\documentclass[english,british,10pt,twocolumn,letterpaper]{article}
\usepackage[T1]{fontenc}
\usepackage[utf8]{inputenc}
\usepackage{color}
\usepackage{array}
\usepackage{multirow}
\usepackage{amsmath}
\usepackage{amssymb}
\usepackage{graphicx}
\PassOptionsToPackage{normalem}{ulem}
\usepackage{ulem}

\makeatletter

\providecommand{\tabularnewline}{\\}

\usepackage{cvpr}
\usepackage{times}
\usepackage{epsfig}
\usepackage{graphicx}
\usepackage{amsmath}
\usepackage{amssymb}

\usepackage{caption} 
\usepackage{colortbl}
\usepackage{arydshln}
\usepackage{makecell, pict2e}

\definecolor{lightgray}{gray}{0.8}

\usepackage[pagebackref=true,breaklinks=true,letterpaper=true,colorlinks,bookmarks=false]{hyperref}

 \cvprfinalcopy 

\ifcvprfinal\fi

\@ifundefined{showcaptionsetup}{}{%
 \PassOptionsToPackage{caption=false}{subfig}}
\usepackage{subfig}
\makeatother

\usepackage{babel}
\begin{document}

\title{Weakly Supervised Object Boundaries}

\author{Anna Khoreva\textsuperscript{1}\hspace{1.5em}Rodrigo Benenson\textsuperscript{1}\hspace{1.5em}Mohamed
Omran\textsuperscript{1}\hspace{1.5em}Matthias Hein\textsuperscript{2}\hspace{1.5em}Bernt
Schiele\textsuperscript{1}\\\\\textsuperscript{1}Max Planck Institute
for Informatics, Saarbr{\"u}cken, Germany\\ \textsuperscript{2}Saarland
University, Saarbr{\"u}cken, Germany}
\maketitle
\begin{abstract}
State-of-the-art learning based boundary detection methods require
extensive training data. Since labelling object boundaries is one
of the most expensive types of annotations, there is a need to relax
the requirement to carefully annotate images to make both the training
more affordable and to extend the amount of training data. In this
paper we propose a technique to generate weakly supervised annotations
and show that bounding box annotations alone suffice to reach high-quality
object boundaries without\emph{ }using any object-specific boundary
annotations. With the proposed weak supervision techniques we achieve
the top performance on the object boundary detection task, outperforming
by a large margin the current fully supervised state-of-the-art methods.
\end{abstract}
\global\long\def\FH{\mbox{F\ensuremath{\mathfrak{\&}}H}}

\makeatletter 
\renewcommand{\paragraph}{%
\@startsection{paragraph}{4}%
{\z@}{0.5ex \@plus 1ex \@minus .2ex}{-0.5em}%
{\normalfont \normalsize \bfseries}%
}
\makeatother\setlength{\textfloatsep}{1.5em}

\section{\label{sec:Introduction}Introduction}

Boundary detection is a classic computer vision problem. It is an
enabling ingredient for many vision tasks such as image/video segmentation
\cite{ArbelaezMaireFowlkesMalikPAMI11,Galasso2013Iccv}, object proposals
\cite{Hosang2015Pami}, object detection \cite{Zhu2015CvprDeepM},
and semantic labelling \cite{BanicaS15}. Rather than image edges,
many of these tasks require class specific objects boundaries. These
are the external boundaries of object instances belonging to a specific
class (or class set).

State-of-the-art boundary detection is obtained via machine learning
which requires extensive training data. Yet, instance-wise boundaries
are amongst the most expensive types of annotations. Compared to two
clicks for a bounding box, delineating an object requires a polygon
with 20\textasciitilde{}100 points, i.e. at least $10\times$ more
effort per object.

In order to make the training of new object classes affordable, and/or
to increase the size of the models we train, there is a need to relax
the requirement of high-quality image annotations. Hence the starting
point of this paper is the following question: is it possible to obtain
object-specific boundaries without having any object boundary annotations
at training time?

\begin{figure}
\vspace{-1em}

\noindent \begin{centering}
\hfill{}\subfloat[\label{fig:Image-teaser}Image]{\protect\includegraphics[width=0.32\columnwidth,height=0.08\textheight]{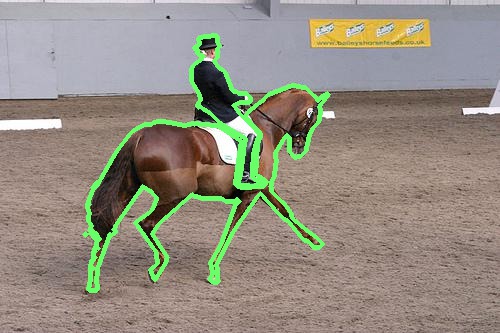}}\hfill{}\subfloat[SE(VOC)]{\protect\includegraphics[width=0.32\columnwidth,height=0.08\textheight]{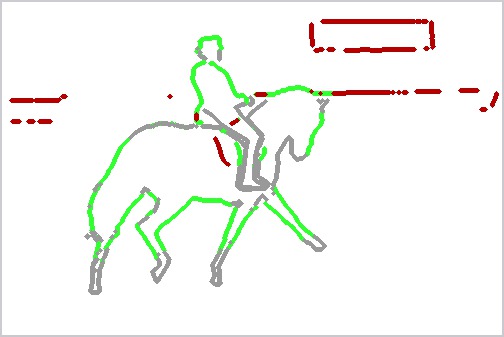}}\hfill{}\subfloat[$\mbox{Det.}\negmedspace+\negmedspace\mbox{SE}\left(\mbox{VOC}\right)$]{\protect\includegraphics[width=0.32\columnwidth,height=0.08\textheight]{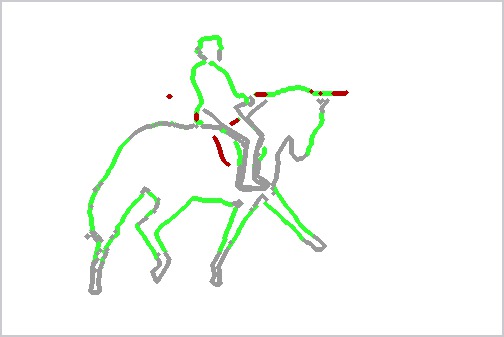}}\hfill{}\vspace{-0.5em}

\par\end{centering}

\noindent \begin{centering}
\hfill{}\subfloat[\label{fig:SE(BSDS)-example}SE(BSDS)]{\protect\includegraphics[width=0.32\columnwidth,height=0.08\textheight]{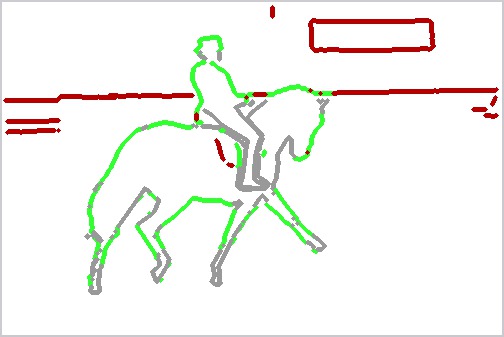}}\hfill{}\subfloat[\label{fig:HED-weak-example}$\textrm{SE}\left(\textrm{weak}\right)$]{\protect\includegraphics[width=0.32\columnwidth,height=0.08\textheight]{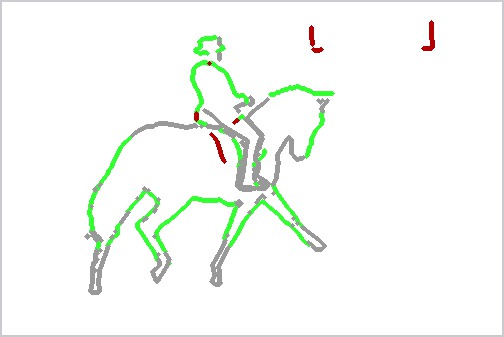}}\hfill{}\subfloat[\label{fig:SE-weak-example}$\mbox{Det.}\negmedspace+\negmedspace\mbox{SE}\left(\mbox{weak}\right)$]{\protect\includegraphics[width=0.32\columnwidth,height=0.08\textheight]{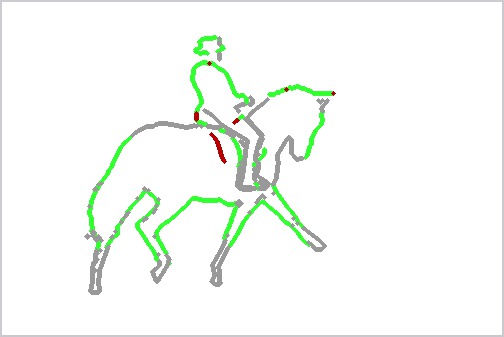}}\hfill{}
\par\end{centering}

\protect\caption{\label{fig:teaser}Object-specific boundaries \ref{fig:Image-teaser}
differ from generic boundaries (such as the ones detected in \ref{fig:SE(BSDS)-example}).
The proposed weakly supervised approach drives boundary detection
towards the objects of interest. Example results in \ref{fig:HED-weak-example}
and \ref{fig:SE-weak-example}. Red/green indicate false/true positive
pixels, grey is missing recall. All methods shown at $50\%$ recall.}
\end{figure}

In this paper we focus on learning object boundaries in a weakly supervised
fashion and show that high quality object boundary detection can be
obtained without\emph{ }using any class-specific boundary annotations.
We propose several ways of generating object boundary annotations
with different levels of supervision, from just using a bounding box
oriented object detector to using the boundary detector trained on
generic boundaries. For generating weak object boundary annotations
we consider different sources, fusing unsupervised image segmentation
\cite{Felzenszwalb2004Ijcv} and object proposal methods \cite{Uijlings2013Ijcv,PontTuset2015ArxivMcg}
with object detectors \cite{Girshick2015IccvFastRCNN,Ren2015NipsFasterRCNN}.
We show that bounding box annotations alone suffice to achieve objects
boundary estimates with high quality.

We present results using a decision forest \cite{Dollar2015Pami}
and a convnet edge detector \cite{Xie2015Iccv}. We report top performance
on Pascal object boundary detection \cite{Hariharan2011Iccv,Everingham15}
with our weak-supervision approaches already surpassing previously
reported fully supervised results.

Our main contributions are summarized below:

\selectlanguage{english}%
$\bullet$\foreignlanguage{british}{ We introduce the problem of weakly
supervised object-specific boundary detection.}

$\bullet$\foreignlanguage{british}{ We show that good performance
can be obtained on BSDS, PascalVOC12, and SBD boundary estimation
using only weak-supervision (leveraging bounding box detection annotations
without the need of instance-wise object boundary annotations).}

$\bullet$\foreignlanguage{british}{ We report best known results
on PascalVOC12, and SBD datasets. Our weakly supervised results alone
improve over the previous fully supervised state-of-the-art.}

\selectlanguage{british}%
The rest of this paper is organized as follows. Section \ref{sec:tasks}
describes different types of boundary detection and the considered
datasets. In Section \ref{sec:Robustness} we investigate the robustness
to annotation noise during training. We leverage our findings and
propose several approaches for generating weak boundary annotations
in Section \ref{sec:Weak-groundtruth}. Sections \ref{sec:Pascal-boundaries-SE}-\ref{sec:SBD-boundaries}
report results using the two different classifier architectures.

\section{\label{sub:Related-work}Related work}

\paragraph{Generic boundaries}

Boundary detection has been regained attention recently. Early methods
are based on a fixed prior model of what constitutes a boundary (e.g.
Canny \cite{Canny1986Pami}). Modern methods leverage machine learning
to push performance. From well crafted features and simple classifiers
(gPb \cite{ArbelaezMaireFowlkesMalikPAMI11}), to powerful decision
trees over fixed features (SE \cite{Dollar2015Pami}, OEF \cite{Hallman2015Cvpr}),
and recently to end-to-end learning via convnets (DeepEdge \cite{Bertasius2015Cvpr},
N4 \cite{Ganin2014Accv}, HED \cite{Xie2015Iccv}). Convnets are usually
pre-trained on large classification datasets, so as to be initialized
with reasonable features. The more sophisticated the model, the more
data is needed to learn it.\\
Other than pure boundary detection, segmentation techniques (such
as $\FH$ \cite{Felzenszwalb2004Ijcv}, gPb-owt-ucm \cite{ArbelaezMaireFowlkesMalikPAMI11},
and MCG \cite{PontTuset2015ArxivMcg}), can also be used to improve
or to generate closed contours.\\
A few works have addressed unsupervised detection of generic boundaries
\cite{Isola2014Eccv,Li2015ArxivULE}. PMI \cite{Isola2014Eccv} detects
boundaries by modelling them as statistical anomalies amongst all
local image patches, reaching competitive performance without the
need for training. Recently \cite{Li2015ArxivULE} proposes to train
edge detectors using motion boundaries obtained from a large corpus
of video data in place of human supervision. Both approaches reach
similar detection performance.

\paragraph{Object-specific boundaries}

In many applications, there is interest to focus on boundaries of
specific object classes. The class-specific object boundary detectors
need then to be trained or tuned to the classes of interest. This
problem is more recent and still relatively unexplored. \cite{Hariharan2011Iccv}
introduced the SBD dataset to measure this task over the 20 pascal
categories. \cite{Hariharan2011Iccv} proposes to re-weight generic
boundaries using the activation regions of a detector. \cite{Uijlings2015Cvpr}
proposed to train class-specific boundary detectors, and weighted
them at test time according to an image classifier.

\paragraph{Weakly supervised learning}

In this work we are interested in object-specific boundaries \emph{without
using class specific boundary annotation}s. We only use bounding box
annotations, and in some experiments, generic boundaries (from BSDS
\cite{ArbelaezMaireFowlkesMalikPAMI11}). Multiple works have addressed
weakly supervised learning for object localization \cite{Oquab2015Cvpr,Cao2015Iccv},
object detection \cite{Prest2012Cvpr,Wang2014EccvWeakLoc}, or semantic
labelling \cite{Vezhnevets2011Iccv,Xu2015CvprWeakSegmentation,Pinheiro2015Cvpr}.
To the best of our knowledge there is no previous work attempting
to learn object boundaries in a weakly supervised fashion.

\section{\label{sec:tasks}Boundary detection tasks}

\begin{figure}
\vspace{-1em}

\begin{centering}
\hfill{}\subfloat[BSDS \cite{ArbelaezMaireFowlkesMalikPAMI11}]{\protect\centering{}\protect\includegraphics[width=0.45\columnwidth,height=0.23\columnwidth]{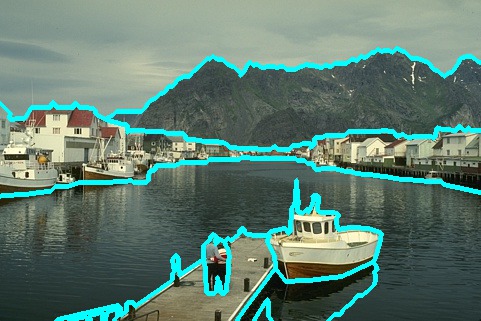}\protect}\hfill{}\subfloat[VOC12 \cite{Everingham15}]{\protect\centering{}\protect\includegraphics[width=0.45\columnwidth,height=0.23\columnwidth]{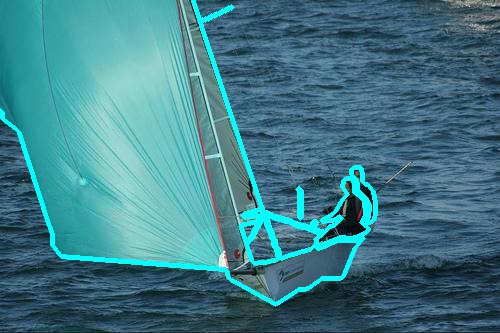}\protect}\hfill{}\linebreak{}
\vspace{-1em}

\par\end{centering}

\begin{centering}
\hfill{}\subfloat[COCO \cite{Lin2014EccvCoco}]{\protect\begin{centering}
\protect\includegraphics[width=0.45\columnwidth,height=0.23\columnwidth]{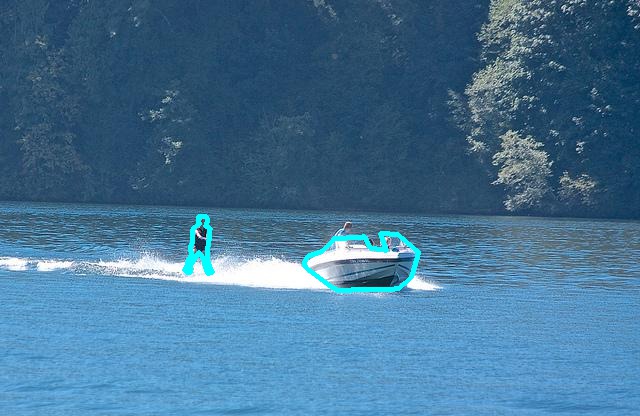}\protect
\par\end{centering}

\protect\centering{}\protect}\hfill{}\subfloat[SBD \cite{Hariharan2011Iccv}]{\protect\centering{}\protect\includegraphics[width=0.45\columnwidth,height=0.23\columnwidth]{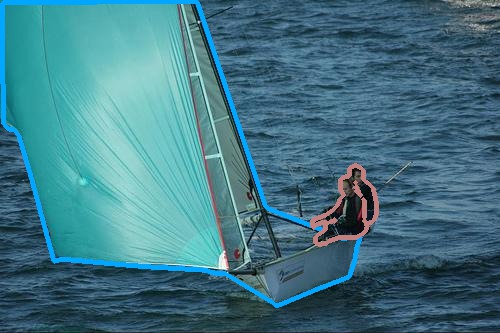}\protect}\hfill{}\protect\caption{\label{fig:datasets}Datasets considered.}

\par\end{centering}

\vspace{-0.5em}
\end{figure}
In this work we distinguish three types of boundaries: generic boundaries
(``things'' and ``stuff''), instance-wise boundaries (external
object instance boundaries), and class specific boundaries (object
instance boundaries of a certain semantic class). For detecting these
three types of boundaries we consider different datasets: BSDS500
\cite{ArbelaezMaireFowlkesMalikPAMI11,MartinFTM01}, Pascal VOC12
\cite{Everingham15}, MS COCO \cite{Lin2014EccvCoco}, and SBD \cite{Hariharan2011Iccv},
where each represents boundary annotations of a given boundary type
(see Figure \ref{fig:datasets}).

\paragraph{BSDS}

We first present our results on the Berkeley Segmentation Dataset
and Benchmark (BSDS) \cite{ArbelaezMaireFowlkesMalikPAMI11,MartinFTM01},
the most established benchmark for generic boundary detection task.
The dataset contains $200$ training, $100$ validation and $200$
test images. Each image has multiple ground truth annotations. For
evaluating the quality of estimated boundaries three measures are
used: fixed contour threshold (ODS), per-image best threshold (OIS),
and average precision (AP). Following the standard approach \cite{Dollar2015Pami,Canny1986Pami}
prior to evaluation we apply a non-maximal suppression technique to
boundary probability maps to obtain thinned edges.

\paragraph{VOC}

For evaluating instance-wise boundaries we propose to use the PASCAL
VOC 2012 (VOC) segmentation dataset \cite{Everingham15}. The dataset
contains $1\,464$ training and $1\,449$ validation images, annotated
with contours for 20 object classes for all instances. The dataset
was originally designed for semantic segmentation. Therefore only
object interior pixels are marked and the boundary location is recovered
from the segmentation mask. Here we consider only object boundaries
without distinguishing the semantics, treating all 20 classes as one.
For measuring the quality of predicted boundaries the BSDS evaluation
software is used. Following \cite{Uijlings2015Cvpr} the maxDist (maximum
tolerance for edge match) is set to $0.01$.

\paragraph{COCO}

To show generalization of the proposed method for instance-wise boundary
detection we use the MS COCO (COCO) dataset \cite{Lin2014EccvCoco}.
The dataset provides semantic segmentation masks for 80 object classes.
For our experiments we consider only images that contain the 20 Pascal
classes and objects larger than 200 pixels. The subset of COCO that
contains Pascal classes consists of $65\,813$ training and $30\,163$
validation images. For computational reasons we limit evaluation to
$5\,000$ randomly chosen images of the validation set. The BSDS evaluation
software is used ($\textrm{maxDist}=0.01$). Only object boundaries
are evaluated without distinguishing the semantics.

\paragraph{SBD}

We use the Semantic Boundaries Dataset (SBD) \cite{Hariharan2011Iccv}
for evaluating class specific object boundaries. The dataset consists
of $11\,318$ images from the trainval set of the PASCAL VOC2011 challenge,
divided into $8\,498$ training and $2\,820$ test images. This dataset
has object instance boundaries with accurate figure/ground masks that
are also labeled with one of 20 Pascal VOC classes. The boundary detection
accuracy for each class is evaluated using the official evaluation
software \cite{Hariharan2011Iccv}. During the evaluation process
all internal object-specific boundaries are set to zero and the maxDist
is set to $0.02$. We report the mean ODS F-measure (F), and average
precision (AP) across 20 classes. 

Note that VOC and SBD datasets have overlap between their train and
test sets. When doing experiments across datasets we make sure not
to re-use any images included in the test set considered.\vspace{-0.1em}

\paragraph{Baselines}

For our experiments we consider two different types of boundary detectors
- SE \cite{Dollar2015Pami} and HED \cite{Xie2015Iccv} - as baselines.\\
SE is at the core of multiple related methods (SCG, MCG, OEF). SE
\cite{Dollar2015Pami} builds a ``structured decision forest'' which
is a modified decision forest, where the leaf outputs are local boundary
patches ($16\times16\ \mbox{pixels}$) that are averaged at test time,
and the split nodes are built taking into account the local segmentation
of the ground truth input patches. It uses binary comparison over
hand-crafted edge and self-similarity features as split decisions.
By construction this method requires closed contours (i.e. segmentations)
as training input. This detector is reasonably fast to train/test
and yields good detection quality.\\
HED \cite{Xie2015Iccv} is currently the top performing convnet for
BSDS boundaries. It builds upon a VGG16 network pre-trained on ImageNet
\cite{Simonyan2015Iclr}, and exploits features from all layers to
build its output boundary probability map. By also exploiting the
lower layers (which have higher resolution) the output is more detailed,
and the fine-tuning is more effective (since all layers are guided
directly towards the boundary detection task). To reach top performance,
HED is trained using a subset of the annotated BSDS pixels, where
all annotators agree \cite{Xie2015Iccv}. These are so called ``consensus''
annotations \cite{Hou2013Cvpr}, and correspond to sparse $\sim\negthinspace15\%$
of all true positives.
\begin{figure}
\noindent \begin{centering}
\includegraphics[bb=0bp 0bp 540bp 371bp,width=0.9\columnwidth,height=0.22\textheight]{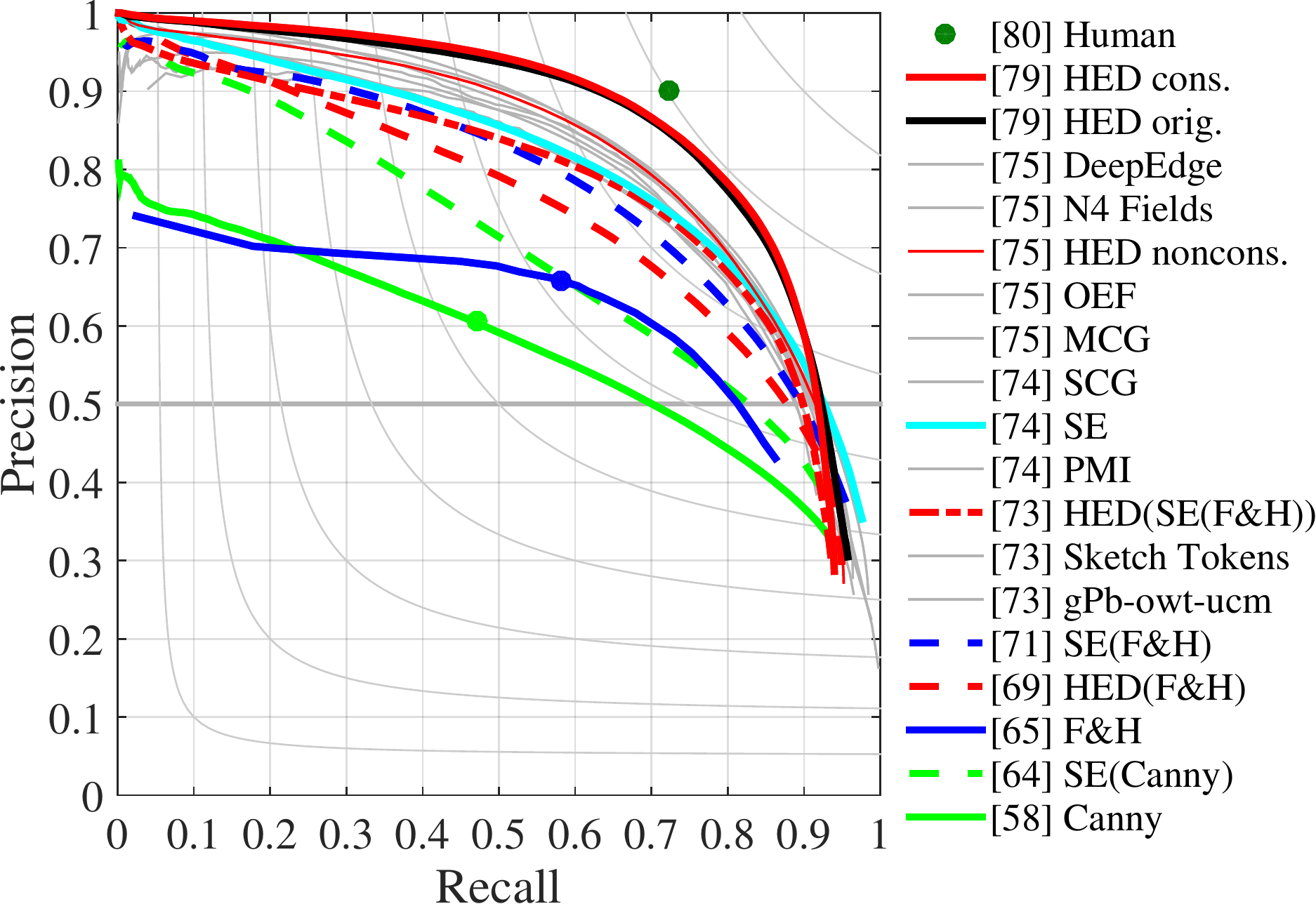}
\par\end{centering}

\protect\caption{\label{fig:BSD500-results}BSDS results. Canny and $\protect\FH$
points indicate the boundaries used as noisy annotations. When trained
over noisy annotations, both SE and HED provide a large quality improvement.}
\end{figure}
\begin{table}
\centering{}\vspace{-1em}
\setlength{\tabcolsep}{2.5pt} 
\renewcommand{\arraystretch}{0.85}%
\begin{tabular}{clcccc}
{\small{}Family} & {\small{}Method} & {\small{}ODS} & {\small{}OIS} & {\small{}AP} & {\small{}$\Delta\mbox{AP}\%$}\tabularnewline
\hline 
\hline 
\multirow{3}{*}{{\small{}}%
\begin{tabular}{c}
{\small{}Unsupervised}\tabularnewline
\end{tabular}} & {\small{}Canny} & {\small{}58} & {\small{}62} & {\small{}55} & {\small{}-}\tabularnewline
 & {\small{}$\FH$} & {\small{}64} & {\small{}67} & {\small{}64} & {\small{}-}\tabularnewline
 & {\small{}PMI} & {\small{}74} & {\small{}77} & {\small{}78} & {\small{}-}\tabularnewline
\hline 
\multirow{4}{*}{{\small{}}%
\begin{tabular}{c}
{\small{}Trained}\tabularnewline
{\small{}on}\tabularnewline
{\small{}ground truth}\tabularnewline
\end{tabular}} & {\small{}gPb-owt-ucm} & {\small{}73} & {\small{}76} & {\small{}73} & {\small{}-}\tabularnewline
 & {\small{}SE(BSDS)} & {\small{}74} & {\small{}76} & {\small{}\uline{79}} & {\small{}-}\tabularnewline
 & {\small{}HED(BSDS) noncons. } & {\small{}75} & {\small{}77} & {\small{}\uline{80}} & {\small{}-}\tabularnewline
 & {\small{}HED(BSDS) cons.} & {\small{}79} & {\small{}81} & {\small{}84} & {\small{}-}\tabularnewline
\hline 
\multirow{6}{*}{{\small{}}%
\begin{tabular}{c}
{\small{}Trained}\tabularnewline
{\small{}on}\tabularnewline
{\small{}unsupervised}\tabularnewline
{\small{}boundary}\tabularnewline
{\small{}estimates}\tabularnewline
\end{tabular}} & {\small{}$\mbox{SE}\left(\mbox{Canny}\right)$} & {\small{}64} & {\small{}67} & {\small{}64} & {\small{}38}\tabularnewline
 & {\small{}$\mbox{SE}\left(\FH\right)$} & {\small{}71} & {\small{}74} & \textbf{\small{}76} & {\small{}80}\tabularnewline
 & {\small{}$\mbox{SE}\left(\mbox{SE}\left(\FH\right)\right)$} & {\small{}72} & {\small{}74} & {\small{}76} & {\small{}80}\tabularnewline
 & {\small{}SE(PMI)} & {\small{}72} & {\small{}75} & {\small{}77} & {\small{}-}\tabularnewline
 & {\small{}$\mbox{HED}\left(\FH\right)$} & {\small{}69} & {\small{}72} & {\small{}73} & {\small{}56}\tabularnewline
 & {\small{}$\mbox{HED}\left(\mbox{SE}\left(\FH\right)\right)$} & {\small{}73} & {\small{}76} & \textbf{\small{}75} & {\small{}69}\tabularnewline
\end{tabular}\protect\caption{\label{tab:BSD500-results}Detailed BSDS results, see Figure \ref{fig:BSD500-results}
and Section \ref{sec:Robustness}. Underline indicates ground truth
baselines, and bold are our best weakly supervised results. $\left(\cdot\right)$
denotes the data used for training. $\Delta\mbox{AP}\%$ indicates
the ratio between the same model trained on ground truth, and the
noisy input boundaries. The closer to $100\%$, the lower the drop
due to using noisy inputs instead of ground truth. }
\vspace{-0em}
\end{table}

\section{\label{sec:Robustness}Robustness to annotation noise}

We start by exploring weakly supervised training for generic boundary
detection, as considered in BSDS.

Model based approaches such as Canny \cite{Canny1986Pami} and $\FH$
\cite{Felzenszwalb2004Ijcv} are able to provide low quality boundary
detections. We notice that correct boundaries tend to have consistent
appearance, while erroneous detections are mostly inconsistent. Robust
training methods should be able to pick-up the signal in such noisy
detections.

\paragraph{SE}

In Figure \ref{fig:BSD500-results} and Table \ref{tab:BSD500-results}
we report our results when training a structured decision forest (SE)
and a convnet (HED) with noisy boundary annotations. By $\left(\cdot\right)$
we denote the data used for training. When training SE using either
Canny ({\small{}$\mbox{SE}\left(\mbox{Canny}\right)$}) or $\FH$
($\mbox{SE}\left(\FH\right)$) we observe a notable jump in boundary
detection quality. Comparing SE trained with the BSDS ground truth
(fully supervised, $\mbox{SE}\left(\mbox{BSDS}\right)$), with the
noisy labels from $\FH$, $\mbox{SE}\left(\FH\right)$ closes up to
$80\%$ of the gap between $\mbox{SE}\left(\FH\right)$ and $\mbox{SE}\left(\mbox{BSDS}\right)$
($\Delta\mbox{AP}\%$ column in Table \ref{tab:BSD500-results}).
Using only noisy weak supervision $\mbox{SE}\left(\FH\right)$ is
only $3$ AP percent points behind from the fully supervised case
($76$ vs. $79$). \\
We believe that the strong noise robustness of SE can be attributed
to the way it builds its leaves. The final output of each leaf is
the medoid of all segments reaching it. If the noisy boundaries are
randomly spread in the image appearance space, the medoid selection
will be robust.
\begin{figure*}
\noindent \begin{centering}
\vspace{-1em}

\par\end{centering}

\noindent \begin{centering}
\hfill{}\subfloat[\label{fig:Ground-truth}Ground truth]{\protect\includegraphics[width=0.19\textwidth]{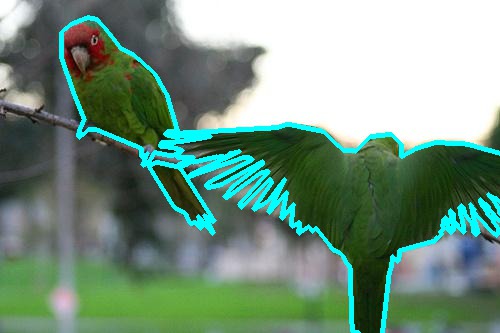}}\hfill{}\subfloat[\label{fig:F=000026H}$\protect\FH$]{\protect\includegraphics[width=0.19\textwidth]{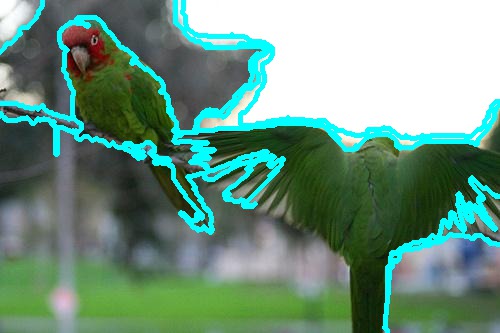}}\hfill{}\subfloat[\label{fig:F=000026H-BBs}$\protect\FH\cap\mbox{BBs}$]{\protect\includegraphics[width=0.19\textwidth]{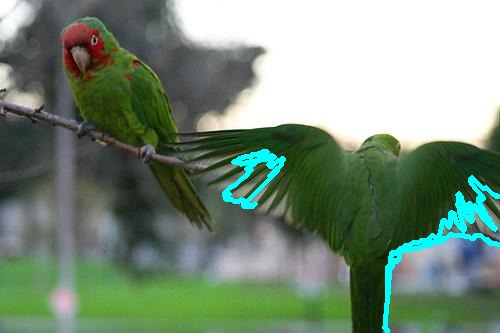}}\hfill{}\subfloat[\label{fig:GrabCut-BBs}GrabCut $\cap$ BBs]{\protect\includegraphics[width=0.19\textwidth]{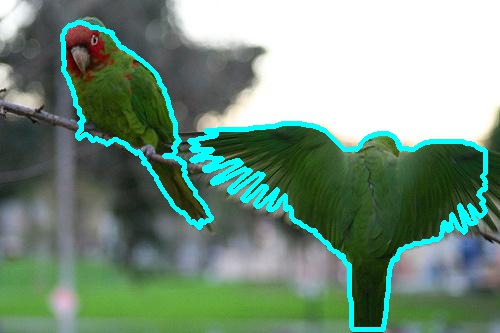}}\hfill{}\subfloat[\label{fig:SeSe-BBs}SeSe $\cap$ BBs]{\protect\includegraphics[width=0.19\textwidth]{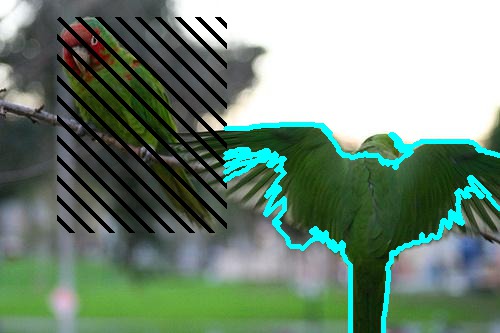}}\hfill{}\vspace{-1em}

\par\end{centering}

\noindent \begin{centering}
\hfill{}\subfloat[\label{fig:MCG-BBs}MCG $\cap$ BBs]{\protect\includegraphics[width=0.19\textwidth]{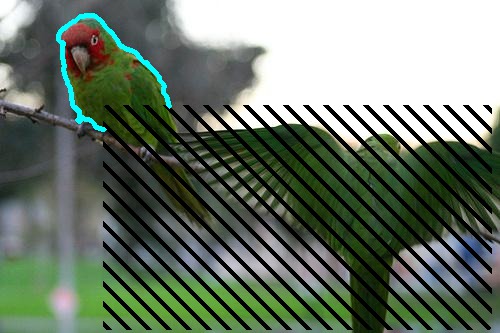}}\hfill{}\subfloat[\label{fig:cons.-MCG-BBs}cons. MCG $\cap$ BBs]{\protect\includegraphics[width=0.19\textwidth]{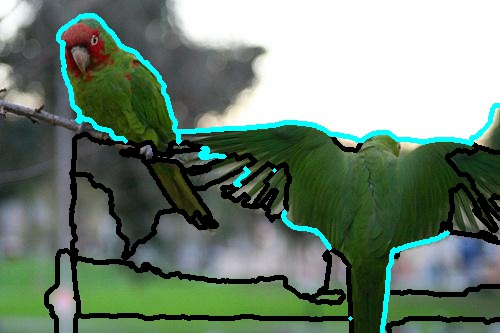}}\hfill{}\subfloat[\label{fig:SE(SeSe-BBs)}SE(SeSe $\cap$ BBs)]{\protect\includegraphics[width=0.19\textwidth]{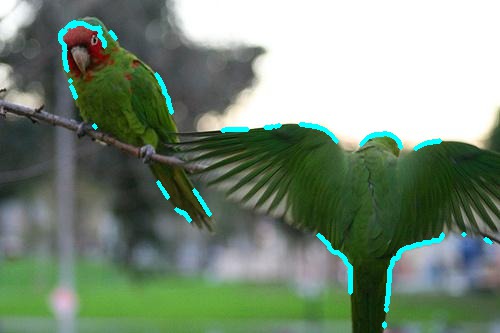}}\hfill{}\subfloat[\label{fig:cons.-all-methods-sese}cons. S\&G$\cap$BBs]{\protect\includegraphics[width=0.19\textwidth]{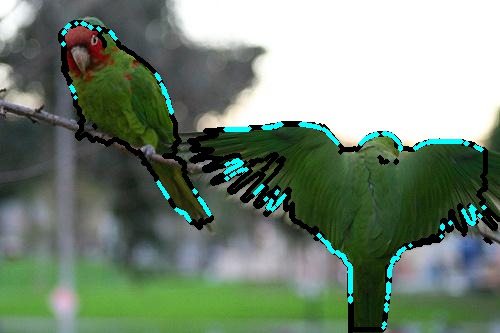}}\hfill{}\subfloat[\label{fig:cons.-all-methods}cons. all methods $\cap$ BBs]{\protect\includegraphics[width=0.19\textwidth]{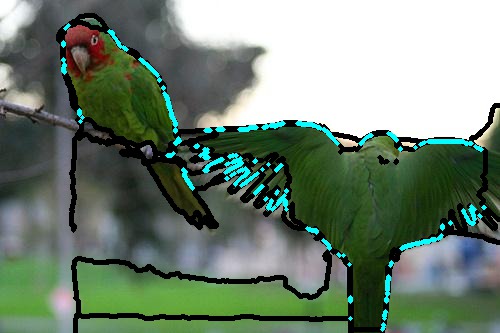}}\hfill{}
\par\end{centering}

\protect\caption{\label{fig:Noisy-boundaries}Different generated boundary annotations.
Cyan/black indicates positive/ignored boundaries.}

\vspace{-0.5em}
\end{figure*}

\paragraph{HED}

The HED convnet \cite{Xie2015Iccv} reaches top quality when trained
over consensus annotations. When using all annotations (``non consensus''),
its performance is comparable to other convnet alternatives. When
trained over $\FH$ the relative improvement is smaller than for the
SE case, when combined with SE (denoted ``$\mbox{HED(SE}\left(\FH\right))$'')
it reaches $69\ \Delta\mbox{AP}\%$ . $\mbox{HED}\left(\mbox{SE}\left(\FH\right)\right)$
provides better boundaries than $\mbox{SE}\left(\FH\right)$ alone,
and reaches a quality comparable to the classic gPb method \cite{ArbelaezMaireFowlkesMalikPAMI11}
($75$ vs. $73$).

On BSDS the unsupervised PMI methods provides better boundaries than
our weakly supervised variants. However PMI cannot be adapted to provide
object-specific boundaries. For this we need to rely on methods than
can be trained, such as SE and HED.

\paragraph{Conclusion}

SE is surprisingly robust annotation noise during training. HED is
also robust but to a lesser degree. By using noisy boundaries generated
from unsupervised methods, we can reach a performance comparable to
the bulk of current methods.

\section{\label{sec:Weak-groundtruth}Weakly supervised generation of boundary
annotations}

Based on the observations in Section \ref{sec:Robustness}, we propose
to train boundary detectors using data generated from weak annotations.
Our weakly supervised models are trained in a regular fashion, but
use generated (noisy) training data as input instead of human annotations. 

We consider boundary annotations generated with three different levels
of supervision: fully unsupervised, using only detection annotations,
and using both detection annotations and BSDS boundary annotations
(e.g. using generic boundary annotation, but zero object-specific
boundaries). In this section we present the different variants of
weakly supervised boundary annotations. Some of them are illustrated
in Figure \ref{fig:Noisy-boundaries}.

\paragraph{$\mbox{BBs}$}

We use the bounding box annotations to train a class-specific object
detector \cite{Ren2015NipsFasterRCNN,Girshick2015IccvFastRCNN}. We
then apply this detector over the training set (and possibly a larger
set of images), and retain boxes with confidence scores above $0.8$.
(We also experimented using directly the ground truth annotations,
but saw no noticeable difference; thus we report only numbers using
the ``detections over the training set'').

\paragraph{$\protect\FH$}

As a source of unsupervised boundaries we consider the classical graph
based image segmentation technique proposed by \cite{Felzenszwalb2004Ijcv}
($\FH$). To focus the training data on the classes of interest, we
intersect these boundaries with detection bounding boxes from \cite{Ren2015NipsFasterRCNN}
(\textbf{$\FH\cap\mbox{BBs}$}). Only the boundaries of segments that
are contained inside a bounding box are retained.

\paragraph{GrabCut}

Boundaries from $\FH$ will trigger on any kind of boundary, including
the internal boundaries of objects. A way to exclude internal object
boundaries, is to extract object contours via\textcolor{black}{{} }figure-ground
segmentation of the detection bounding box. We use \textbf{GrabCut}
\cite{RotherGrabCut} for this purpose. We also experimented with
DenseCut \cite{Cheng2015CgfDenseCut} and CNN+GraphCut \cite{Simonyan2014Iclr},
but did not obtain any gain; thus we report only GrabCut results.\\
For the experiments reported below, for \textbf{$\mbox{GrabCut}\cap\mbox{BBs}$}
a segment is only accepted if a detection from \cite{Ren2015NipsFasterRCNN}
has the intersection-over-union score ($\mbox{IoU}$) $\geq0.7$.
If a detection bounding boxes has no matching segment, the whole region
is marked as ignore (see Figure \ref{fig:SeSe-BBs}) and not used
during the training of boundary detectors.

\paragraph{Object proposals}

Another way to bias generation of boundary annotations towards object
contours is to consider object proposals. \textbf{SeSe }\cite{Uijlings2013Ijcv}
is based on the $\FH$ \cite{Felzenszwalb2004Ijcv} segmentation (thus
it is fully unsupervised), while \textbf{MCG} \cite{PontTuset2015ArxivMcg}
employs boundaries estimated via $\mbox{SE}\left(\mbox{BSDS}\right)$
(thus uses generic boundaries annotations).

Similar to $\mbox{GrabCut}\cap\mbox{BBs}$, \textbf{$\mbox{SeSe}\cap\mbox{BBs}$}
and \textbf{$\mbox{MCG}\cap\mbox{BBs}$} are generated by matching
proposals to bounding boxes (if $\mbox{IoU}\geq0.9$). BBs come from
\cite{Girshick2015IccvFastRCNN} with the corresponding object proposals.
When more than one proposal is matched to a detection bounding box
we use the union of the proposal boundaries as positive annotations.
This maximizes the recall of boundaries, and somewhat imitates the
multiple human annotators in BSDS. We also experimented using only
the highest overlapping proposal, but the union provides marginally
better results; thus we report only the later. Since proposals matching
a bounding box might have boundaries outside it, we consider them
all since the bounding box itself might not cover well the underlying
object.

\paragraph{Consensus boundaries}

As pointed out in Table \ref{tab:BSD500-results}, HED requires consensus
boundaries to reach good performance. Thus rather than taking the
union between proposal boundaries, we consider using the consensus
between object proposal boundaries. The boundary is considered to
be present if the agreement is higher than $70\%$, otherwise the
boundary is ignored. We denote such generated annotations as ``cons.'',
e.g. \textbf{$\mbox{cons. MCG}\cap\mbox{BBs}$} (see Figure \ref{fig:cons.-MCG-BBs}).
\\
Another way to generate sparse (consensus-like) boundaries, is to
threshold the boundary probability map out of an $\mbox{SE}\left(\cdot\right)$
model. \textbf{$\mbox{SE}\left(\mbox{SeSe}\cap\mbox{BBs}\right)$}
uses the top $15\%$ quantile per image as weakly supervised annotations.\\
Finally, other than consensus between proposals, we can also do consensus
between methods. \textbf{$\mbox{cons. S\&G}\cap\mbox{BBs}$} is the
intersection between $\mbox{SE}\left(\mbox{SeSe}\cap\mbox{BBs}\right)$,
SeSe and GrabCut boundaries (fully unsupervised); while \textbf{$\mbox{cons. all methods}\cap\mbox{BBs}$}
is the intersection between MCG, SeSe and GrabCut (uses BSDS data).

\paragraph{Datasets}

Since we generate boundary annotations in a weakly supervised fashion,
we are able to generate boundaries over arbitrary image sets. In our
experiments we consider SBD, VOC (segmentation), and $\mbox{VOC}_{+}$
(VOC plus images from Pascal VOC12 detection task). Methods using
$\mbox{VOC}_{+}$are denoted using $\cdot_{+}$ (e.g. $\mbox{SE}\left(\mbox{SeSe}_{+}\cap\mbox{BBs}\right)$).

\section{\label{sec:Pascal-boundaries-SE}Structured forest VOC boundary detection}

In this section we analyse the variants of weakly supervised methods
for object boundary detection proposed in Section \ref{sec:Weak-groundtruth}
as opposed to the fully supervised ones. From now on we are interested
in external boundaries of objects. Therefore we employ the Pascal
VOC12, treating all 20 Pascal classes as one. See details of the evaluation
protocol in Section \ref{sec:tasks}. We start by discussing results
using SE; convnet results are presented in Section \ref{sec:Pascal-boundaries-convnets}.

\subsection{Training models with ground truth}

\paragraph{SE}

\begin{figure}
\noindent \begin{centering}
\includegraphics[width=0.85\columnwidth,height=0.75\columnwidth]{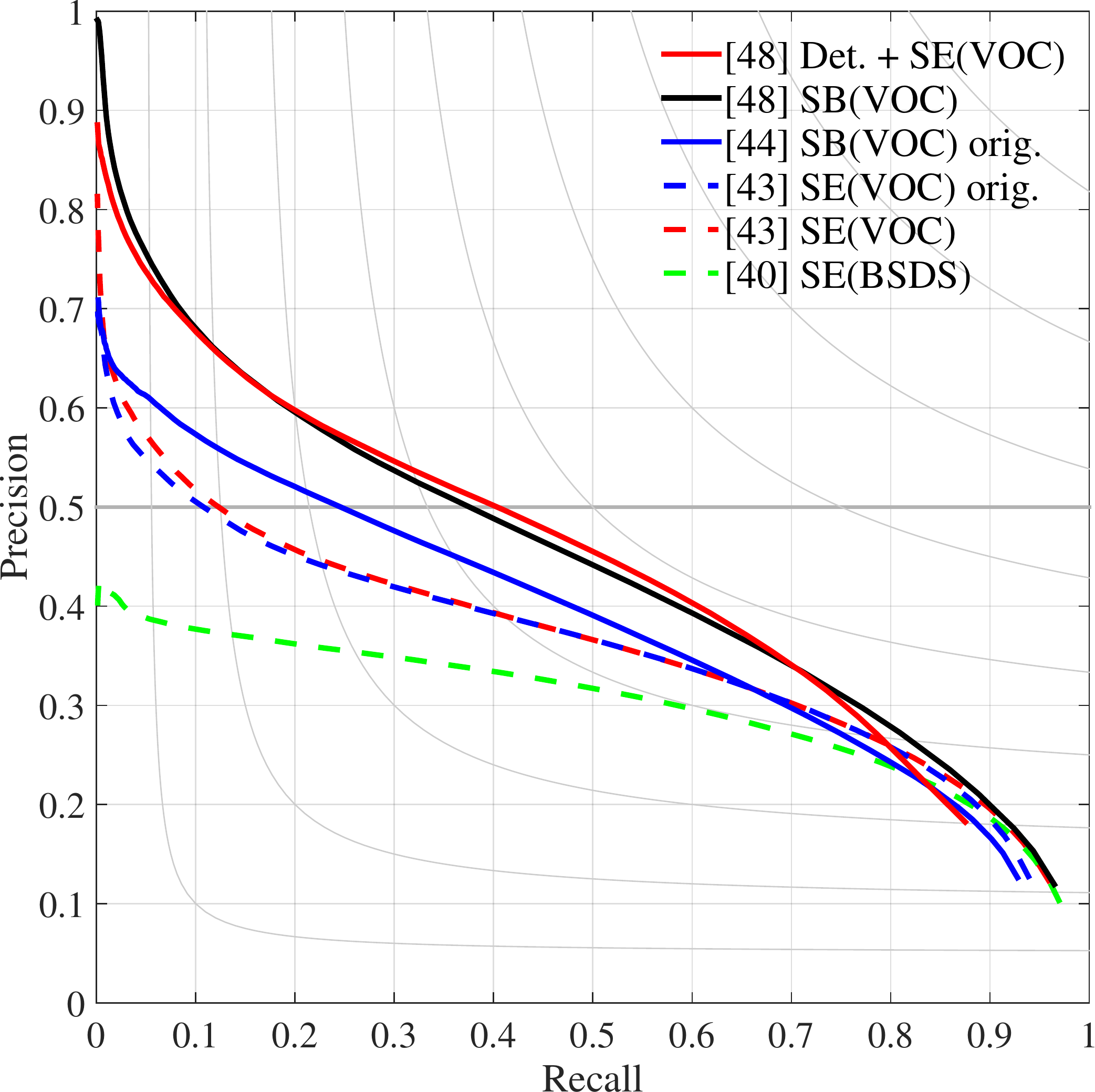}
\par\end{centering}

\protect\caption{\label{fig:VOC-SE-GT-results}VOC12 results, fully supervised SE models.
$\left(\cdot\right)$ denotes the data used for training. Continuous/dashed
line indicates models using/not using a detector at test time. Legend
indicates AP numbers.}
\vspace{-0.5em}
\end{figure}

Figure \ref{fig:VOC-SE-GT-results} and Table \ref{tab:VOC-SE-results}
show results of SE trained over the ground truth of different datasets
(dashed lines). Our results of $\mbox{SE}\left(\mbox{VOC}\right)$
are on par to the ones reported in \cite{Uijlings2015Cvpr}. The gap
between $\mbox{SE}\left(\mbox{VOC}\right)$ and $\mbox{SE}\left(\mbox{BSDS}\right)$
reflects the difference between generic boundaries and boundaries
specific to the 20 VOC object categories (see also Figure \ref{fig:teaser}).

\paragraph{SB }

To improve object-specific boundary detection, the situational boundary
method SB \cite{Uijlings2015Cvpr}, trains 20 class-specific SE models.
These models are combined at test time using a convnet image classifier.
The original SB results and our re-implementation $\mbox{SB}\left(\mbox{VOC}\right)$
are shown in Figure \ref{fig:VOC-SE-GT-results}. Our version obtains
better results ($4$ percent points gain in AP) due to training the
SE models with more samples per image, and using a stronger image
classifier \cite{Simonyan2015Iclr}.

\paragraph{Detector + $\mbox{SE}$}

Rather than training and testing with $20$ SE models plus an image
classifier, we propose to leverage the same training data using a
single SE model together with a detector \cite{Girshick2015IccvFastRCNN}.
By computing a per-pixel maximum among all detection bounding boxes
and their score, we construct an ``objectness map'' that we multiply
with the boundary probability map from SE. False positive boundaries
are thus down-scored, and boundaries in high confidence regions for
the detector get boosted. The detector is trained with the same per
object boundary annotations used to train the SE model, no additional
data is required.

Our $\mbox{Det.}\negmedspace+\negmedspace\mbox{SE}\left(\mbox{VOC}\right)$
obtains the same detection quality as $\mbox{SB}\left(\mbox{VOC}\right)$
while using only a single SE model. These are the best reported results
on this task (top of Table \ref{tab:VOC-SE-results}), when using
the fully supervised training data.

At the cost of more expensive training and test, one could in principle
also combine object detection with the situational boundaries method
\cite{Uijlings2015Cvpr}, this is out of scope of this paper and considered
as future work.

\subsection{Training models using weak annotations}

Given the reference performance of $\mbox{Det.}\negmedspace+\negmedspace\mbox{SE}\left(\mbox{VOC}\right)$,
can we reach similar boundary detection quality without using the
boundary annotations from VOC?

\paragraph{$\mbox{SE}\left(\cdot\right)$}

\begin{figure}
\noindent \begin{centering}
\includegraphics[bb=0bp 0bp 601bp 599bp,width=0.85\columnwidth,height=0.75\columnwidth]{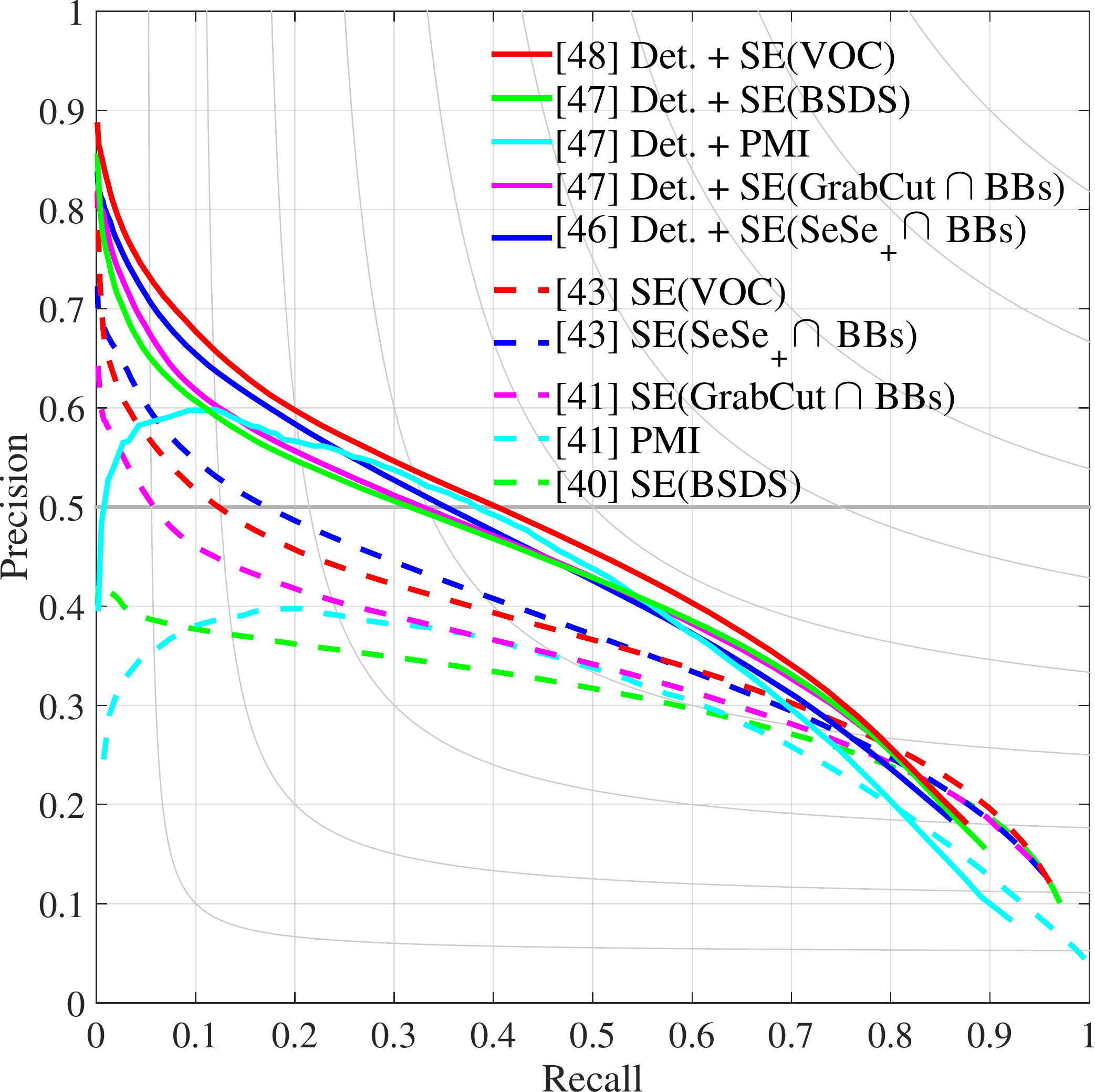}
\par\end{centering}

\protect\caption{\label{fig:VOC-SE-noisy-GT-results}VOC12 results, weakly supervised
SE models. $\left(\cdot\right)$ denotes the data used for training.
Continuous/dashed line indicates models using/not using a detector
at test time. Legend indicates AP numbers.}
\vspace{-0.5em}
\end{figure}

First we consider using a SE model alone at test time. Using only
the $\mbox{BSDS}$ annotations leads to rather low performance (see
$\mbox{SE}\left(\mbox{BSDS}\right)$ in Figure \ref{fig:VOC-SE-noisy-GT-results}).
$\mbox{PMI}$ shows a similar gap. The same BSDS data can be used
to generate $\mbox{MCG}$ object proposals over the VOC training data,
and a detector trained on VOC bounding boxes can generate bounding
boxes over the same images. We combined them together to generate
boundary annotations via $\mbox{MCG}\cap\mbox{BBs}$, as described
in Section \ref{sec:Weak-groundtruth}. The weak supervision from
the bounding boxes can be used to improve the performance of $\mbox{SE}\left(\mbox{BSDS}\right)$.
By extending the training set to additional pascal images ($\mbox{SE}\left(\mbox{MCG}_{+}\cap\mbox{BBs}\right)$
in Table \ref{tab:VOC-SE-results}) we can reach \emph{the same performance}
as when using the ground truth VOC data.\\
We also consider variants that do not leverage the BSDS boundary annotations,
such as $\mbox{SeSe}$ and $\mbox{GrabCut}$. $\mbox{SeSe}$ provides
essentially the same result as MCG.
\begin{table}
\setlength{\tabcolsep}{2.5pt} 
\renewcommand{\arraystretch}{0.85}

\begin{centering}
{\small{}}%
\begin{tabular}{ccc|ccc|ccc}
\multirow{2}{*}{{\footnotesize{}Family}} & \multirow{2}{*}{{\footnotesize{}Method}} & \multirow{2}{*}{{\footnotesize{}Data}} & \multicolumn{3}{c|}{{\footnotesize{}Without BBs}} & \multicolumn{3}{c}{{\footnotesize{}With BBs}}\tabularnewline
 &  &  & {\footnotesize{}F} & {\footnotesize{}AP} & {\footnotesize{}$\textrm{\ensuremath{\Delta}}$AP} & {\footnotesize{}F} & {\footnotesize{}AP} & {\footnotesize{}$\Delta$AP}\tabularnewline
\hline 
\hline 
\multirow{1}{*}{{\footnotesize{}GT}} & {\footnotesize{}SE} & \multirow{1}{*}{{\footnotesize{}VOC}} & {\footnotesize{}\uline{43}} & {\footnotesize{}\uline{35}} & {\footnotesize{}-} & {\footnotesize{}\uline{48}} & {\footnotesize{}\uline{41}} & {\footnotesize{}-}\tabularnewline
\hline 
\multirow{3}{*}{{\footnotesize{}}%
\begin{tabular}{c}
{\footnotesize{}Other}\tabularnewline
{\footnotesize{}GT}\tabularnewline
\end{tabular}} & {\footnotesize{}SE} & {\footnotesize{}COCO} & {\footnotesize{}44} & {\footnotesize{}37} & {\footnotesize{}2} & {\footnotesize{}49} & {\footnotesize{}42} & {\footnotesize{}1}\tabularnewline
 & {\footnotesize{}SE} & \multirow{2}{*}{{\footnotesize{}BSDS}} & {\footnotesize{}40} & {\footnotesize{}29} & {\footnotesize{}-6} & {\footnotesize{}47} & {\footnotesize{}39} & {\footnotesize{}-2}\tabularnewline
 & {\footnotesize{}MCG} &  & {\footnotesize{}41} & {\footnotesize{}28} & {\footnotesize{}-7} & {\footnotesize{}48} & {\footnotesize{}39} & {\footnotesize{}-2}\tabularnewline
\hline 
\multirow{6}{*}{{\footnotesize{}}%
\begin{tabular}{c}
{\footnotesize{}Weakly}\tabularnewline
{\footnotesize{}super-}\tabularnewline
{\footnotesize{}vised}\tabularnewline
\end{tabular}} & \multirow{6}{*}{{\footnotesize{}SE}} & {\footnotesize{}$\FH\cap\mbox{BBs}$} & {\footnotesize{}40} & {\footnotesize{}29} & {\footnotesize{}-6} & {\footnotesize{}46} & {\footnotesize{}36} & {\footnotesize{}-5}\tabularnewline
 &  & {\footnotesize{}$\mbox{GrabCut}\cap\mbox{BBs}$} & {\footnotesize{}41} & {\footnotesize{}32} & {\footnotesize{}-3} & {\footnotesize{}47} & {\footnotesize{}39} & {\footnotesize{}-2}\tabularnewline
 &  & {\footnotesize{}$\mbox{SeSe}\cap\mbox{BBs}$} & {\footnotesize{}42} & {\footnotesize{}35} & {\footnotesize{}0} & {\footnotesize{}46} & {\footnotesize{}39} & {\footnotesize{}-2}\tabularnewline
 &  & {\footnotesize{}$\mbox{SeSe}_{+}\cap\mbox{BBs}$} & \textbf{\footnotesize{}43} & \textbf{\footnotesize{}36} & \textbf{\footnotesize{}+1} & {\footnotesize{}46} & {\footnotesize{}39} & {\footnotesize{}-2}\tabularnewline
 &  & {\footnotesize{}$\mbox{MCG}\cap\mbox{BBs}$} & {\footnotesize{}43} & {\footnotesize{}34} & {\footnotesize{}-1} & {\footnotesize{}47} & {\footnotesize{}39} & {\footnotesize{}-2}\tabularnewline
 &  & {\footnotesize{}$\mbox{MCG}_{+}\cap\mbox{BBs}$} & {\footnotesize{}43} & {\footnotesize{}35} & {\footnotesize{}0} & \textbf{\footnotesize{}48} & \textbf{\footnotesize{}40} & \textbf{\footnotesize{}-1}\tabularnewline
\hline 
\multirow{2}{*}{{\footnotesize{}}%
\begin{tabular}{c}
{\footnotesize{}Unsuper-}\tabularnewline
{\footnotesize{}vised}\tabularnewline
\end{tabular}} & {\footnotesize{}$\FH$} & \multirow{2}{*}{{\footnotesize{}-}} & {\footnotesize{}34} & {\footnotesize{}15} & {\footnotesize{}-20} & {\footnotesize{}41} & {\footnotesize{}25} & {\footnotesize{}-16}\tabularnewline
 & {\footnotesize{}PMI} &  & {\footnotesize{}41} & {\footnotesize{}29} & {\footnotesize{}-6} & {\footnotesize{}47} & {\footnotesize{}38} & {\footnotesize{}-3}\tabularnewline
\end{tabular}
\par\end{centering}{\small \par}

\protect\caption{\label{tab:VOC-SE-results}VOC results for SE models, see Figures
\ref{fig:VOC-SE-GT-results} and \ref{fig:VOC-SE-noisy-GT-results}.
Underline indicates ground truth baselines, and bold are our best
weakly supervised results.}
\vspace{-0.5em}
\end{table}

\paragraph{$\mbox{Det.}\negmedspace+\negmedspace\mbox{SE}\left(\cdot\right)$}

Applying object detection at test time squashes the differences among
all weakly supervised methods. $\mbox{Det.}\negmedspace+\negmedspace\mbox{PMI}$
shows strong results, but (since not trained on boundaries) fails
to reach high precision. The high quality of $\mbox{Det.}\negmedspace+\negmedspace\mbox{BSDS}$
indicates that $\mbox{BSDS}$ annotations, despite being in principle
``generic boundaries'' in practice reflect well object boundaries,
at least in the proximity of an object. This is further confirmed
in Section \ref{sec:Pascal-boundaries-convnets}.\\
Compared to $\mbox{Det.}\negmedspace+\negmedspace\mbox{BSDS}$ our
weakly supervised annotation variants further close the gap to $\mbox{Det.}\negmedspace+\negmedspace\mbox{SE}\left(\mbox{VOC}\right)$
(especially in high precision area), even when not using any BSDS
data.

\paragraph{Conclusion}

Based only on bonding box annotations, our weakly supervised boundary
annotations enable the $\mbox{Det.}\negmedspace+\negmedspace\mbox{SE}$
model to match the fully supervised case, improving over the best
reported results on the task. We also observe that BSDS data allows
to train models that describe well object boundaries.

\section{\label{sec:Pascal-boundaries-convnets}Convnet VOC boundary detection
results}

\begin{figure}
\noindent \begin{centering}
\includegraphics[width=0.9\columnwidth,height=0.8\columnwidth]{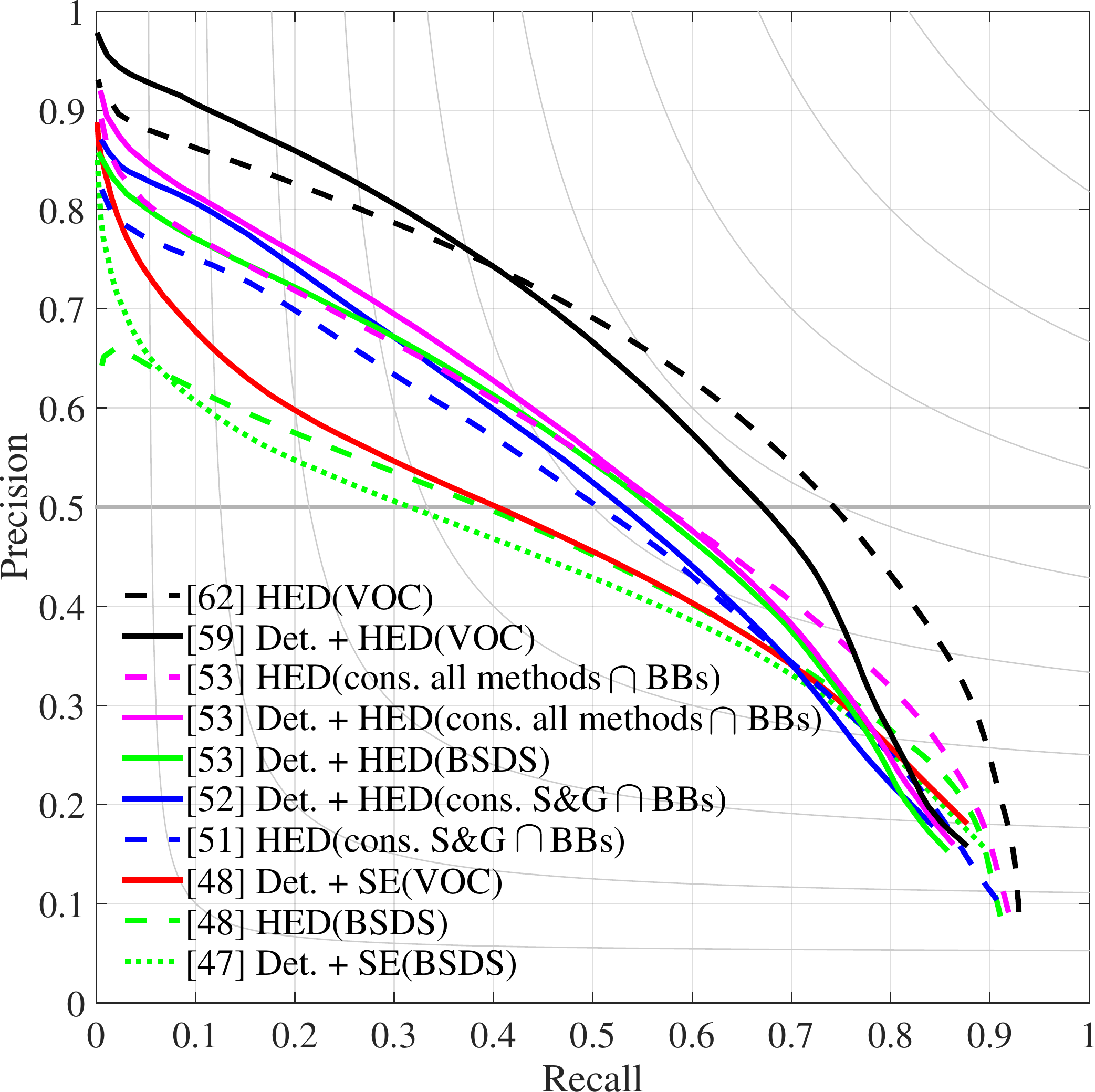}
\par\end{centering}

\protect\caption{\label{fig:VOC-HED-results}VOC12 HED results. $\left(\cdot\right)$
denotes the data used for training. Continuous/dashed line indicates
models using/not using a detector at test time. Legend indicates AP
numbers.}
\vspace{-1em}
\end{figure}

This section analyses the performance of the HED \cite{Xie2015Iccv}
trained with the weakly supervised variants proposed in Section \ref{sec:Weak-groundtruth}.
We use our HED re-implementation of HED which is on par performance
with the original (see Figure \ref{fig:BSD500-results}). We use the
same evaluation setup as in the previous section. Figure \ref{fig:VOC-HED-results}
and Table \ref{tab:VOC-HED-results} show the results.

\paragraph{$\mbox{HED}\left(\cdot\right)$}

The $\textrm{HED(VOC)}$ model outperforms the $\textrm{SE(VOC)}$
model by a large margin. We observe in the test images that HED manages
to suppress well the internal object boundaries, while SE fails to
do so due to its more local nature. 

\begin{table}
\setlength{\tabcolsep}{2.5pt} 
\renewcommand{\arraystretch}{0.85}

\centering{}\vspace{-0.5em}
\begin{tabular}{ccc|ccc|ccc}
\multirow{2}{*}{{\footnotesize{}Family}} & \multirow{2}{*}{{\footnotesize{}Method}} & \multirow{2}{*}{{\footnotesize{}Data}} & \multicolumn{3}{c|}{{\footnotesize{}Without BBs}} & \multicolumn{3}{c}{{\footnotesize{}With BBs}}\tabularnewline
 &  &  & {\footnotesize{}F} & {\footnotesize{}AP} & {\footnotesize{}$\textrm{\ensuremath{\Delta}}$AP} & {\footnotesize{}F} & {\footnotesize{}AP} & {\footnotesize{}$\Delta$AP}\tabularnewline
\hline 
\hline 
\multirow{2}{*}{{\footnotesize{}GT}} & {\footnotesize{}SE} & \multirow{2}{*}{{\footnotesize{}VOC}} & {\footnotesize{}\uline{43}} & {\footnotesize{}\uline{35}} & {\footnotesize{}-} & {\footnotesize{}\uline{48}} & {\footnotesize{}\uline{41}} & {\footnotesize{}-}\tabularnewline
 & {\footnotesize{}HED} &  & {\footnotesize{}62} & {\footnotesize{}61} & {\footnotesize{}26} & {\footnotesize{}59} & {\footnotesize{}58} & {\footnotesize{}17}\tabularnewline
\hline 
\multirow{2}{*}{{\footnotesize{}}%
\begin{tabular}{c}
{\footnotesize{}Other}\tabularnewline
{\footnotesize{}GT}\tabularnewline
\end{tabular}} & \multirow{2}{*}{{\footnotesize{}HED}} & {\footnotesize{}BSDS} & {\footnotesize{}48} & {\footnotesize{}41} & {\footnotesize{}6} & {\footnotesize{}53} & {\footnotesize{}48} & {\footnotesize{}7}\tabularnewline
 &  & {\footnotesize{}COCO} & {\footnotesize{}59} & {\footnotesize{}60} & {\footnotesize{}25} & {\footnotesize{}56} & {\footnotesize{}55} & {\footnotesize{}14}\tabularnewline
\hline 
\multirow{6}{*}{{\footnotesize{}}%
\begin{tabular}{c}
{\footnotesize{}Weakly}\tabularnewline
{\footnotesize{}super-}\tabularnewline
{\footnotesize{}vised}\tabularnewline
\end{tabular}} & {\footnotesize{}SE} & {\footnotesize{}$\mbox{MCG}\cap\mbox{BBs}$} & {\footnotesize{}43} & {\footnotesize{}34} & {\footnotesize{}-1} & {\footnotesize{}47} & {\footnotesize{}39} & {\footnotesize{}-2}\tabularnewline
\cline{2-9} 
 & \multirow{5}{*}{{\footnotesize{}HED}} & {\footnotesize{}$\textrm{SE(SeSe}\cap\mbox{BBs})$} & {\footnotesize{}45} & {\footnotesize{}37} & {\footnotesize{}3} & {\footnotesize{}49} & {\footnotesize{}40} & {\footnotesize{}-1}\tabularnewline
 &  & {\footnotesize{}$\mbox{MCG}\cap\mbox{BBs}$} & {\footnotesize{}50} & {\footnotesize{}44} & {\footnotesize{}9} & {\footnotesize{}48} & {\footnotesize{}42} & {\footnotesize{}1}\tabularnewline
 &  & {\footnotesize{}$\mbox{cons. }\mbox{S\&G}\cap\mbox{BBs}$} & \textbf{\textcolor{black}{\footnotesize{}51}} & \textbf{\textcolor{black}{\footnotesize{}46}} & \textbf{\textcolor{black}{\footnotesize{}+11}} & \textbf{\textcolor{black}{\footnotesize{}52}} & \textbf{\textcolor{black}{\footnotesize{}47}} & \textbf{\textcolor{black}{\footnotesize{}+8}}\tabularnewline
 &  & {\footnotesize{}cons. $\mbox{MCG}\cap\mbox{BBs}$} & {\footnotesize{}53} & {\footnotesize{}50} & {\footnotesize{}15} & {\footnotesize{}52} & {\footnotesize{}49} & {\footnotesize{}8}\tabularnewline
 &  & {\footnotesize{}\hspace*{-1.5em}cons. all methods$\cap\mbox{BBs}$} & \textbf{\textcolor{black}{\footnotesize{}53}} & \textbf{\textcolor{black}{\footnotesize{}50}} & \textbf{\textcolor{black}{\footnotesize{}+15}} & \textbf{\textcolor{black}{\footnotesize{}53}} & \textbf{\textcolor{black}{\footnotesize{}50}} & \textbf{\textcolor{black}{\footnotesize{}+9}}\tabularnewline
\end{tabular}\protect\caption{\label{tab:VOC-HED-results}VOC results for HED models, see Figure
\ref{fig:VOC-HED-results}. Underline indicates ground truth baselines,
and bold are our best weakly supervised results.}
\vspace{-0.5em}
\end{table}
Even though trained on the generic boundaries HED(BSDS) achieves high
performance on the object boundary detection task. HED(BSDS) is trained
on the ``consensus'' annotations and they are closer to object-like
boundaries as the fraction of annotators agreeing on the presence
of external object boundaries is much higher than for non-object or
internal object boundaries.

For training HED, in contrast to SE model, we do not need closed contours
and can use the consensus between different weak annotation variants.
This results in better performance. Using the consensus between boundaries
of MCG proposals $\textrm{HED}(\textrm{cons.}\mbox{ MCG}\cap\mbox{BBs})$
improves AP by $6\%$ compared to using the union of object proposals
$\textrm{HED}(\mbox{MCG}\cap\mbox{BBs})$ (see Table \ref{tab:VOC-HED-results})
. 

The HED models trained with weak annotations outperform the fully
supervised $\textrm{SE(VOC})$ and do not reach the performance of
$\textrm{HED(VOC)}$. As has been shown in Section \ref{sec:Robustness}
the HED detector is less robust to noise than SE.

\paragraph{$\mbox{Det.}\negmedspace+\negmedspace\mbox{HED}\left(\cdot\right)$}

Combining an object detector with $\textrm{HED(VOC)}$ (see $\mbox{Det.}\negmedspace+\negmedspace\mbox{HED}\left(\textrm{VOC}\right)$
in Figure \ref{fig:VOC-HED-results}) is not beneficial to the performance
as the HED detector already has notion of objects and their location
due to pixel-to-pixel end-to-end learning of the network. 

For HED models trained with the weakly supervised variants, employing
an object detector at test time brings only a slight improvement of
the performance in the high precision area . The reason for this is
that we already use information from the bounding box detector to
generate the annotation and the convnet method is able to learn it
during training. 

$\mbox{Det.}\negmedspace+\negmedspace\mbox{HED}\left(\mbox{MCG}\cap\mbox{BBs}\right)$
outperforms $\mbox{Det.}\negmedspace+\negmedspace\mbox{HED}\left(\mbox{BSDS}\right)$
(see Table \ref{tab:VOC-HED-results}). Note that the HED trained
with the proposed annotations, generated without using boundary ground
truth, performs on par with the HED model trained on generic boundaries
($\mbox{Det.}\negmedspace+\negmedspace\mbox{HED}\left(\mbox{\mbox{cons. }\mbox{S\&G}\ensuremath{\cap}\mbox{BBs}}\right)$
and $\mbox{Det.}\negmedspace+\negmedspace\mbox{HED}\left(\mbox{BSDS}\right)$in
Figure \ref{fig:VOC-HED-results}).

\begin{figure*}
\begingroup{\arrayrulecolor{lightgray}
\setlength{\tabcolsep}{0pt} 
\renewcommand{\arraystretch}{0}

\begin{tabular}[b]{|c|c|c|c|c|c|c|}
\hline 
\includegraphics[width=0.14\textwidth,height=0.06\textheight]{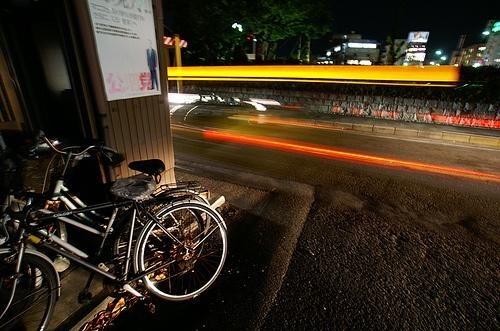} & \includegraphics[width=0.14\textwidth,height=0.06\textheight]{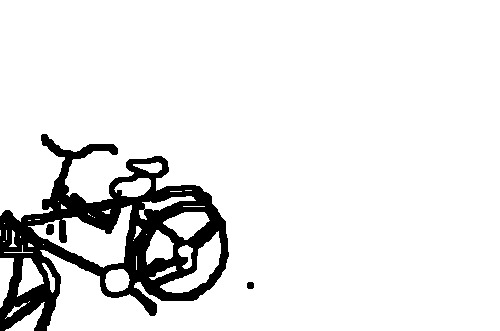} & \includegraphics[width=0.14\textwidth,height=0.06\textheight]{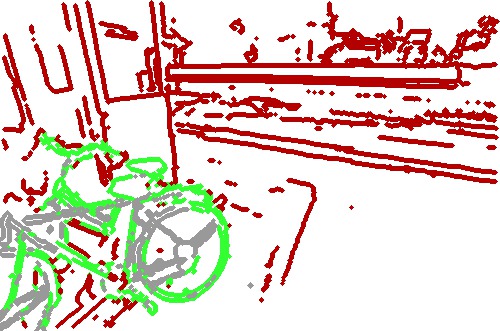} & \includegraphics[width=0.14\textwidth,height=0.06\textheight]{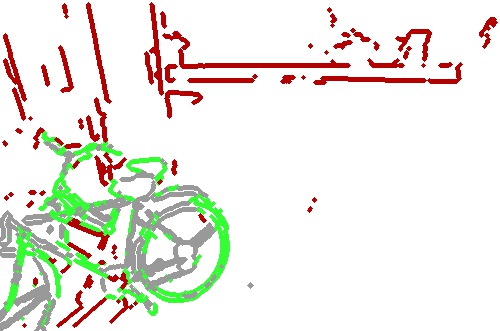} & \includegraphics[width=0.14\textwidth,height=0.06\textheight]{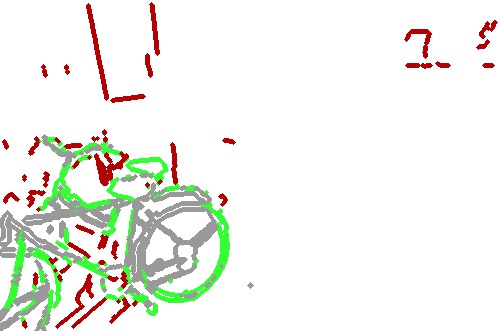} & \includegraphics[width=0.14\textwidth,height=0.06\textheight]{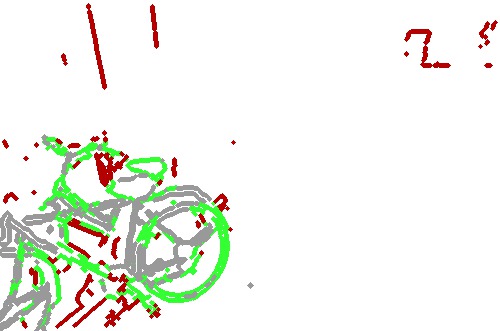} & \includegraphics[width=0.16\textwidth,height=0.06\textheight]{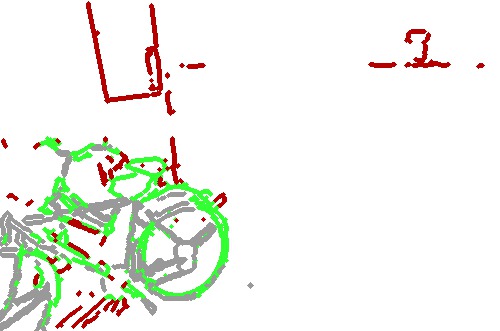}\tabularnewline
\hline 
\includegraphics[width=0.14\textwidth,height=0.065\textheight]{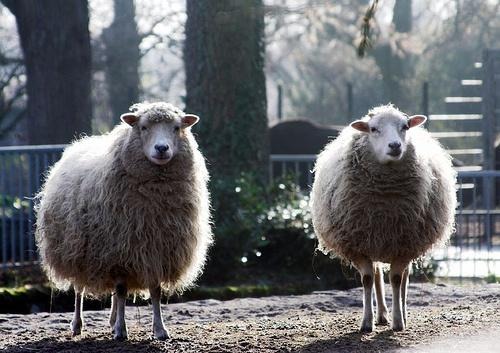} & \includegraphics[width=0.14\textwidth,height=0.065\textheight]{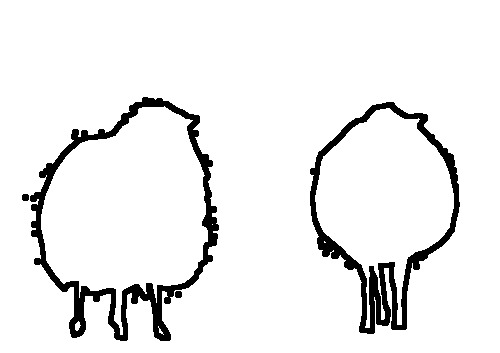} & \includegraphics[width=0.14\textwidth,height=0.065\textheight]{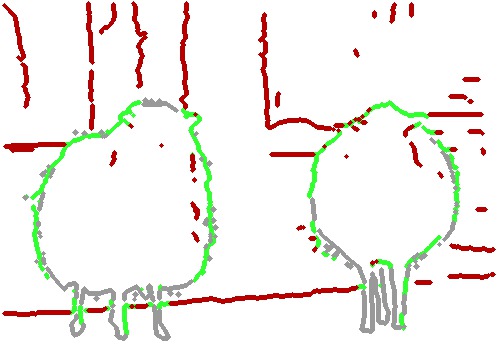} & \includegraphics[width=0.14\textwidth,height=0.065\textheight]{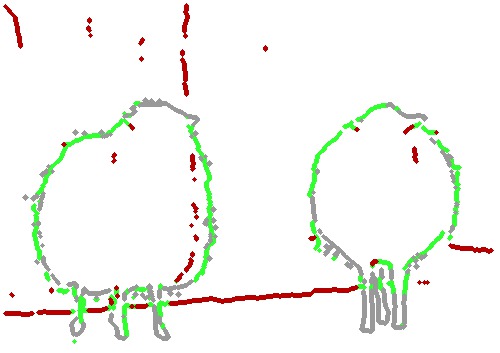} & \includegraphics[width=0.14\textwidth,height=0.065\textheight]{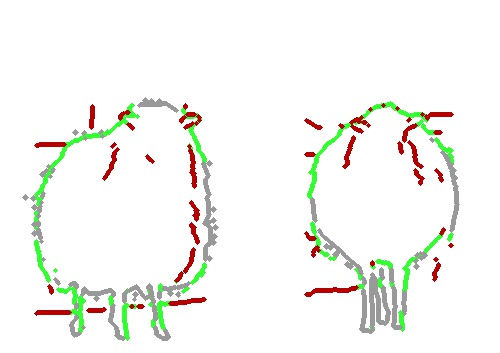} & \includegraphics[width=0.14\textwidth,height=0.065\textheight]{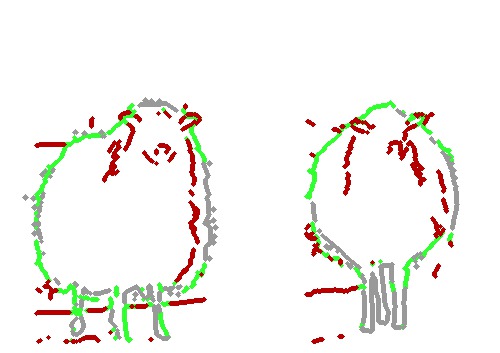} & \includegraphics[width=0.14\textwidth,height=0.065\textheight]{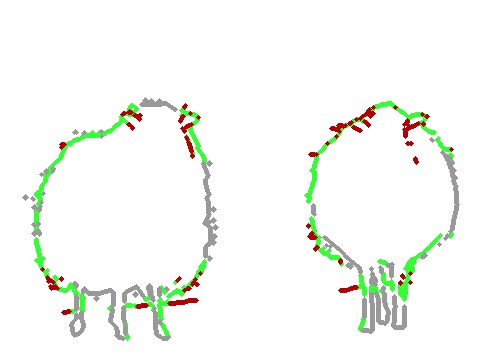}\tabularnewline
\hline 
\includegraphics[width=0.14\textwidth,height=0.06\textheight]{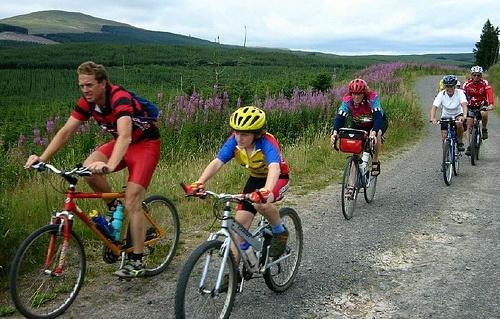} & \includegraphics[width=0.14\textwidth,height=0.06\textheight]{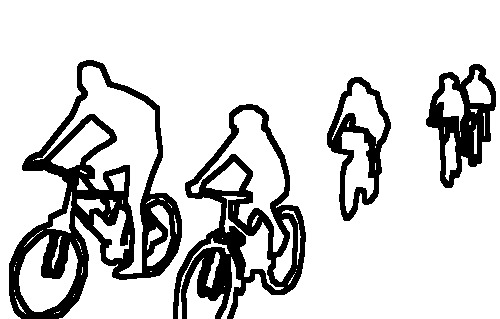} & \includegraphics[width=0.14\textwidth,height=0.06\textheight]{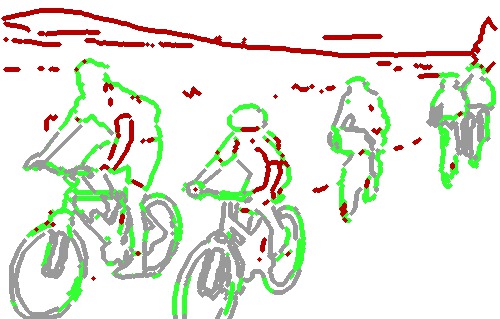} & \includegraphics[width=0.14\textwidth,height=0.06\textheight]{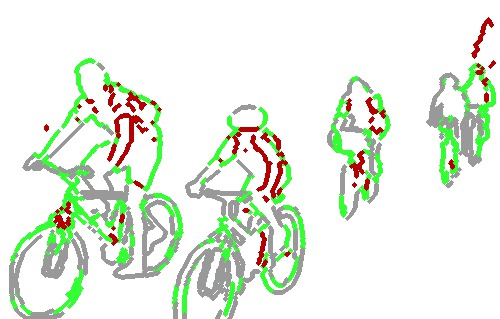} & \includegraphics[width=0.14\textwidth,height=0.06\textheight]{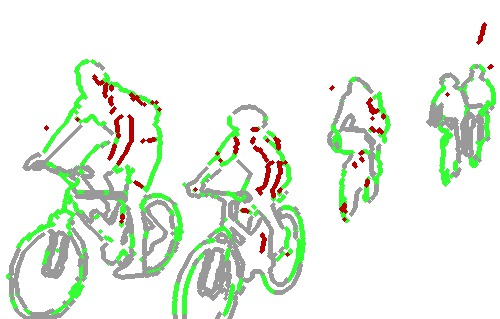} & \includegraphics[width=0.14\textwidth,height=0.06\textheight]{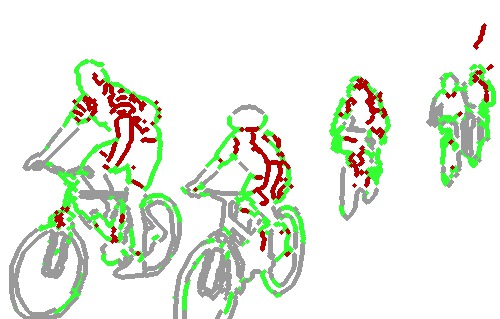} & \includegraphics[width=0.16\textwidth,height=0.06\textheight]{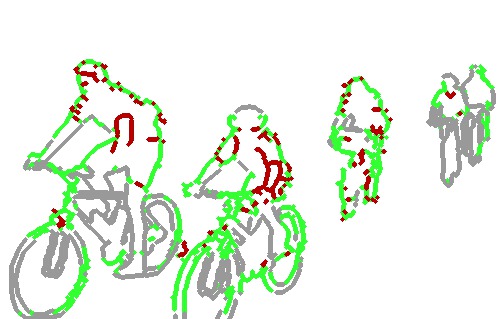}\tabularnewline
\hline 
\multicolumn{1}{c}{Image} & \multicolumn{1}{c}{Ground truth} & \multicolumn{1}{c}{SE(BSDS)} & \multicolumn{1}{c}{SB(VOC)} & \multicolumn{1}{c}{$\mbox{Det.}\negmedspace+\negmedspace\mbox{SE}\left(\mbox{VOC}\right)$} & \multicolumn{1}{c}{$\mbox{Det.}\negmedspace+\negmedspace\mbox{SE}\left(\mbox{weak}\right)$} & \multicolumn{1}{c}{$\mbox{Det.}\negthickspace+\negthickspace\mbox{HED}\left(\textrm{weak}\right)$}\tabularnewline
\end{tabular}

\protect\caption{\label{fig:Qualitative-results}Qualitative results on VOC. $\left(\cdot\right)$
denotes the data used for training. Red/green indicate false/true
positive pixels, grey is missing recall. All methods are shown at
$50\%$ recall. $\mbox{Det.}\negmedspace+\negmedspace\mbox{SE}\left(\textrm{weak}\right)$
refers to the model $\mbox{Det.}\negmedspace+\negmedspace\mbox{SE}\left(\mbox{SeSe}_{+}\cap\mbox{BBs}\right)$
$\mbox{Det.}\negmedspace+\negmedspace\mbox{HED}\left(\textrm{weak}\right)$
refers to $\mbox{Det.}\negmedspace+\negmedspace\mbox{HED}\left(\mbox{cons. }\mbox{S\&G}\cap\mbox{BBs}\right)$.
Object-specific boundaries differ from generic boundaries (such as
the ones detected by SE(BSDS)). By using an object detector we can
suppress non-object boundaries and focus boundary detection on the
classes of interest. The proposed weakly supervised techniques allow
to achieve high quality boundary estimates that are similar to the
ones obtained by fully supervised methods. }
\vspace{-1em}
}\endgroup
\arrayrulecolor{black}
\end{figure*}

The qualitative results are presented in Figure \ref{fig:Qualitative-results}
and support the quantitative evaluation.

\paragraph{Conclusion}

Similar to other computer vision tasks deep convnet methods show superior
performance. Due to the pixel-to-pixel training and global view of
the image the convnet models have a notion of object and its location
which allows to omit the use of the detector at test time. With our
weakly supervised boundary annotations we can gain fair performance
without using any instance-wise object boundary or generic boundary
annotations and leave out object detection at test time by feeding
object bounding box information during training.

\section{\label{sec:COCO-boundaries}COCO boundary detection results}

\begin{table}
\begin{centering}
\setlength{\tabcolsep}{1pt} 
\renewcommand{\arraystretch}{0.85}%
\begin{tabular}{ccc|ccc|ccc}
\multirow{2}{*}{{\footnotesize{}Method}} & \multirow{2}{*}{{\footnotesize{}Family}} & \multirow{2}{*}{{\footnotesize{}Data}} & \multicolumn{3}{c|}{{\footnotesize{}Without BBs}} & \multicolumn{3}{c}{{\footnotesize{}With BBs}}\tabularnewline
 &  &  & {\footnotesize{}F} & {\footnotesize{}AP} & {\footnotesize{}$\textrm{\ensuremath{\Delta}}$AP} & {\footnotesize{}F} & {\footnotesize{}AP} & {\footnotesize{}$\Delta$AP}\tabularnewline
\hline 
\hline 
\multirow{4}{*}{{\footnotesize{}SE}} & {\footnotesize{}GT} & \multirow{1}{*}{{\footnotesize{}COCO}} & {\footnotesize{}\uline{40}} & {\footnotesize{}\uline{32}} & {\footnotesize{}-} & {\footnotesize{}\uline{45}} & {\footnotesize{}\uline{37}} & {\footnotesize{}-}\tabularnewline
 & {\footnotesize{}Other GT} & \multirow{1}{*}{{\footnotesize{}BSDS}} & {\footnotesize{}34} & {\footnotesize{}23} & {\footnotesize{}-9} & {\footnotesize{}43} & {\footnotesize{}33} & {\footnotesize{}-4}\tabularnewline
\cline{3-9} 
 & \multirow{2}{*}{{\footnotesize{}}%
\begin{tabular}{c}
{\footnotesize{}Weakly}\tabularnewline
{\footnotesize{}supervised}\tabularnewline
\end{tabular}} & {\footnotesize{}$\mbox{SeSe}_{+}\cap\mbox{BBs}$} & \textbf{\footnotesize{}40} & \textbf{\footnotesize{}31} & \textbf{\footnotesize{}-1} & \textbf{\footnotesize{}44} & \textbf{\footnotesize{}35} & \textbf{\footnotesize{}-2}\tabularnewline
 &  & {\footnotesize{}$\mbox{MCG}_{+}\cap\mbox{BBs}$} & {\footnotesize{}39} & {\footnotesize{}30} & {\footnotesize{}-2} & {\footnotesize{}44} & {\footnotesize{}35} & {\footnotesize{}-2}\tabularnewline
\cline{2-9} 
\multirow{4}{*}{{\footnotesize{}HED}} & {\footnotesize{}GT} & {\footnotesize{}COCO} & {\footnotesize{}60} & {\footnotesize{}59} & {\footnotesize{}27} & {\footnotesize{}56} & {\footnotesize{}55} & {\footnotesize{}18}\tabularnewline
 & {\footnotesize{}Other GT} & {\footnotesize{}BSDS} & {\footnotesize{}44} & {\footnotesize{}34} & {\footnotesize{}2} & {\footnotesize{}49} & {\footnotesize{}42} & {\footnotesize{}5}\tabularnewline
\cline{3-9} 
 & \multirow{2}{*}{{\footnotesize{}}%
\begin{tabular}{c}
{\footnotesize{}Weakly}\tabularnewline
{\footnotesize{}supervised}\tabularnewline
\end{tabular}} & {\footnotesize{}cons. $\mbox{S\&G}$$\cap$BBs} & {\footnotesize{}47} & {\footnotesize{}39} & {\footnotesize{}7} & {\footnotesize{}48} & {\footnotesize{}42} & {\footnotesize{}5}\tabularnewline
 &  & {\footnotesize{}cons. all methods$\cap\mbox{BBs}$} & \textbf{\footnotesize{}49} & \textbf{\footnotesize{}43} & \textbf{\footnotesize{}+11} & \textbf{\footnotesize{}50} & \textbf{\footnotesize{}44} & \textbf{\footnotesize{}+7}\tabularnewline
\end{tabular}
\par\end{centering}

\protect\caption{\label{tab:COCO-results}COCO results, curves in supplementary material.
Underline indicates ground truth baselines.}
\vspace{-1em}
\end{table}
Additionally we show the generalization of the proposed weakly supervised
variants for object boundary detection on the COCO dataset. We use
the same evaluation protocol as for VOC. For weakly supervised cases
the results are shown with the models trained on VOC, without re-training
on COCO. 

The results are summarized in Table \ref{tab:COCO-results}. On the
COCO benchmark for both SE and HED the models trained on the proposed
weak annotations perform as well as the fully supervised SE models.
Similar to the VOC benchmark the HED model trained on ground truth
shows superior performance.
\begin{table}
\begin{centering}
\setlength{\tabcolsep}{2.5pt} 
\renewcommand{\arraystretch}{0.85}{\small{}}%
\begin{tabular}{ccllc}
 & {\footnotesize{}Family } & {\footnotesize{}Method} & {\footnotesize{}mF} & {\footnotesize{}mAP}\tabularnewline
\hline 
\hline 
\multirow{1}{*}{{\footnotesize{}Other}} & \multirow{1}{*}{{\footnotesize{}GT}} & {\footnotesize{}Hariharan et al. \cite{Hariharan2011Iccv}} & {\footnotesize{}28} & {\footnotesize{}21}\tabularnewline
\hline 
\multirow{9}{*}{{\footnotesize{}SE}} & \multirow{3}{*}{{\footnotesize{}GT}} & {\footnotesize{}SB(SBD) orig. \cite{Uijlings2015Cvpr}} & {\footnotesize{}39} & {\footnotesize{}32}\tabularnewline
 &  & {\footnotesize{}SB(SBD)} & {\footnotesize{}43} & {\footnotesize{}37}\tabularnewline
 &  & {\footnotesize{}$\mbox{Det.}\negmedspace+\negmedspace\mbox{SE}\left(\mbox{SBD}\right)$} & {\footnotesize{}\uline{51}} & {\footnotesize{}\uline{45}}\tabularnewline
\cline{3-5} 
 & \multirow{2}{*}{{\footnotesize{}}%
\begin{tabular}{c}
{\footnotesize{}Other}\tabularnewline
{\footnotesize{}GT}\tabularnewline
\end{tabular}} & {\footnotesize{}$\mbox{Det.}\negmedspace+\negmedspace\mbox{SE}\left(\mbox{BSDS}\right)$} & {\footnotesize{}51} & {\footnotesize{}44}\tabularnewline
 &  & {\footnotesize{}$\mbox{Det.}\negmedspace+\negmedspace\mbox{MCG}\left(\mbox{BSDS}\right)$} & {\footnotesize{}50} & {\footnotesize{}42}\tabularnewline
\cline{3-5} 
 & \multirow{4}{*}{{\footnotesize{}}%
\begin{tabular}{c}
{\footnotesize{}Weakly}\tabularnewline
{\footnotesize{}super-}\tabularnewline
{\footnotesize{}vised}\tabularnewline
\end{tabular}} & {\footnotesize{}SB($\mbox{SeSe}\cap\mbox{BBs}$)} & {\footnotesize{}40} & {\footnotesize{}34}\tabularnewline
 &  & {\footnotesize{}$\mbox{SB}\left(\mbox{MCG}\cap\mbox{BBs}\right)$} & {\footnotesize{}42} & {\footnotesize{}35}\tabularnewline
 &  & {\footnotesize{}$\mbox{Det.}\negmedspace+\negmedspace\mbox{SE}\left(\mbox{SeSe}\cap\mbox{BBs}\right)$ } & {\footnotesize{}48} & {\footnotesize{}42}\tabularnewline
 &  & {\footnotesize{}$\mbox{Det.}\negmedspace+\negmedspace\mbox{SE}\left(\mbox{MCG}\cap\mbox{BBs}\right)$ } & \textbf{\footnotesize{}51} & \textbf{\footnotesize{}45}\tabularnewline
\hline 
\multirow{8}{*}{{\footnotesize{}HED}} & \multirow{2}{*}{{\footnotesize{}GT}} & {\footnotesize{}$\mbox{HED}\left(\mbox{SBD}\right)$} & {\footnotesize{}44} & {\footnotesize{}41}\tabularnewline
 &  & {\footnotesize{}$\mbox{Det.}\negmedspace+\negmedspace\mbox{HED}\left(\mbox{SBD}\right)$} & {\footnotesize{}\uline{49}} & {\footnotesize{}\uline{45}}\tabularnewline
\cline{3-5} 
 & \multirow{2}{*}{{\footnotesize{}}%
\begin{tabular}{c}
{\footnotesize{}Other}\tabularnewline
{\footnotesize{}GT}\tabularnewline
\end{tabular}} & {\footnotesize{}HED(BSDS)} & {\footnotesize{}38} & {\footnotesize{}32}\tabularnewline
 &  & {\footnotesize{}$\mbox{Det.}\negmedspace+\negmedspace\mbox{HED}\left(\mbox{BSDS}\right)$} & {\footnotesize{}49} & {\footnotesize{}44}\tabularnewline
\cline{3-5} 
 & \multirow{4}{*}{{\footnotesize{}}%
\begin{tabular}{c}
{\footnotesize{}Weakly}\tabularnewline
{\footnotesize{}super-}\tabularnewline
{\footnotesize{}vised}\tabularnewline
\end{tabular}} & {\footnotesize{}HED(cons. $\mbox{MCG}\cap\mbox{BBs}$)} & {\footnotesize{}41} & {\footnotesize{}37}\tabularnewline
 &  & {\footnotesize{}$\mbox{HED}\left(\mbox{cons. }\mbox{S\&G}\cap\mbox{BBs}\right)$} & {\footnotesize{}44} & {\footnotesize{}39}\tabularnewline
 &  & {\footnotesize{}$\mbox{Det.}\negmedspace+\negmedspace\mbox{HED}\left(\mbox{cons. }\mbox{MCG}\cap\mbox{BBs}\right)$} & {\footnotesize{}48} & {\footnotesize{}44}\tabularnewline
 &  & {\footnotesize{}$\mbox{Det.}\negmedspace+\negmedspace\mbox{HED}\left(\mbox{cons. }\mbox{S\&G}\cap\mbox{BBs}\right)$} & \textbf{\footnotesize{}52} & \textbf{\footnotesize{}47}\tabularnewline
\end{tabular}
\par\end{centering}{\small \par}

\protect\caption{\label{tab:SBD-results}SBD results. Results are mean F(ODS)/AP across
all 20 categories. $\left(\cdot\right)$ denotes the data used for
training. See also Figure \ref{fig:SBD-results}. Underline indicates
ground truth baselines, and bold are our best weakly supervised results.}
\vspace{-0.5em}
\end{table}
\begin{figure}
\noindent \begin{centering}
\includegraphics[bb=0bp 0bp 604bp 262bp,width=1\columnwidth]{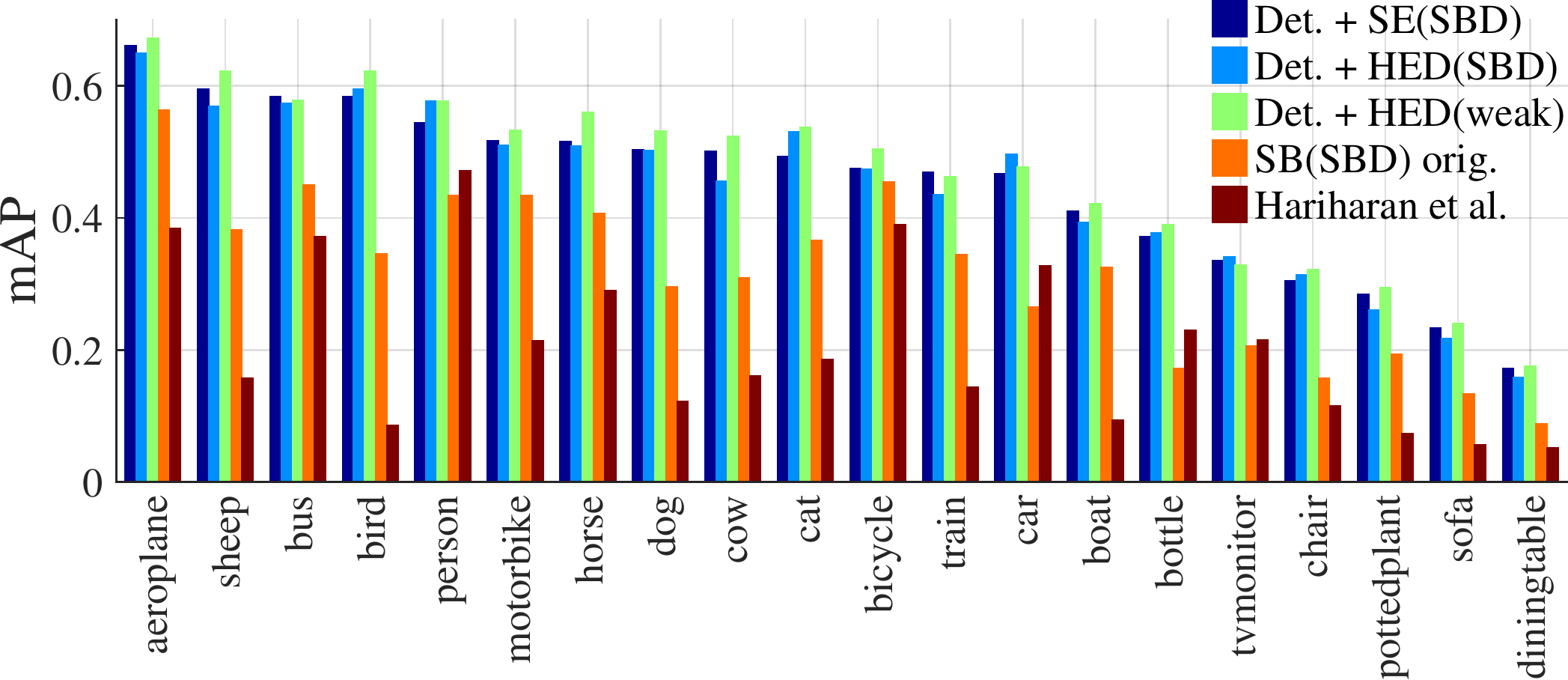}
\par\end{centering}

\protect\caption{\label{fig:SBD-results}SBD results per class. $\left(\cdot\right)$
denotes the data used for training. $\mbox{Det.}\negmedspace+\negmedspace\mbox{HED}\left(\textrm{weak}\right)$
refers to the model $\mbox{Det.}\negmedspace+\negmedspace\mbox{HED}\left(\mbox{cons. }\mbox{S\&G}\cap\mbox{BBs}\right)$.}
\vspace{-0.5em}
\end{figure}

\section{\label{sec:SBD-boundaries}SBD boundary detection results}

In this section we analyse the performance of the proposed weakly
supervised boundary variants trained with SE and HED on the SBD dataset
\cite{Hariharan2011Iccv}. In contrast to the VOC benchmark we move
from object boundaries to class specific object boundaries. We are
interested in external boundaries of all annotated objects of the
specific semantic class and all internal boundaries are ignored during
evaluation following the benchmark \cite{Hariharan2011Iccv}. The
results are presented in Figure \ref{fig:SBD-results} and in Table
\ref{tab:SBD-results}.

\paragraph{Fully supervised}

Applying SE model plus object detection at test time outperforms the
class specific situational boundary detector (for both \cite{Uijlings2015Cvpr}
and our re-implementation) as well as the Inverse Detectors \cite{Hariharan2011Iccv}.
The model trained with SE on ground truth performs as well as the
HED detector. Both of the models are good at detecting external object
boundaries; however SE, being a more local, triggers more on internal
boundaries than HED. In the VOC evaluation detecting internal object
boundaries is penalized, while in SBD these are ignored. This explains
the small gap in the performance between SE and HED on this benchmark.

\paragraph{Weakly supervised}

The models trained with the proposed weakly-supervised boundary variants
perform on par with the fully supervised detectors, while only using
bounding boxes or generic boundary annotations. We show in Table \ref{tab:SBD-results}
the top result with the Det. + HED(cons. S\&G$\cap$BBs) model, achieving
the state-of-the-art performance on the SBD benchmark. As Figure \ref{fig:SBD-results}
shows our weakly supervised approach considerably outperforms \cite{Uijlings2015Cvpr,Hariharan2011Iccv}
on all 20 classes.

\section*{\label{sec:Conclusion}Conclusion}

The presented experiments show that high quality object boundaries
can be achieved using only detection bounding box annotations. With
these alone, our proposed weak-supervision techniques already improve
over previously reported fully supervised results for object-specific
boundaries. When using generic boundary or ground truth annotations,
we achieve the top performance on the object boundary detection task,
outperforming previously reported results by a large margin.

To facilitate future research all the resources of this project -
 source code, trained models and results - will be made publicly available.

\section*{Acknowledgements}

We would like to thank Jan Hosang for help with Fast R-CNN training;
Seong Joon Oh and Bojan Pepik for valuable discussions and feedback.

\bibliographystyle{plain}
\bibliography{2016_cvpr_weakly_supervised_boundaries}
\cleardoublepage{}

\appendix

\part*{Supplementary material}

\global\long\def\FH{\mbox{F\ensuremath{\mathfrak{\&}}H}}
\makeatletter 
\renewcommand{\paragraph}{%
\@startsection{paragraph}{4}%
{\z@}{1.0ex \@plus 1ex \@minus .2ex}{-1em}%
{\normalfont \normalsize \bfseries}%
}
\makeatother

\section{\label{sec:Content}Content}

This supplementary material provides both additional quantitative
and qualitative results:
\begin{itemize}
\item Section \ref{sec:Noisy-boundaries-estimates} (Figure \ref{fig:Noisy-boundaries-1})
shows visualization of the different variants of the proposed weakly
supervised boundary annotations. 
\item Section \ref{sec:Detailed-results-VOC} provides additional results
for VOC (Figures \ref{fig:GT-bbs} - \ref{fig:VOC-results}). Qualitative
results of boundary detection for VOC can be found in Section \ref{sec:VOC-boundaries-examples}
(Figure \ref{fig:Qualitative-results-voc}). 
\item Detailed results for COCO are shown in Section \ref{sec:Detailed-results-COCO}
(Figure \ref{fig:COCO-results}). Visualization of boundary detection
for COCO can be found in Section \ref{sec:Detailed-results-COCO}
(Figure \ref{fig:Qualitative-results-coco}). 
\item Section \ref{sec:Detailed-results-SBD} reports SBD results per class
(Figure \ref{fig:SBD-boundary-PR}). Boundary detection examples for
SBD are shown in Section \ref{sec:SBD-boundaries-examples} (Figure
\ref{fig:Qualitative-results-sbd}).
\end{itemize}

\section{\label{sec:Noisy-boundaries-estimates}Weakly supervised boundary
annotations}

In this work we propose to train boundary detectors using weakly supervised
annotations. We propose and evaluate multiple strategies to generate
annotations fusing different sources, such as unsupervised image segmentation
\cite{Felzenszwalb2004Ijcv}, object proposal methods \cite{Uijlings2013Ijcv,PontTuset2015ArxivMcg},
and object detectors \cite{Girshick2015IccvFastRCNN,Ren2015NipsFasterRCNN}
(trained on bounding boxes). Figure \ref{fig:Noisy-boundaries-1}
illustrates the examples of the proposed weakly supervised boundary
annotations, these extend the example in Figure 4 of the main paper.
See Section 5 of the main paper for more details.

\section{\label{sec:Detailed-results-VOC}Detailed results for VOC}

We provide here some quantitative results mentioned in Section 5,
and in Table 2 of the main paper.
\begin{figure}
\noindent \begin{centering}
\includegraphics[bb=0bp 90bp 612bp 710bp,clip,width=0.8\columnwidth]{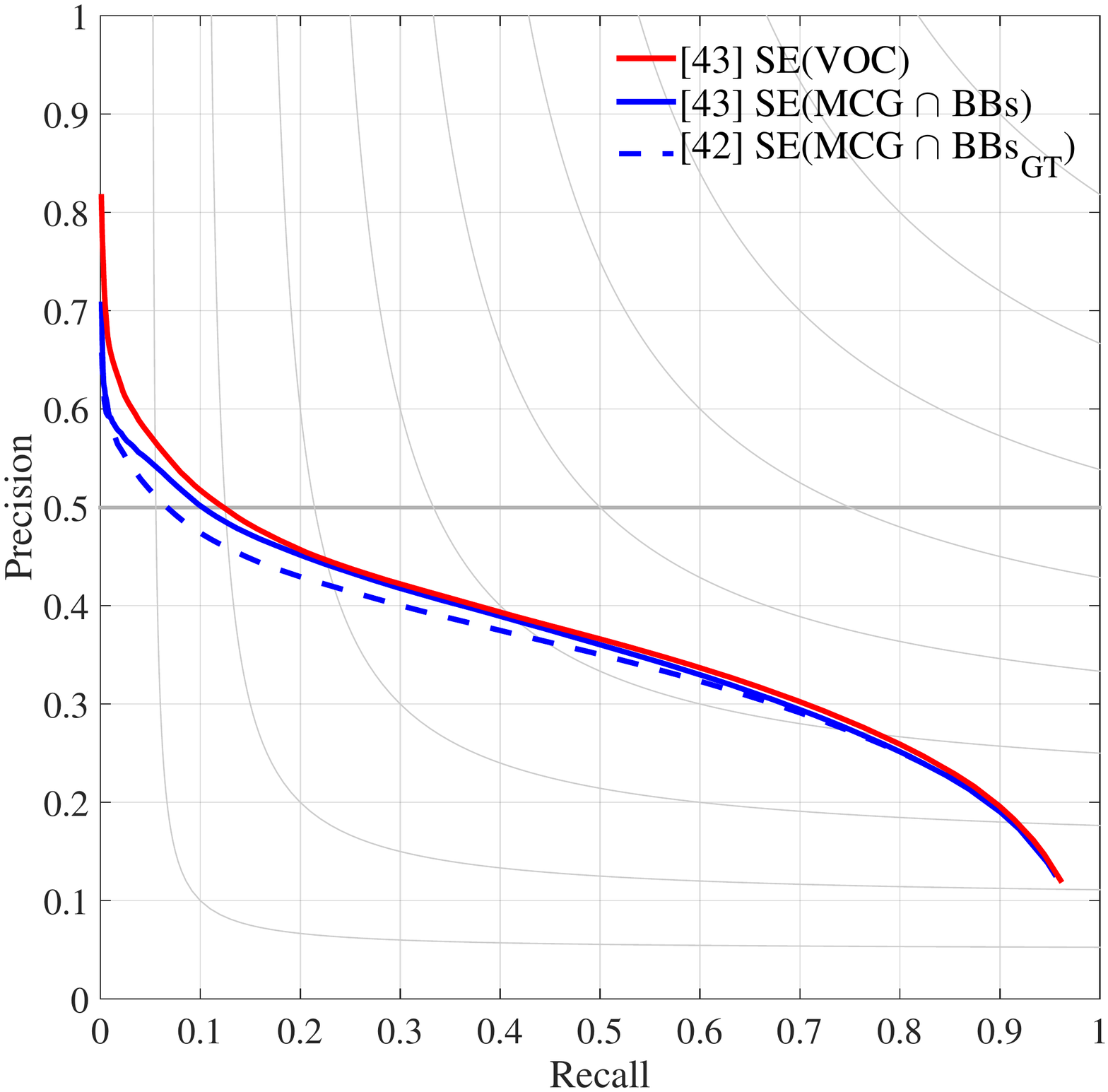}
\par\end{centering}

\protect\caption{\label{fig:GT-bbs}VOC results: Detection BBs versus GT BBs. $\left(\cdot\right)$
denotes the data used for training. Legend indicates AP numbers. $\textrm{BBs}_{\textrm{GT}}$
denotes bounding boxes annotated by humans.}

\noindent \begin{centering}
\includegraphics[bb=0bp 90bp 612bp 710bp,clip,width=0.8\columnwidth]{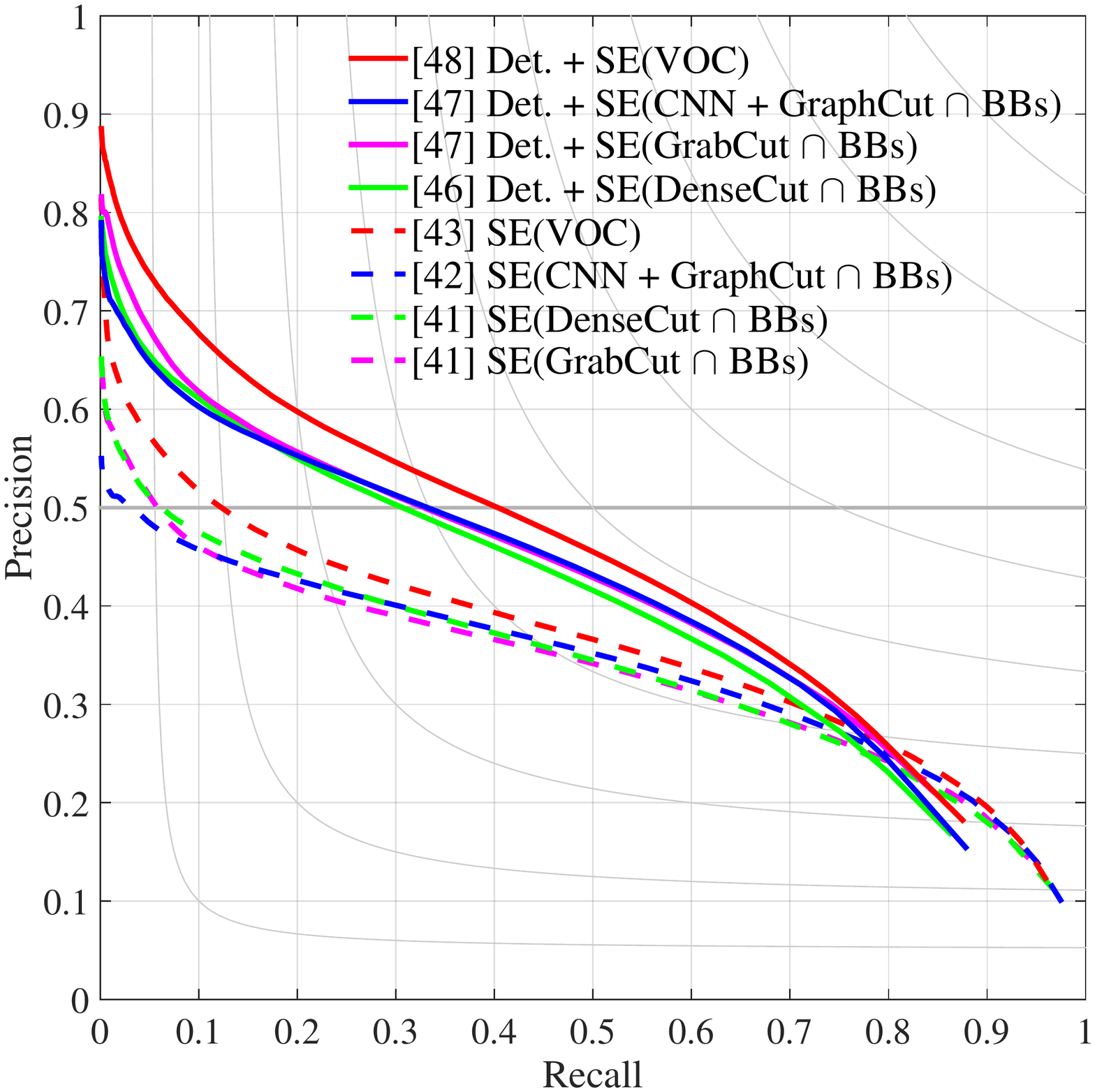}
\par\end{centering}

\protect\caption{\label{fig:caffecut}VOC results: GrabCut versus DenseCut and CNN+GraphCut.
$\left(\cdot\right)$ denotes the data used for training. Continuous/dashed
line indicates models using/not using a detector at test time. Legend
indicates AP numbers. DenseCut and CNN+GraphCut perform on par with
GrabCut.}
\end{figure}
\begin{figure}
\noindent \begin{centering}
\includegraphics[bb=0bp 90bp 612bp 710bp,clip,width=0.8\columnwidth]{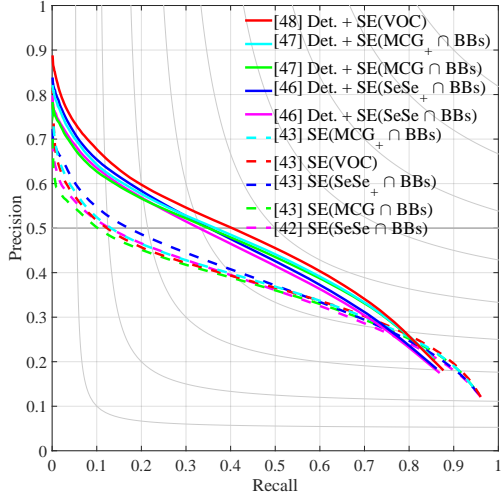}
\par\end{centering}

\protect\caption{\label{fig:VOC-results}VOC results using additional $\mbox{VOC}_{+}$
images. $\left(\cdot\right)$ denotes the data used for training.
Continuous/dashed line indicates models using/not using a detector
at test time. Legend indicates AP numbers. }
\end{figure}

\paragraph{Detection BBs versus GT BBs}

To generate weakly supervised boundary annotations we use a class-specific
object detector \cite{Ren2015NipsFasterRCNN,Girshick2015IccvFastRCNN}.
We also experiment using directly the ground truth bounding box annotations.
This results in a minor drop in the performance. Using object detector
allows to remove hard cases by discarding images which have zero bounding
boxes with confidence scores above $0.8$. from the training set,
hence improvement in the performance. The results are shown in Figure
\ref{fig:GT-bbs}.

\paragraph{GrabCut, DenseCut and CNN+GraphCut}

For generating weakly supervised annotations in addition to GrabCut\cite{RotherGrabCut}
we also experimented with DenseCut \cite{Cheng2015CgfDenseCut} and
CNN+GraphCut \cite{Simonyan2014Iclr}. Employing DenseCut or CNN+GraphCut
does not bring any gain opposed to GrabCut. The results are presented
in Figure \ref{fig:caffecut}.

\paragraph{Using $\mbox{VOC}_{+}$}

Since we generate boundary annotations in a weakly supervised fashion,
we are able to generate boundaries over arbitrary image sets. In our
experiments we consider VOC (Pascal VOC12 segmentation task) and $\mbox{VOC}_{+}$
(VOC plus images from Pascal VOC12 detection task). Figure \ref{fig:VOC-results}
presents the results using VOC and $\mbox{VOC}_{+}$. Methods using
$\mbox{VOC}_{+}$ are denoted by $\cdot_{+}$ (e.g. $\mbox{SE}\left(\mbox{SeSe}_{+}\cap\mbox{BBs}\right)$).
Using a larger set of images for training allows to further improve
the performance of SE trained with the generated boundary annotations.

\section{\label{sec:VOC-boundaries-examples}VOC boundary detection examples}

Figure \ref{fig:Qualitative-results-voc} presents qualitative results
of boundary detection on VOC. The presented boundary estimate examples
show that high quality object boundaries can be achieved using only
detection bounding box annotations. This figure extends Figure 7 of
the main paper.

\section{\label{sec:Detailed-results-COCO}Detailed results for COCO}

\begin{figure}
\noindent \begin{centering}
\includegraphics[bb=0bp 90bp 612bp 710bp,clip,width=0.8\columnwidth]{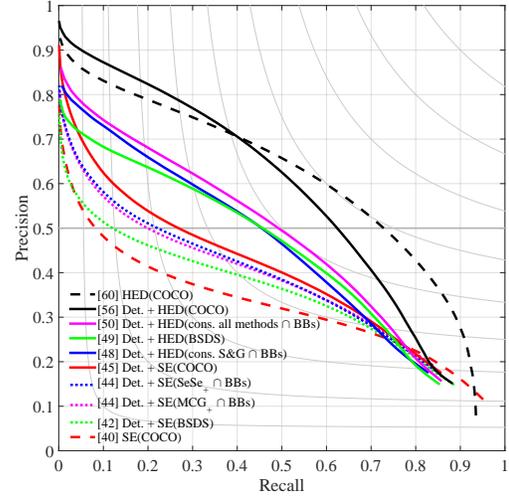}
\par\end{centering}

\protect\caption{\label{fig:COCO-results}COCO results. $\left(\cdot\right)$ denotes
the data used for training. Continuous/dashed line indicates models
using/not using a detector at test time. Legend indicates AP numbers.
For weakly supervised cases the results are shown with the models
trained on VOC, without re-training on COCO. }
\end{figure}

\begin{figure*}
\noindent \begin{centering}
\hfill{}\includegraphics[width=0.15\textwidth,height=0.125\textheight]{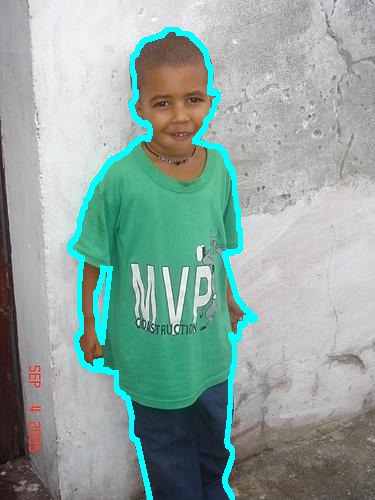}\,\includegraphics[width=0.15\textwidth,height=0.125\textheight]{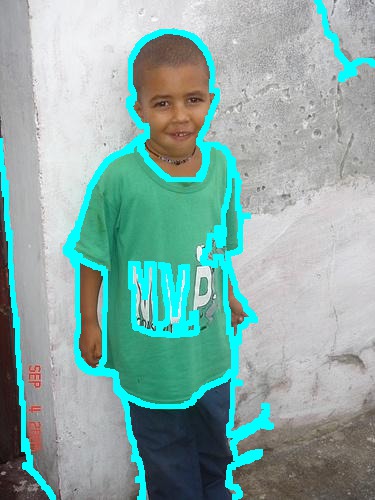}\,\includegraphics[width=0.15\textwidth,height=0.125\textheight]{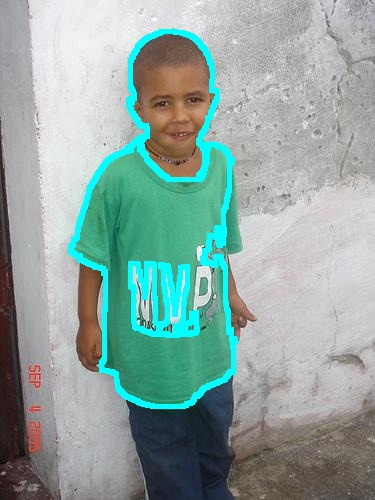}\,\includegraphics[width=0.15\textwidth,height=0.125\textheight]{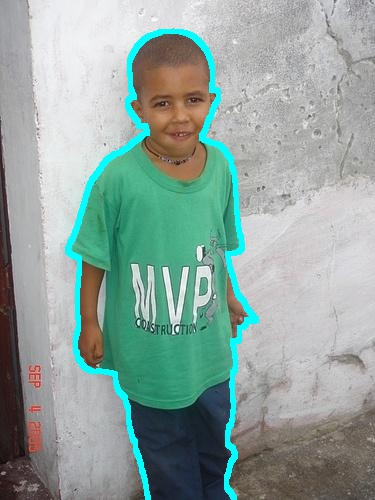}\,\includegraphics[width=0.15\textwidth,height=0.125\textheight]{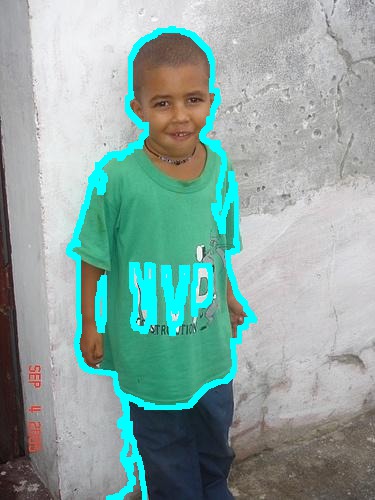}\hfill{}
\par\end{centering}

\noindent \begin{centering}
\vspace{-1.2em}

\par\end{centering}

\noindent \begin{centering}
\hfill{}\includegraphics[width=0.15\textwidth,height=0.085\textheight]{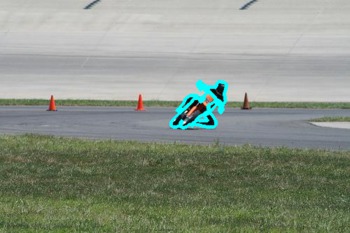}\,\includegraphics[width=0.15\textwidth,height=0.085\textheight]{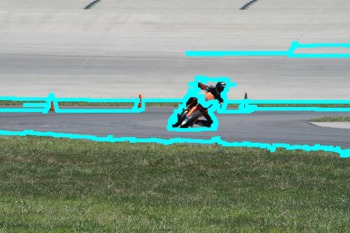}\,\includegraphics[width=0.15\textwidth,height=0.085\textheight]{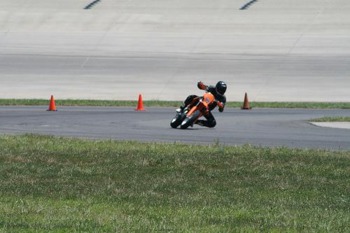}\,\includegraphics[width=0.15\textwidth,height=0.085\textheight]{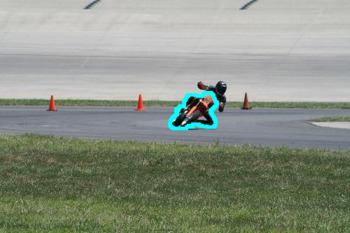}\,\includegraphics[width=0.15\textwidth,height=0.085\textheight]{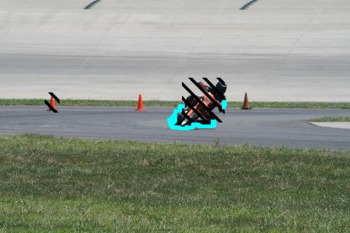}\hfill{}
\par\end{centering}

\noindent \begin{centering}
\vspace{-1.2em}

\par\end{centering}

\noindent \begin{centering}
\hfill{}\includegraphics[width=0.15\textwidth,height=0.085\textheight]{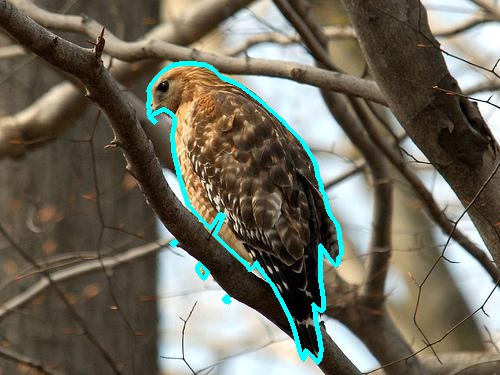}\,\includegraphics[width=0.15\textwidth,height=0.085\textheight]{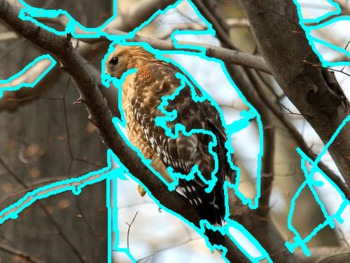}\,\includegraphics[width=0.15\textwidth,height=0.085\textheight]{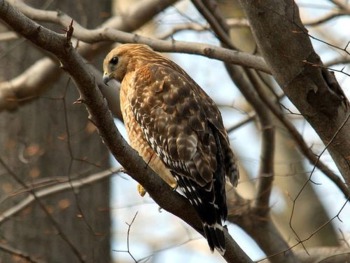}\,\includegraphics[width=0.15\textwidth,height=0.085\textheight]{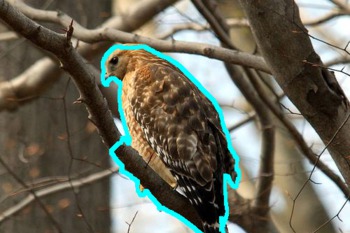}\,\includegraphics[width=0.15\textwidth,height=0.085\textheight]{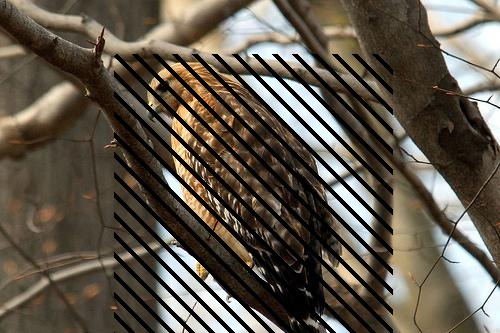}\hfill{}
\par\end{centering}

\noindent \begin{centering}
\vspace{-1.2em}

\par\end{centering}

\noindent \begin{centering}
\hfill{}\includegraphics[width=0.15\textwidth,height=0.08\textheight]{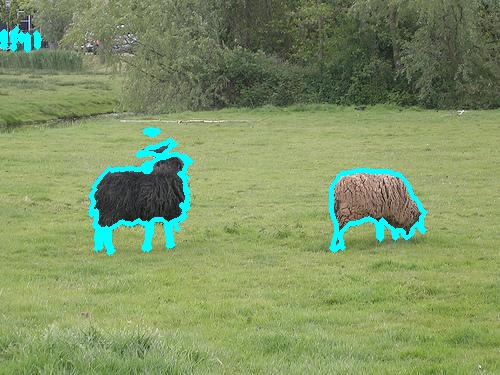}\,\includegraphics[width=0.15\textwidth,height=0.08\textheight]{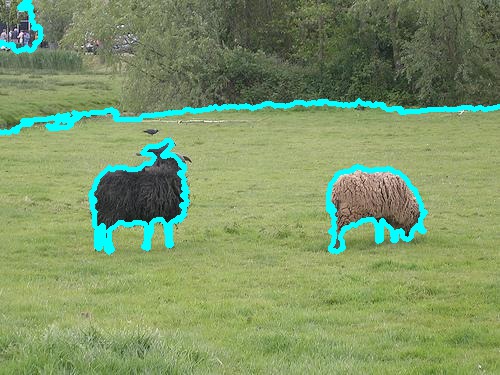}\,\includegraphics[width=0.15\textwidth,height=0.08\textheight]{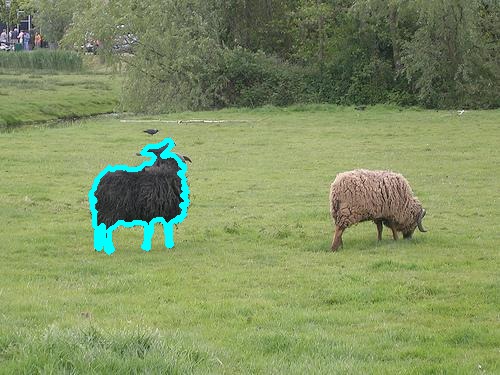}\,\includegraphics[width=0.15\textwidth,height=0.08\textheight]{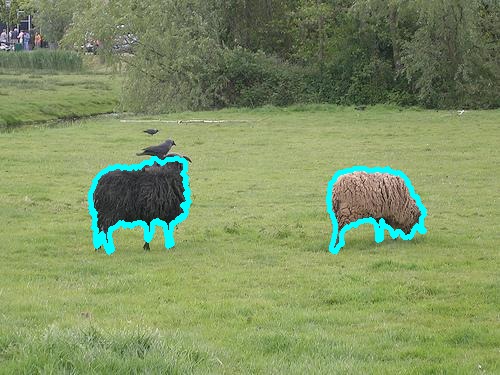}\,\includegraphics[width=0.15\textwidth,height=0.08\textheight]{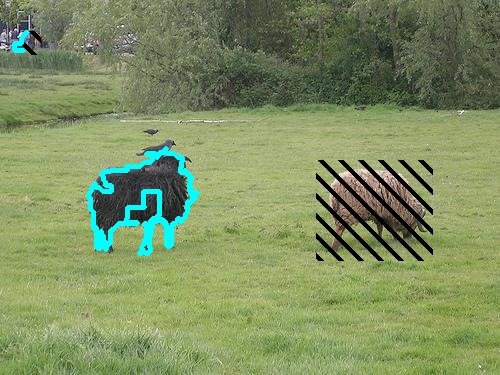}\hfill{}
\par\end{centering}

\noindent \begin{centering}
\vspace{-2.2em}

\par\end{centering}

\noindent \begin{centering}
\hfill{}\subfloat[\label{fig:Ground-truth-1}Ground truth]{\protect\includegraphics[width=0.15\textwidth,height=0.08\textheight]{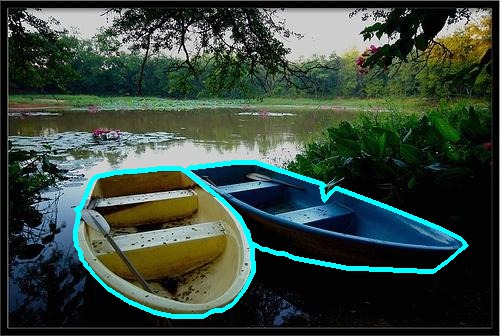}}\,\subfloat[\label{fig:F=000026H-1}$\protect\FH$]{\protect\includegraphics[width=0.15\textwidth,height=0.08\textheight]{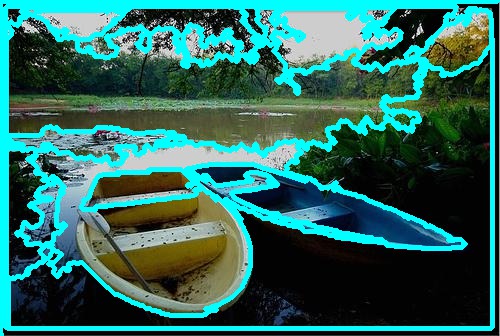}}\,\subfloat[\label{fig:F=000026H-BBs-1}$\protect\FH\cap\mbox{BBs}$]{\protect\includegraphics[width=0.15\textwidth,height=0.08\textheight]{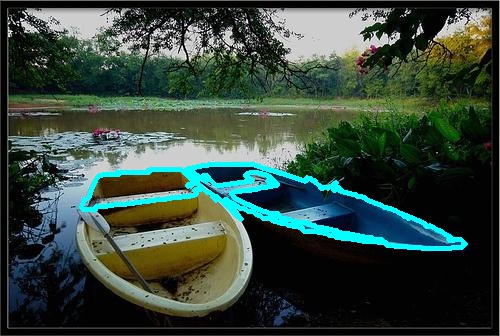}}\,\subfloat[\label{fig:GrabCut-BBs-1}GrabCut $\cap$ BBs]{\protect\includegraphics[width=0.15\textwidth,height=0.08\textheight]{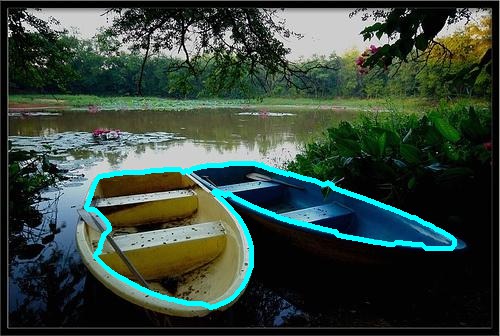}}\,\subfloat[\label{fig:SeSe-BBs-1}SeSe $\cap$ BBs]{\protect\includegraphics[width=0.15\textwidth,height=0.08\textheight]{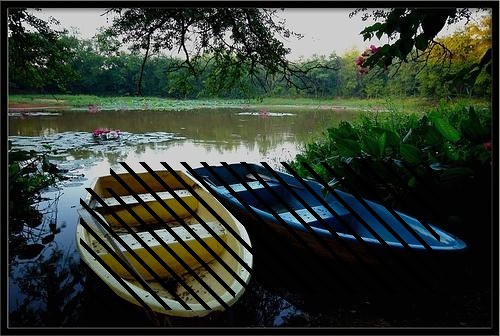}}\hfill{}
\par\end{centering}

\noindent \begin{centering}
\vspace{-0.2em}

\par\end{centering}

\noindent \begin{centering}
\hfill{}\includegraphics[width=0.15\textwidth,height=0.125\textheight]{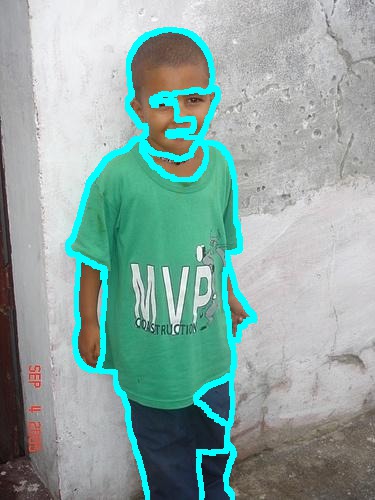}\,\includegraphics[width=0.15\textwidth,height=0.125\textheight]{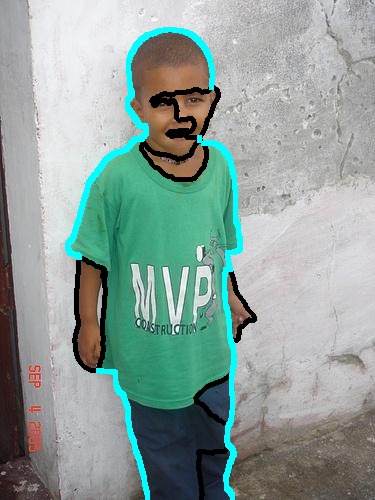}\,\includegraphics[width=0.15\textwidth,height=0.125\textheight]{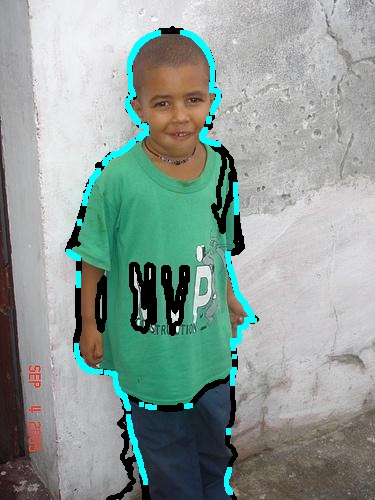}\,\includegraphics[width=0.15\textwidth,height=0.125\textheight]{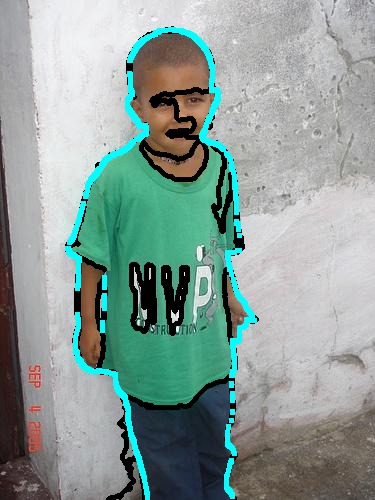}\,\includegraphics[width=0.15\textwidth,height=0.125\textheight]{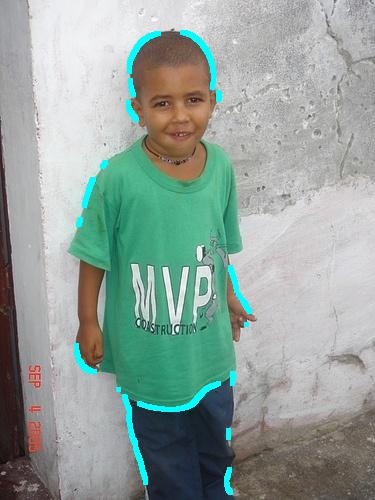}\hfill{}
\par\end{centering}

\noindent \begin{centering}
\vspace{-1.2em}

\par\end{centering}

\noindent \begin{centering}
\hfill{}\includegraphics[width=0.15\textwidth,height=0.085\textheight]{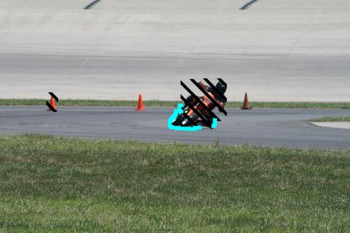}\,\includegraphics[width=0.15\textwidth,height=0.085\textheight]{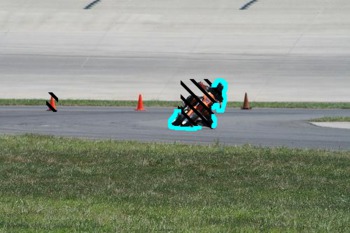}\,\includegraphics[width=0.15\textwidth,height=0.085\textheight]{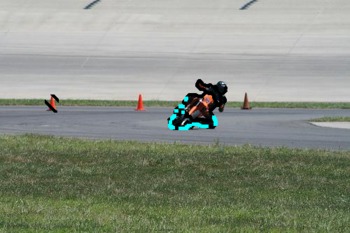}\,\includegraphics[width=0.15\textwidth,height=0.085\textheight]{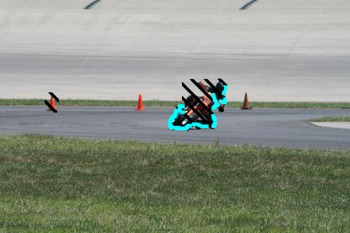}\,\includegraphics[width=0.15\textwidth,height=0.085\textheight]{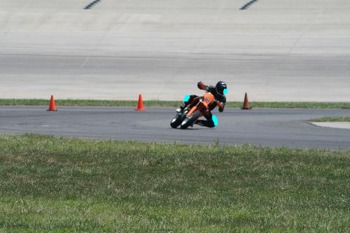}\hfill{}
\par\end{centering}

\noindent \begin{centering}
\vspace{-1.2em}

\par\end{centering}

\noindent \begin{centering}
\hfill{}\includegraphics[width=0.15\textwidth,height=0.085\textheight]{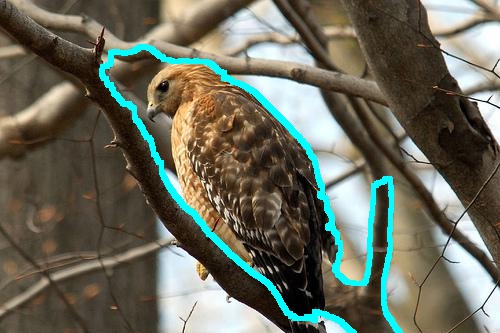}\,\includegraphics[width=0.15\textwidth,height=0.085\textheight]{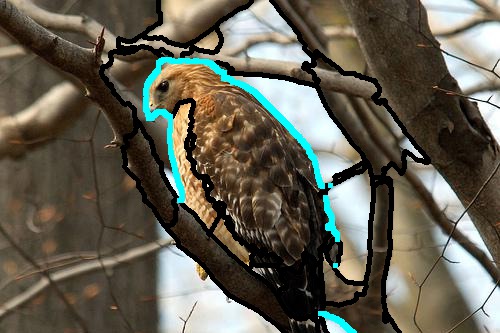}\,\includegraphics[width=0.15\textwidth,height=0.085\textheight]{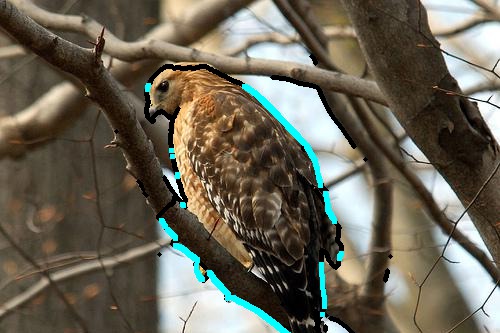}\,\includegraphics[width=0.15\textwidth,height=0.085\textheight]{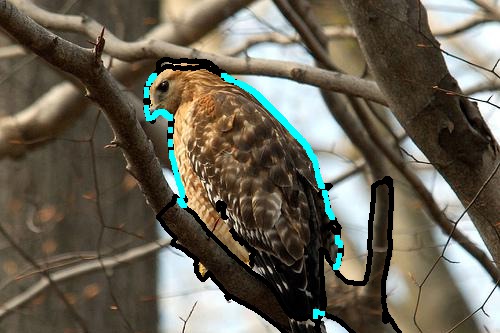}\,\includegraphics[width=0.15\textwidth,height=0.085\textheight]{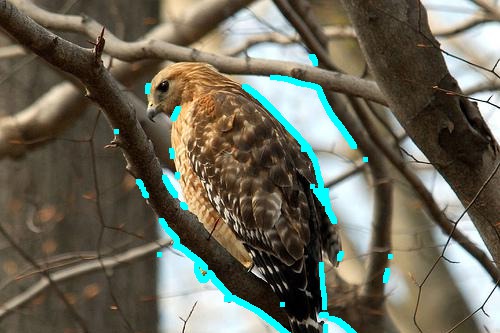}\hfill{}
\par\end{centering}

\noindent \begin{centering}
\vspace{-1.2em}

\par\end{centering}

\noindent \begin{centering}
\hfill{}\includegraphics[width=0.15\textwidth,height=0.08\textheight]{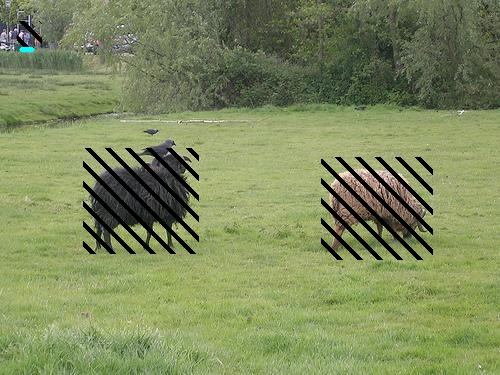}\,\includegraphics[width=0.15\textwidth,height=0.08\textheight]{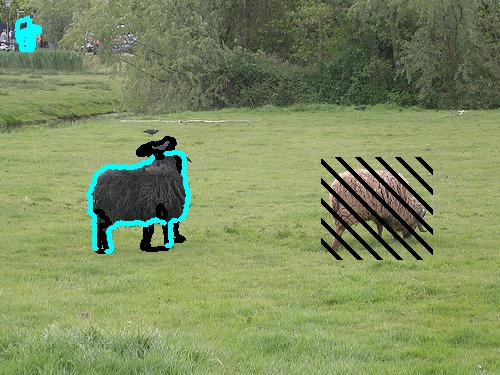}\,\includegraphics[width=0.15\textwidth,height=0.08\textheight]{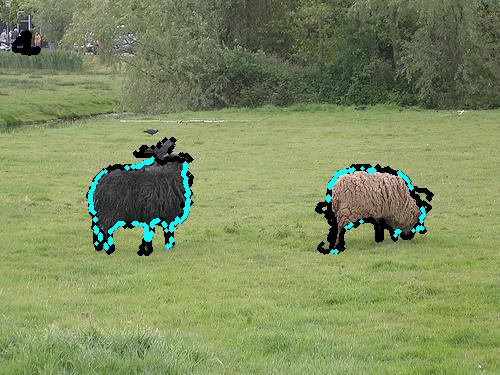}\,\includegraphics[width=0.15\textwidth,height=0.08\textheight]{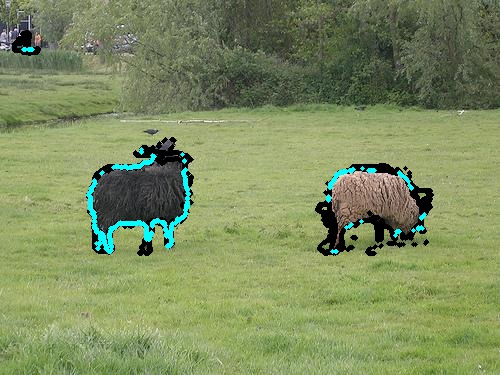}\,\includegraphics[width=0.15\textwidth,height=0.08\textheight]{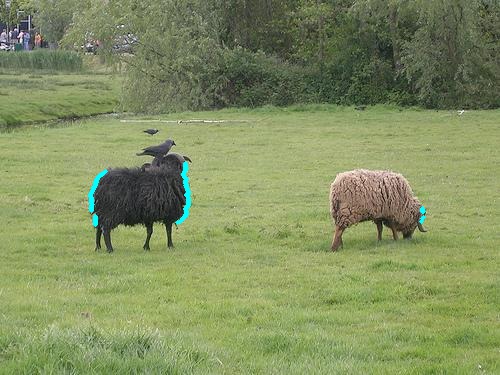}\hfill{}
\par\end{centering}

\noindent \begin{centering}
\vspace{-2.2em}

\par\end{centering}

\noindent \begin{centering}
\hfill{}\subfloat[\label{fig:MCG-BBs-1}MCG $\cap$ BBs]{\protect\includegraphics[width=0.15\textwidth,height=0.08\textheight]{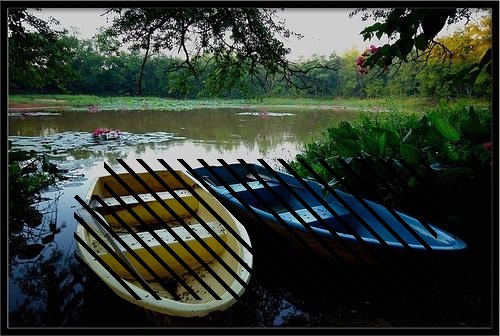}}\,\subfloat[\label{fig:cons.-MCG-BBs-1}cons. MCG $\cap$ BBs]{\protect\includegraphics[width=0.15\textwidth,height=0.08\textheight]{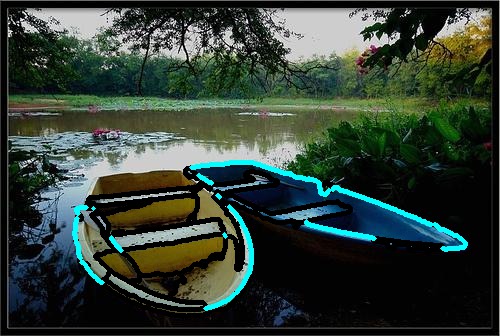}}\,\subfloat[\label{fig:cons.-all-methods-sese-1}cons. S\&G $\cap$ BBs]{\protect\includegraphics[width=0.15\textwidth,height=0.08\textheight]{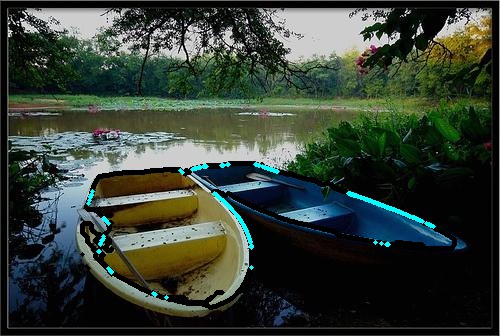}}\,\subfloat[\label{fig:cons.-all-methods-1}cons. all $\cap$ BBs]{\protect\includegraphics[width=0.15\textwidth,height=0.08\textheight]{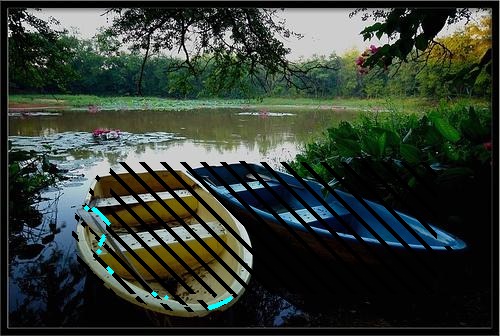}}\,\subfloat[\label{fig:SE(SeSe-BBs)-1}SE(SeSe $\cap$ BBs)]{\protect\includegraphics[width=0.15\textwidth,height=0.08\textheight]{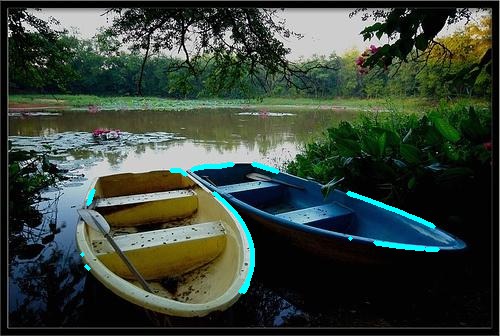}}\hfill{}
\par\end{centering}

\vspace{-0.5em}

\protect\caption{\label{fig:Noisy-boundaries-1}Different generated boundary annotations.
Cyan/black indicates positive/ignored boundaries.}
\end{figure*}

\begin{figure*}
\begingroup{\arrayrulecolor{lightgray}
\setlength{\tabcolsep}{0pt} 
\renewcommand{\arraystretch}{0}

\begin{tabular}[b]{|c|c|c||c|c||c|c|}
\hline 
\includegraphics[width=0.14\textwidth]{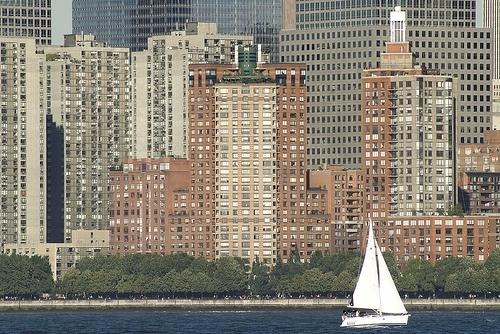} & \includegraphics[width=0.14\textwidth]{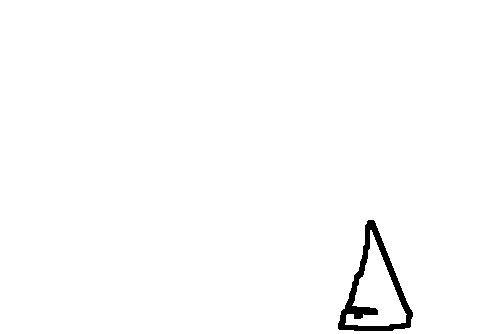} & \includegraphics[width=0.14\textwidth]{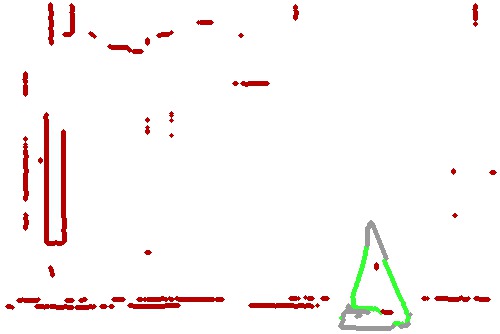} & \includegraphics[width=0.14\textwidth]{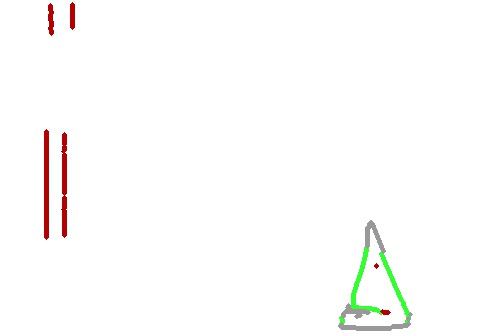} & \includegraphics[width=0.14\textwidth]{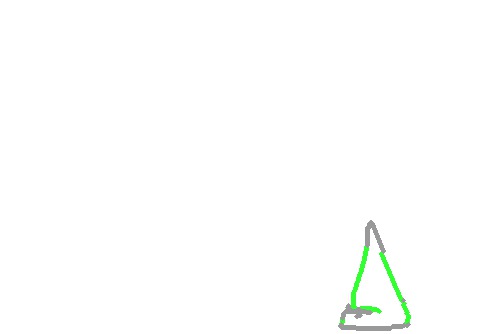} & \includegraphics[width=0.14\textwidth]{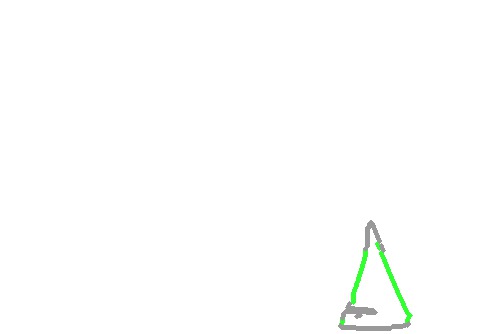} & \includegraphics[width=0.14\textwidth]{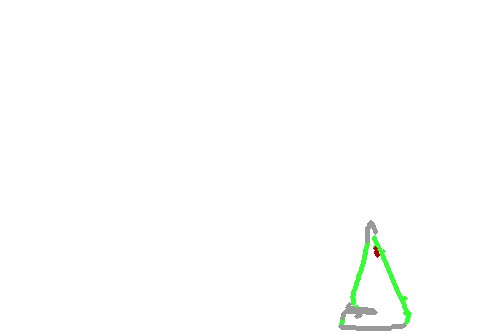}\tabularnewline
\hline 
\includegraphics[width=0.14\textwidth]{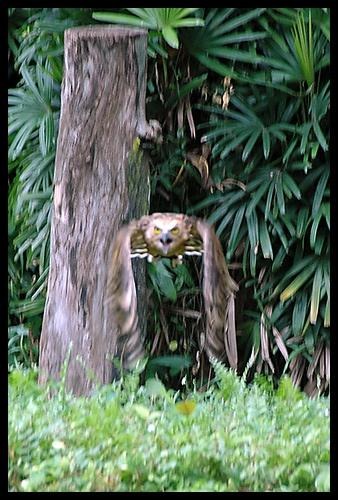} & \includegraphics[width=0.14\textwidth]{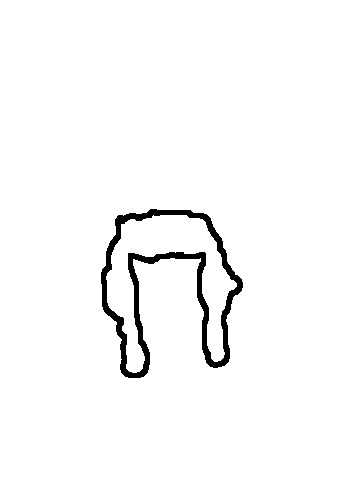} & \includegraphics[width=0.14\textwidth]{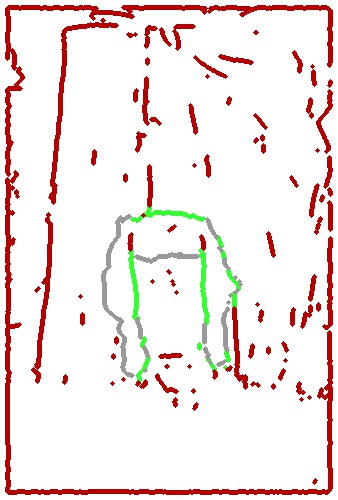} & \includegraphics[width=0.14\textwidth]{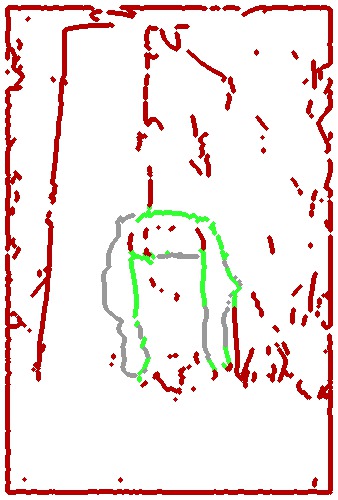} & \includegraphics[width=0.14\textwidth]{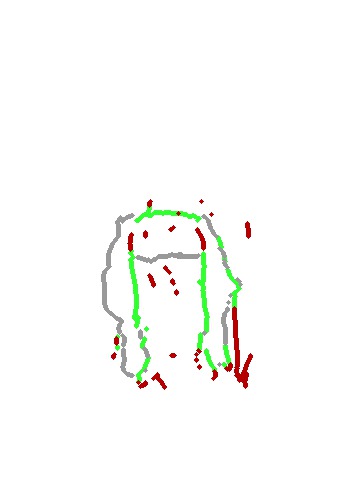} & \includegraphics[width=0.14\textwidth]{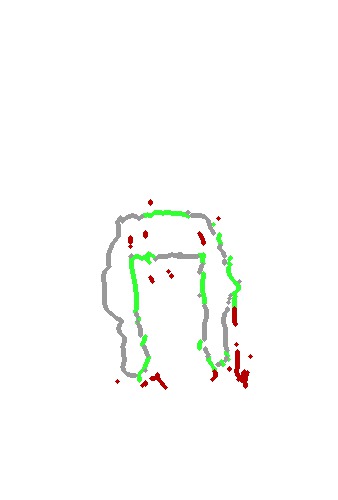} & \includegraphics[width=0.14\textwidth]{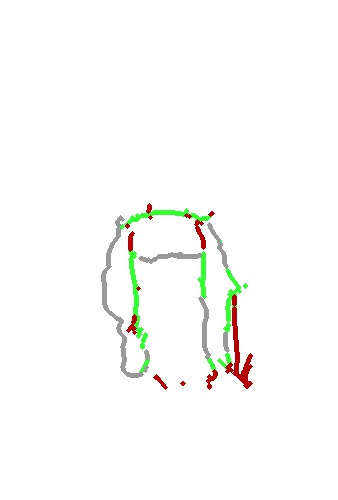}\tabularnewline
\hline 
\includegraphics[width=0.14\textwidth]{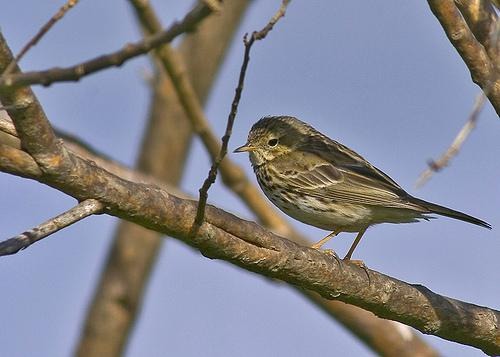} & \includegraphics[width=0.14\textwidth]{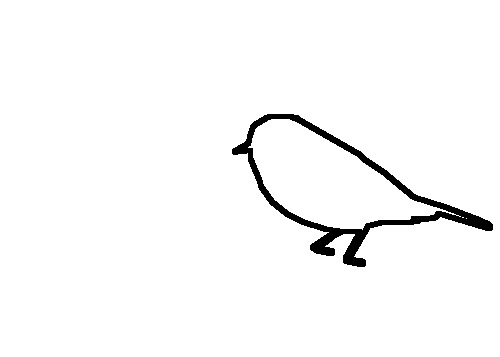} & \includegraphics[width=0.14\textwidth]{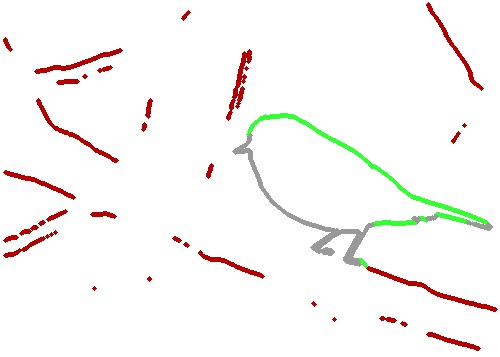} & \includegraphics[width=0.14\textwidth]{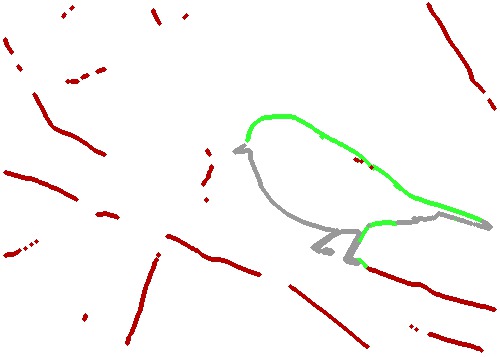} & \includegraphics[width=0.14\textwidth]{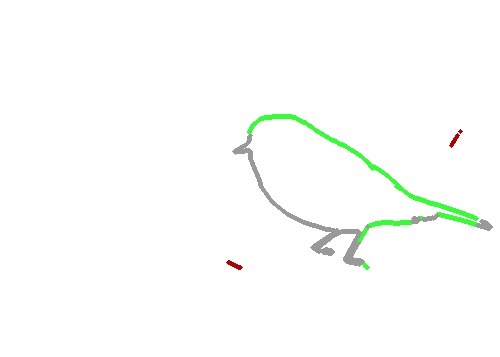} & \includegraphics[width=0.14\textwidth]{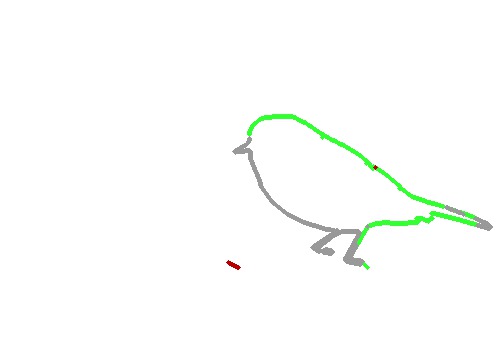} & \includegraphics[width=0.14\textwidth]{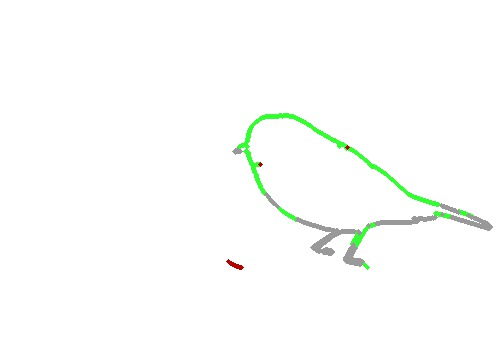}\tabularnewline
\hline 
\includegraphics[width=0.14\textwidth]{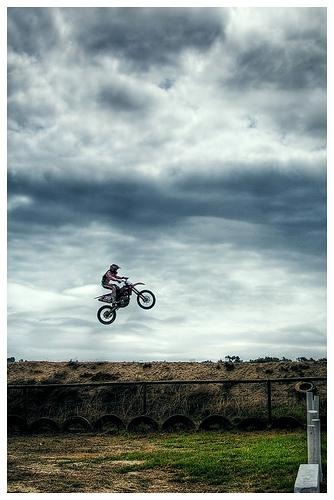} & \includegraphics[width=0.14\textwidth]{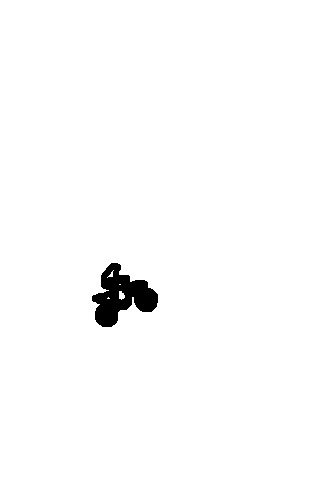} & \includegraphics[width=0.14\textwidth]{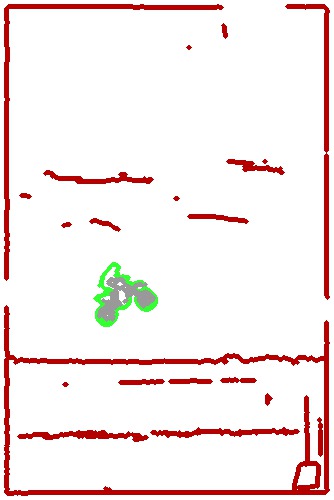} & \includegraphics[width=0.14\textwidth]{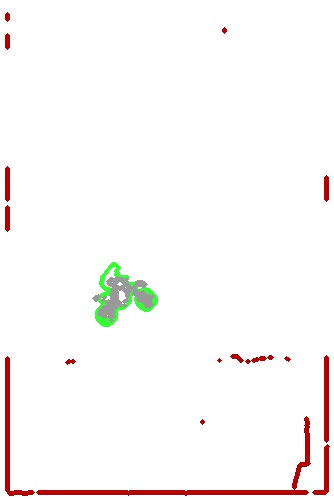} & \includegraphics[width=0.14\textwidth]{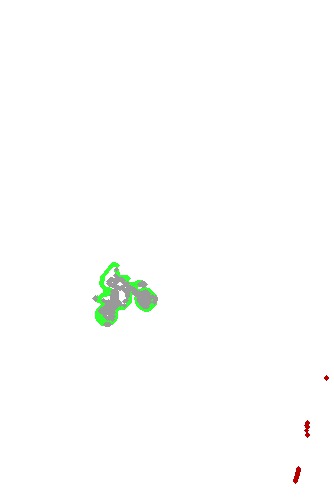} & \includegraphics[width=0.14\textwidth]{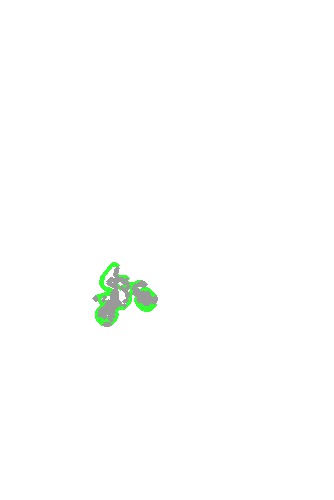} & \includegraphics[width=0.14\textwidth]{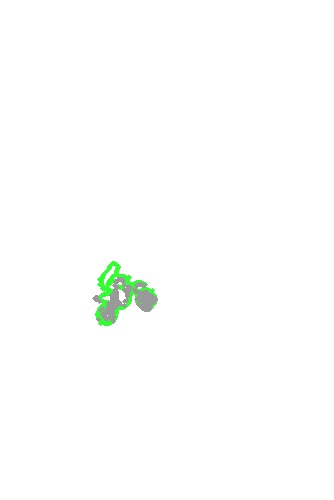}\tabularnewline
\hline 
\includegraphics[width=0.14\textwidth]{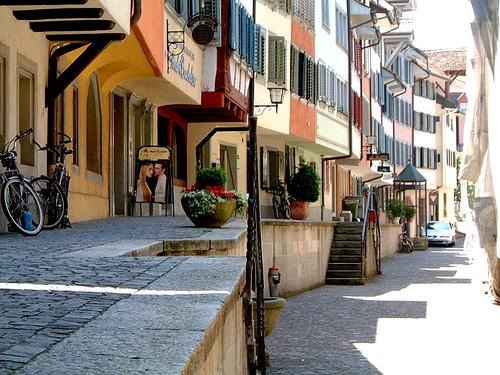} & \includegraphics[width=0.14\textwidth]{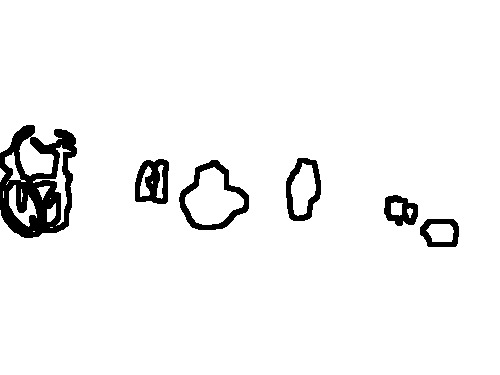} & \includegraphics[width=0.14\textwidth]{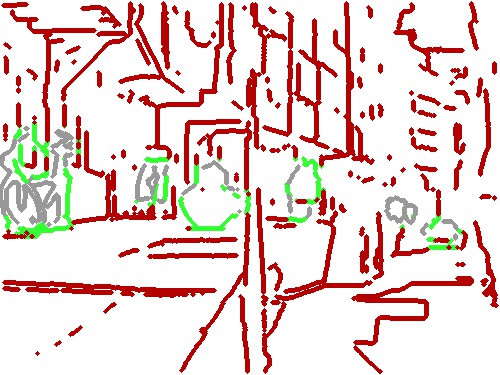} & \includegraphics[width=0.14\textwidth]{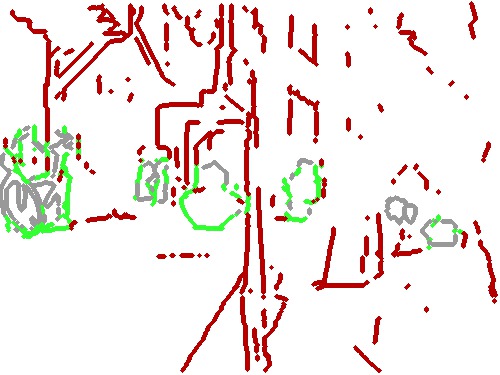} & \includegraphics[width=0.14\textwidth]{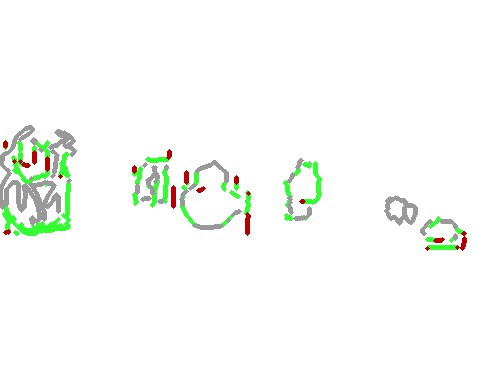} & \includegraphics[width=0.14\textwidth]{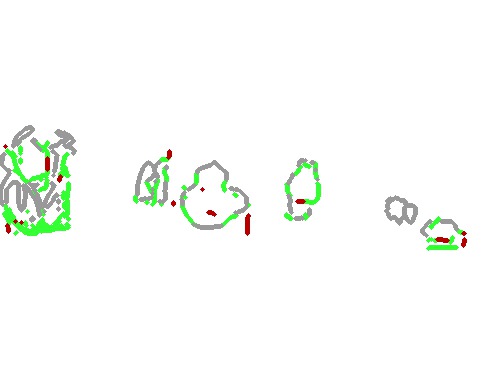} & \includegraphics[width=0.14\textwidth]{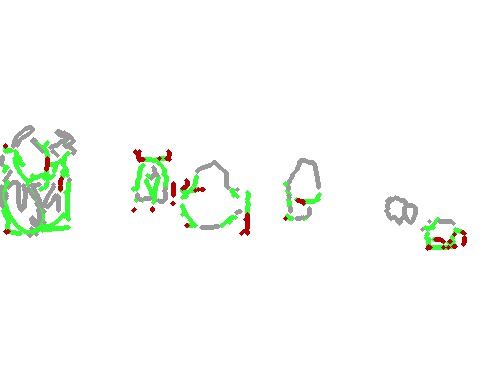}\tabularnewline
\hline 
\includegraphics[width=0.14\textwidth]{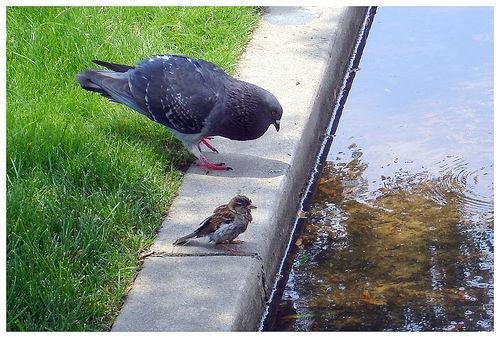} & \includegraphics[width=0.14\textwidth]{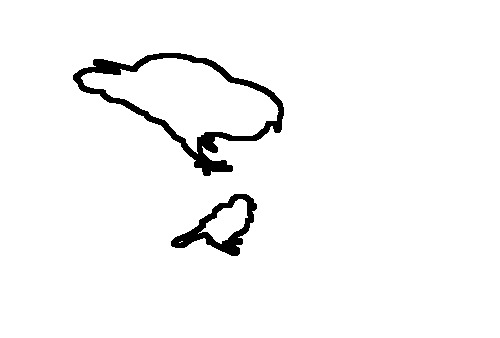} & \includegraphics[width=0.14\textwidth]{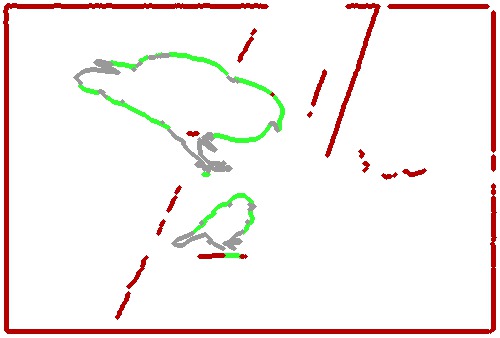} & \includegraphics[width=0.14\textwidth]{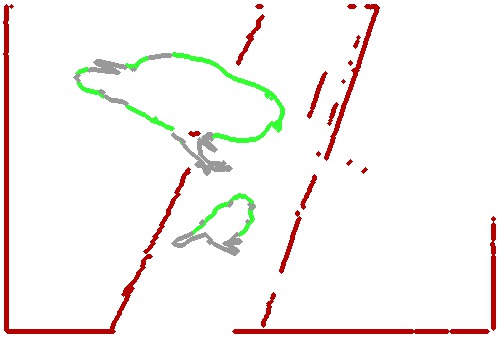} & \includegraphics[width=0.14\textwidth]{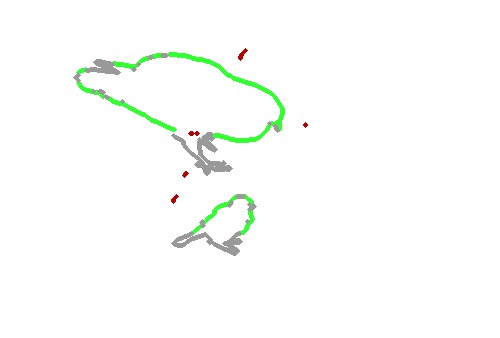} & \includegraphics[width=0.14\textwidth]{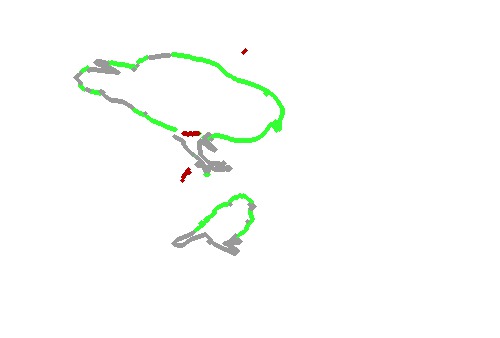} & \includegraphics[width=0.14\textwidth]{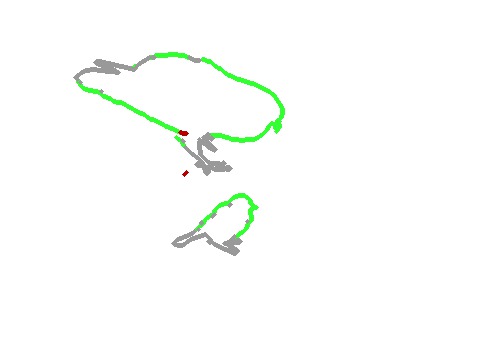}\tabularnewline
\hline 
\includegraphics[width=0.14\textwidth]{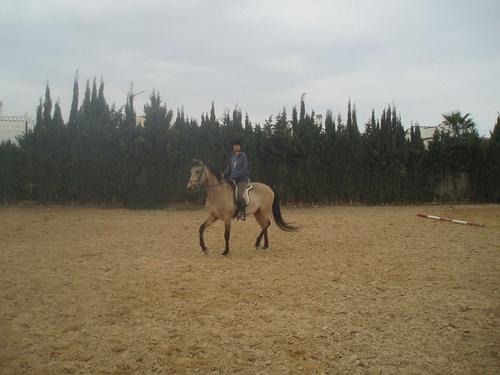} & \includegraphics[width=0.14\textwidth]{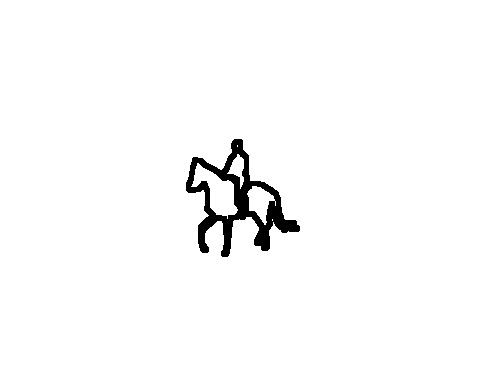} & \includegraphics[width=0.14\textwidth]{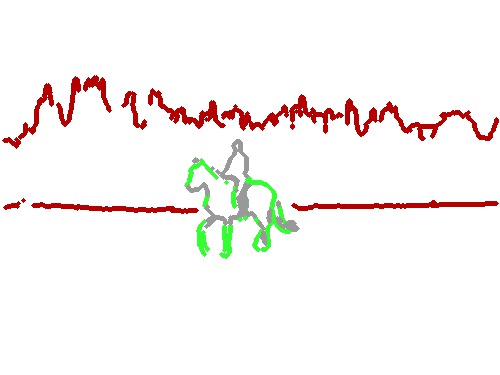} & \includegraphics[width=0.14\textwidth]{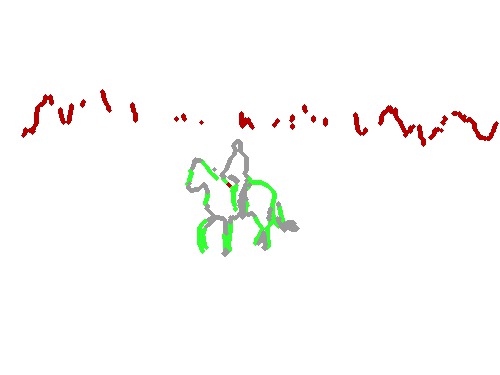} & \includegraphics[width=0.14\textwidth]{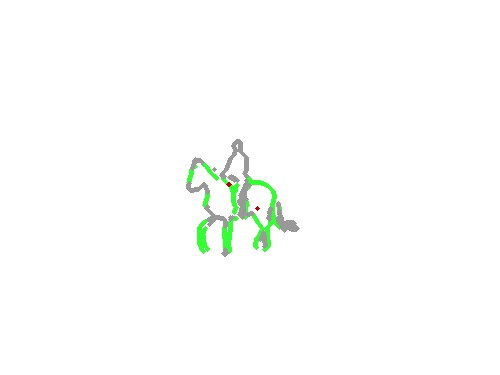} & \includegraphics[width=0.14\textwidth]{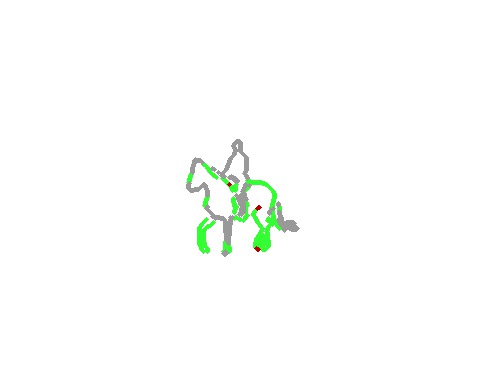} & \includegraphics[width=0.14\textwidth]{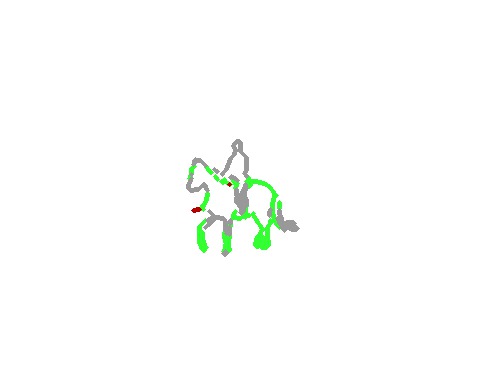}\tabularnewline
\hline 
\multicolumn{1}{c}{\selectlanguage{english}%
{\small{}\rule{0pt}{2.6ex}}\foreignlanguage{british}{{\small{}Image}}\selectlanguage{british}%
} & \multicolumn{1}{c}{{\small{}Ground truth}} & \multicolumn{1}{c}{{\small{}SE(BSDS)}} & \multicolumn{1}{c}{{\small{}SB(VOC)}} & \multicolumn{1}{c}{{\small{}$\mbox{Det.}\negmedspace+\negmedspace\mbox{SE}\left(\mbox{VOC}\right)$}} & \multicolumn{1}{c}{{\small{}$\mbox{Det.}\negmedspace+\negmedspace\mbox{SE}\left(\mathbf{weak}\right)$}} & \multicolumn{1}{c}{{\small{}$\mbox{Det.}\negthickspace+\negthickspace\mbox{HED}\left(\mathbf{weak}\right)$}}\tabularnewline
\end{tabular}

\protect\caption{\label{fig:Qualitative-results-voc}Qualitative results on VOC. $\left(\cdot\right)$
denotes the data used for training. Red/green indicate false/true
positive pixels, grey is missing recall. All methods are shown at
$50\%$ recall. {\small{}$\mbox{Det.}\negmedspace+\negmedspace\mbox{SE}\left(\mathbf{weak}\right)$}
denotes the model {\small{}$\mbox{Det.}\negmedspace+\negmedspace\mbox{SE}\left(\mbox{SeSe}_{+}\cap\mbox{BBs}\right)$}
and {\small{}$\mbox{Det.}\negthickspace+\negthickspace\mbox{HED}\left(\mathbf{weak}\right)$}
denotes {\small{}$\mbox{Det.}\negthickspace+\negthickspace\mbox{HED}\left(\textrm{cons.\mbox{ S\&G}\ensuremath{\cap}BBs}\right)$}.
Object-specific boundaries differ from generic boundaries (such as
the ones detected by SE(BSDS)). By using an object detector we can
suppress non-object boundaries and focus boundary detection on the
classes of interest. The proposed weakly supervised techniques allow
to achieve high quality boundary estimates that are similar to the
ones obtained by fully supervised methods. }
}\endgroup
\arrayrulecolor{black}
\end{figure*}

Figure \ref{fig:COCO-results} shows the generalization of the proposed
weakly supervised variants for object boundary detection on the COCO
dataset. For weakly supervised cases the results are shown with the
models trained on VOC, without re-training on COCO. These curves complement
Table 3 from the main paper.

For both SE and HED the models trained on the proposed weak annotations
perform as well as the fully supervised SE models. Similar to the
VOC benchmark the HED model trained on ground truth shows superior
performance.

\section{\label{sec:COCO-boundaries-examples}COCO boundary detection examples}

\begin{figure*}
\begingroup{\arrayrulecolor{lightgray}
\setlength{\tabcolsep}{0pt} 
\renewcommand{\arraystretch}{0}

\begin{tabular}[b]{|c|c|c||c|c||c|c|}
\hline 
\includegraphics[width=0.14\textwidth]{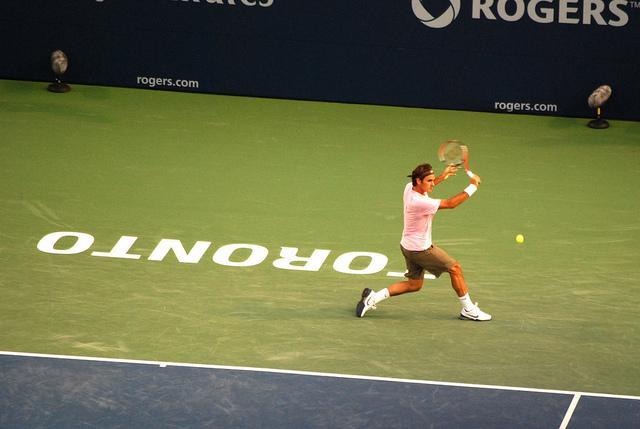} & \includegraphics[width=0.14\textwidth]{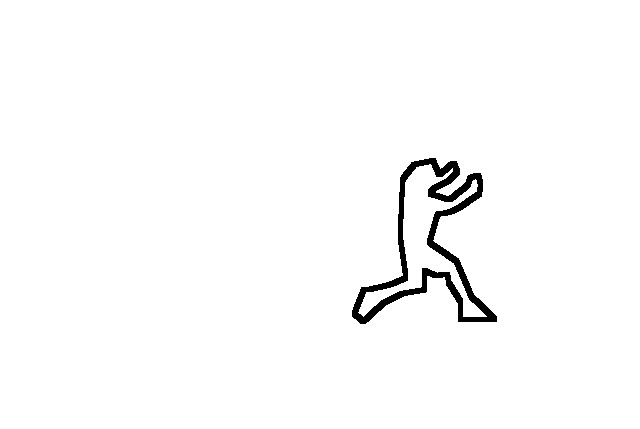} & \includegraphics[width=0.14\textwidth]{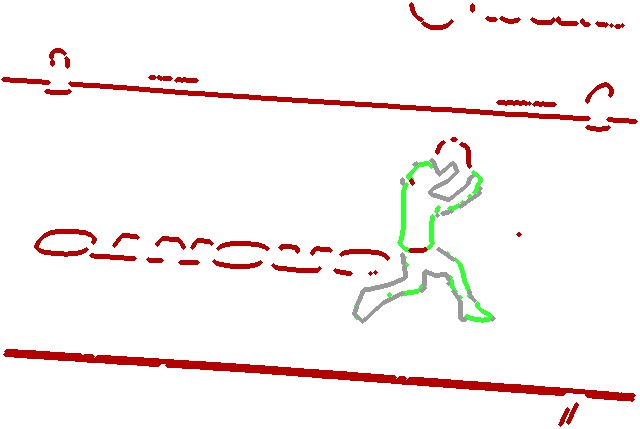} & \includegraphics[width=0.14\textwidth]{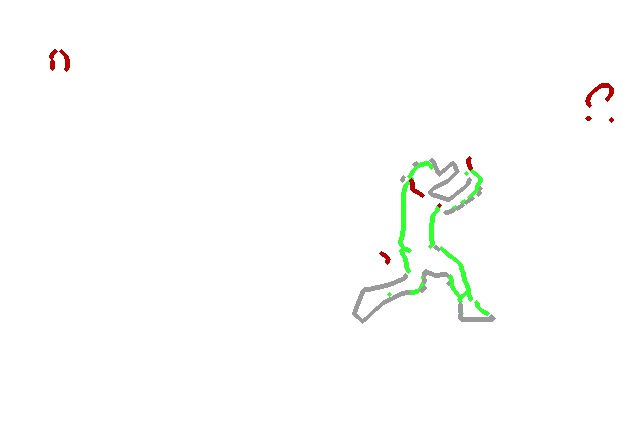} & \includegraphics[width=0.14\textwidth]{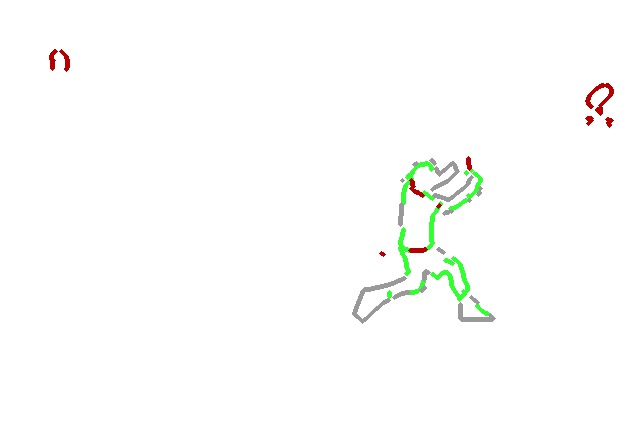} & \includegraphics[width=0.14\textwidth]{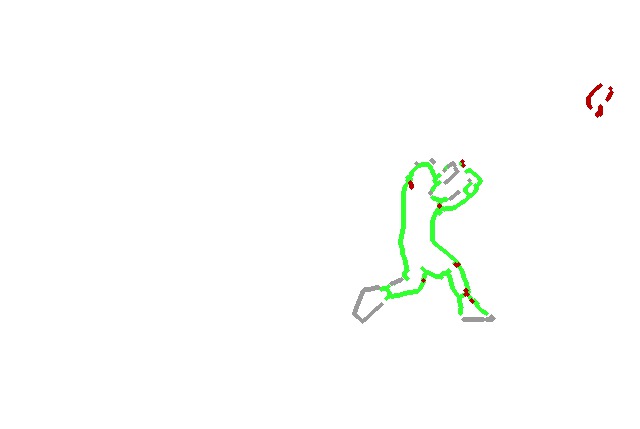} & \includegraphics[width=0.14\textwidth]{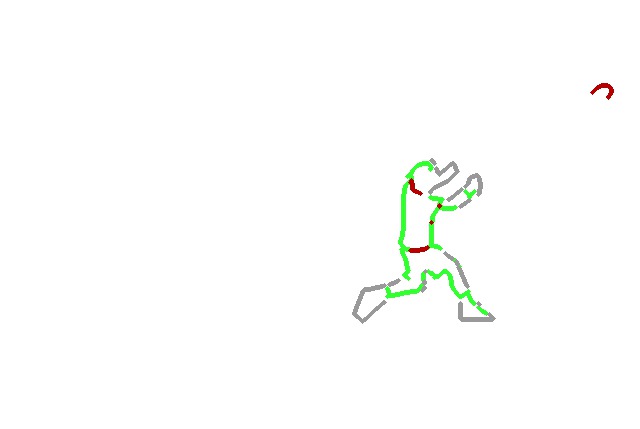}\tabularnewline
\hline 
\includegraphics[width=0.14\textwidth]{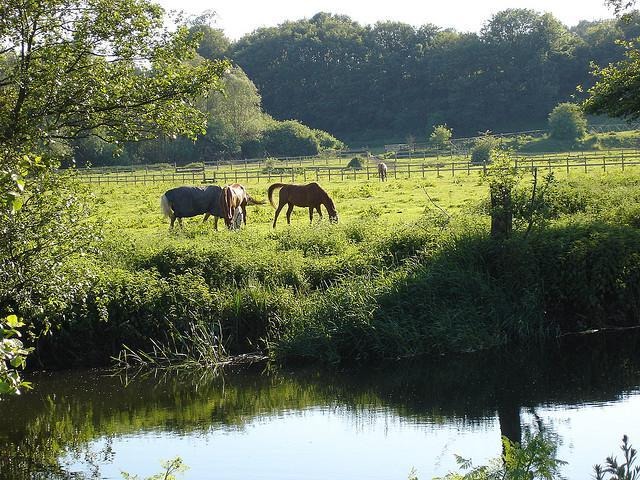} & \includegraphics[width=0.14\textwidth]{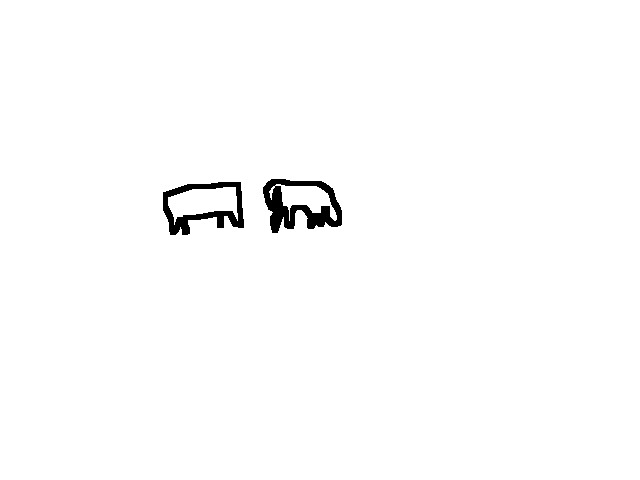} & \includegraphics[width=0.14\textwidth]{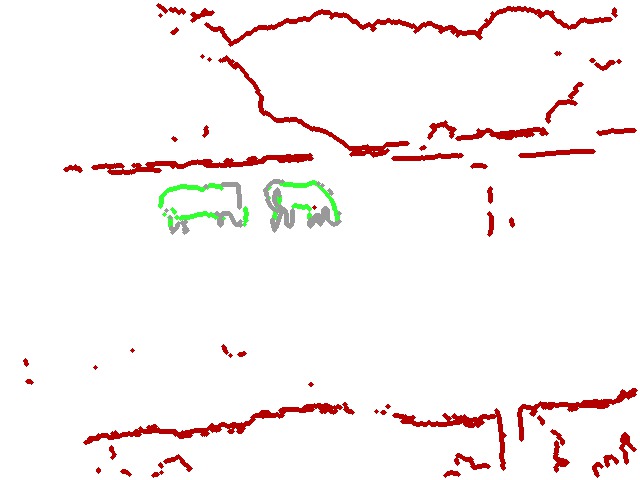} & \includegraphics[width=0.14\textwidth]{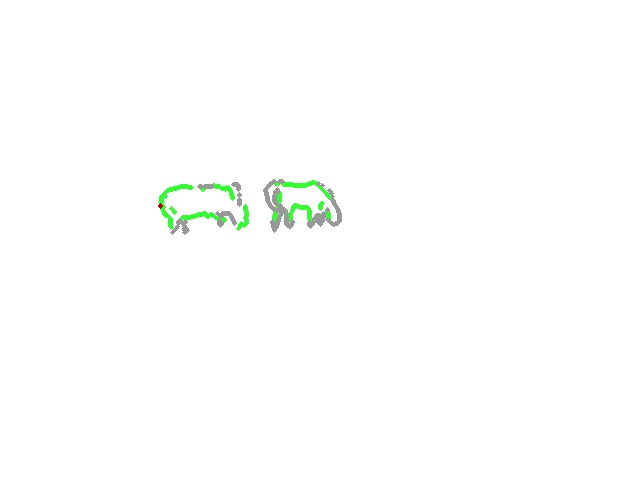} & \includegraphics[width=0.14\textwidth]{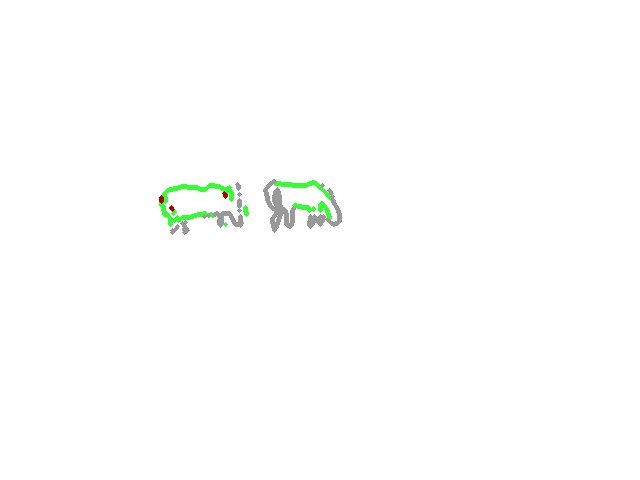} & \includegraphics[width=0.14\textwidth]{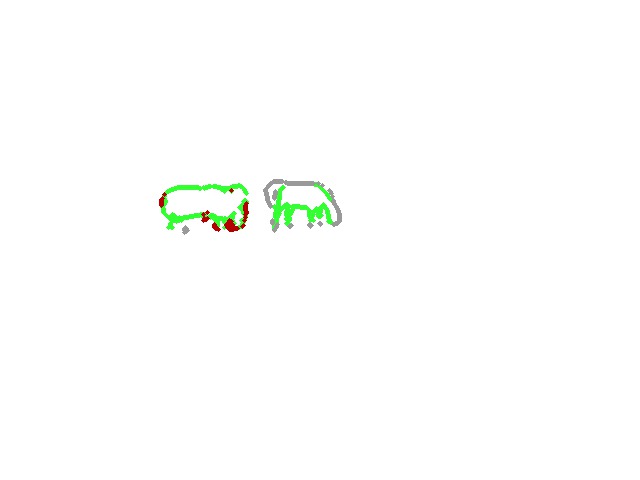} & \includegraphics[width=0.14\textwidth]{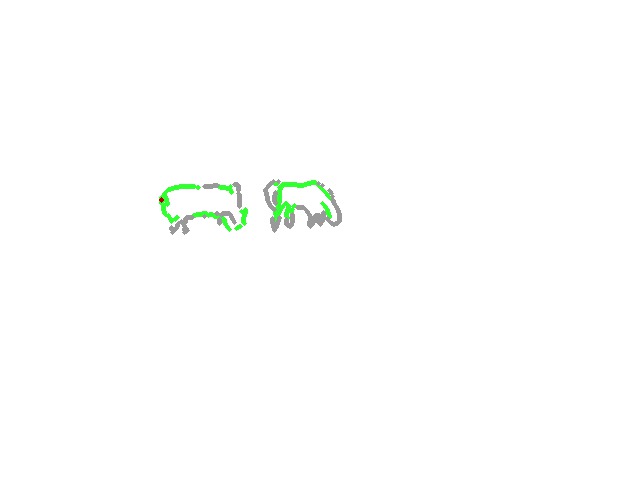}\tabularnewline
\hline 
\includegraphics[width=0.14\textwidth]{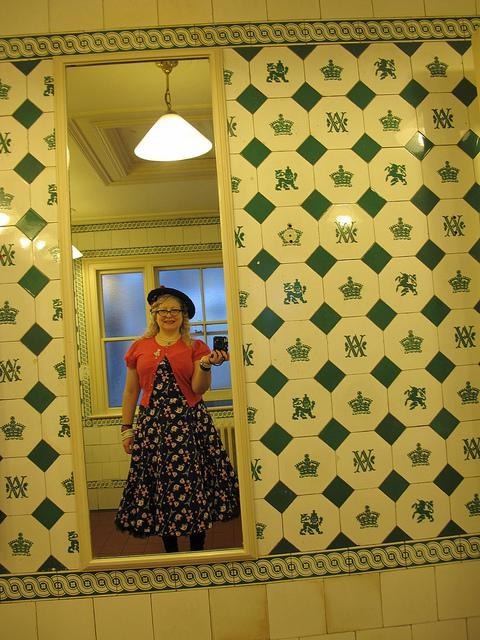} & \includegraphics[width=0.14\textwidth]{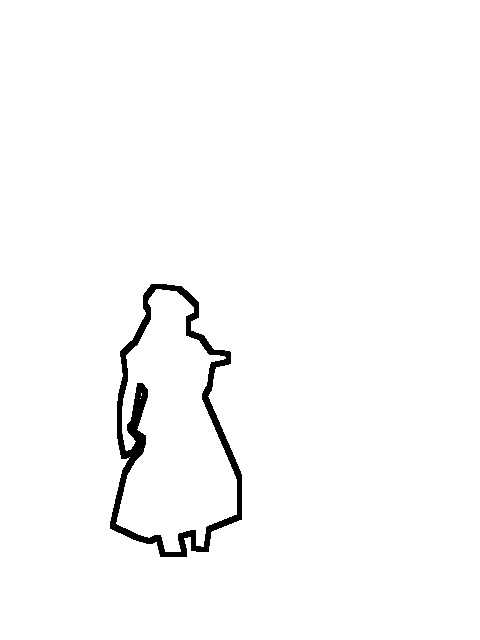} & \includegraphics[width=0.14\textwidth]{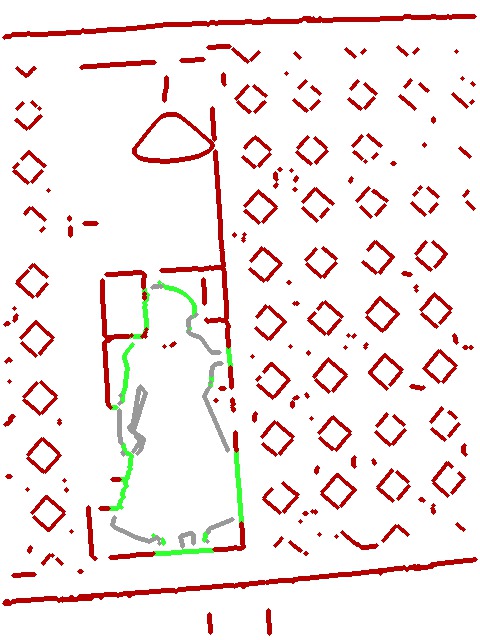} & \includegraphics[width=0.14\textwidth]{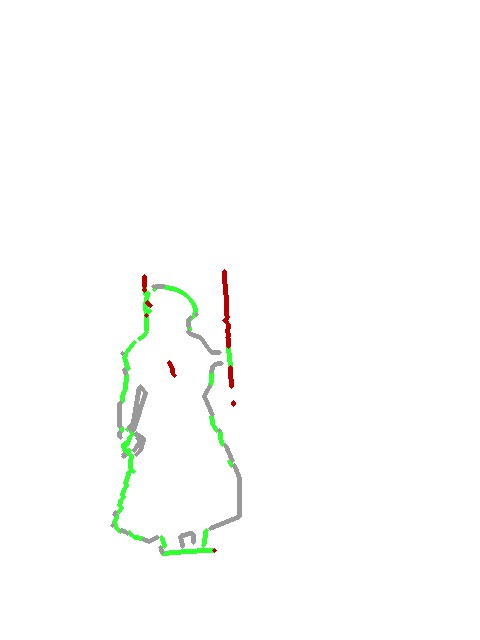} & \includegraphics[width=0.14\textwidth]{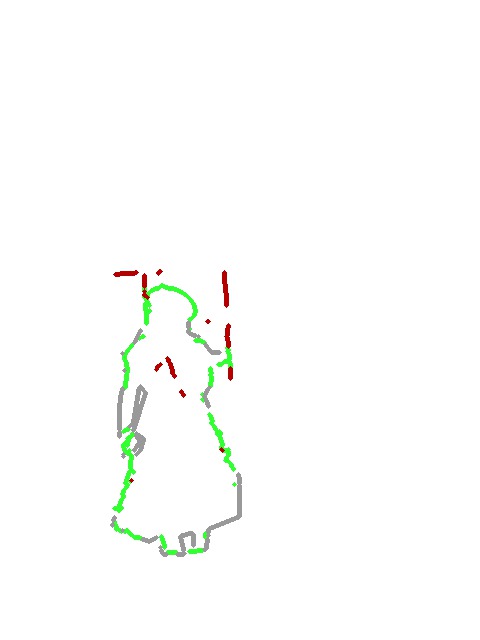} & \includegraphics[width=0.14\textwidth]{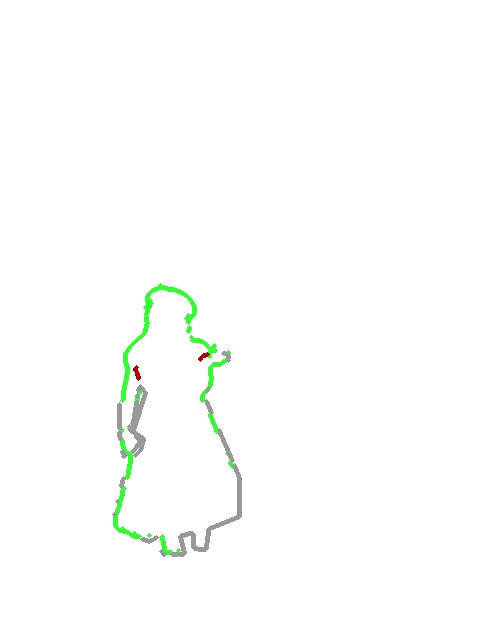} & \includegraphics[width=0.14\textwidth]{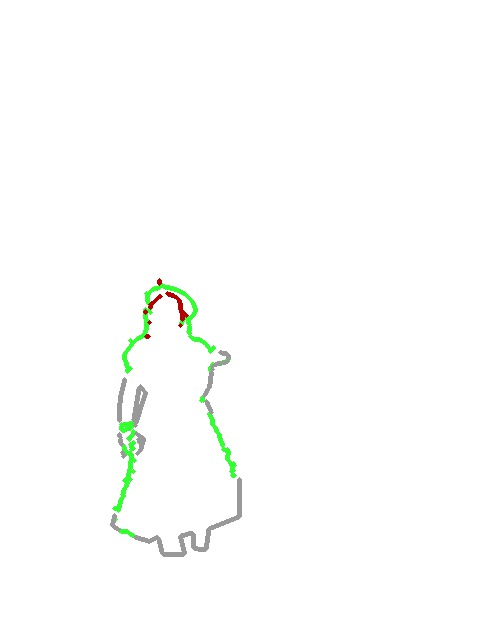}\tabularnewline
\hline 
\includegraphics[width=0.14\textwidth]{figures/sup_material/coco_results/COCO_val2014_000000154087_im} & \includegraphics[width=0.14\textwidth]{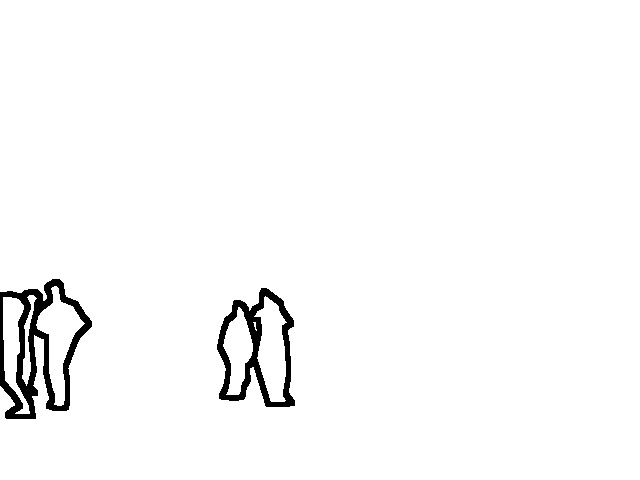} & \includegraphics[width=0.14\textwidth]{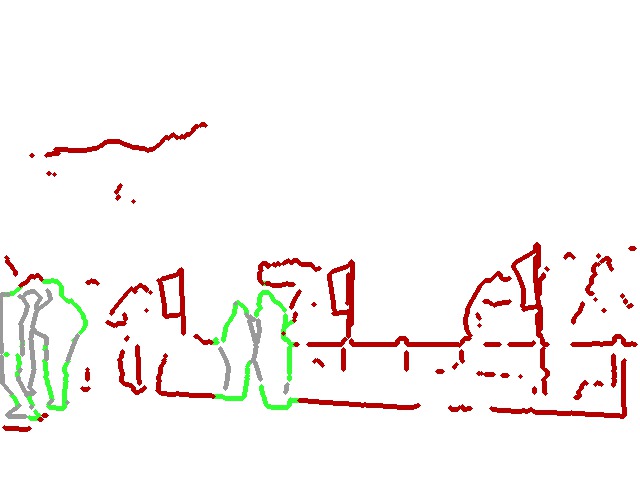} & \includegraphics[width=0.14\textwidth]{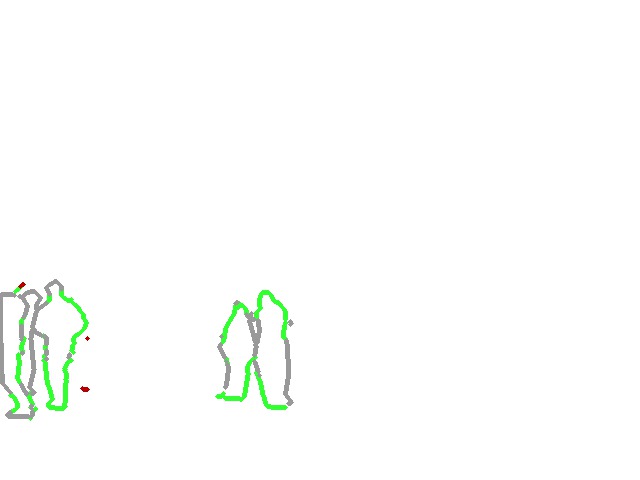} & \includegraphics[width=0.14\textwidth]{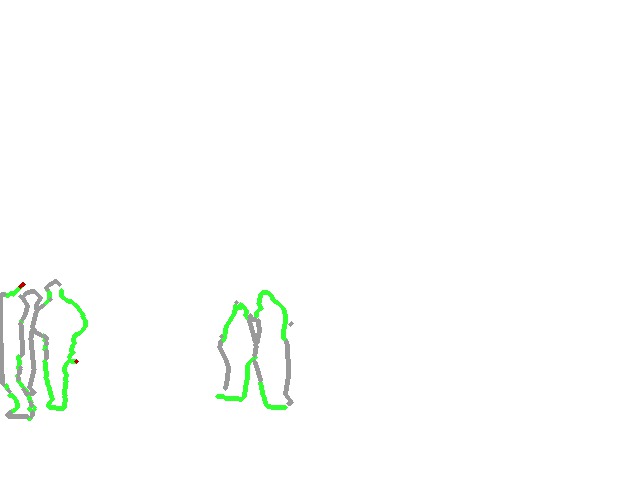} & \includegraphics[width=0.14\textwidth]{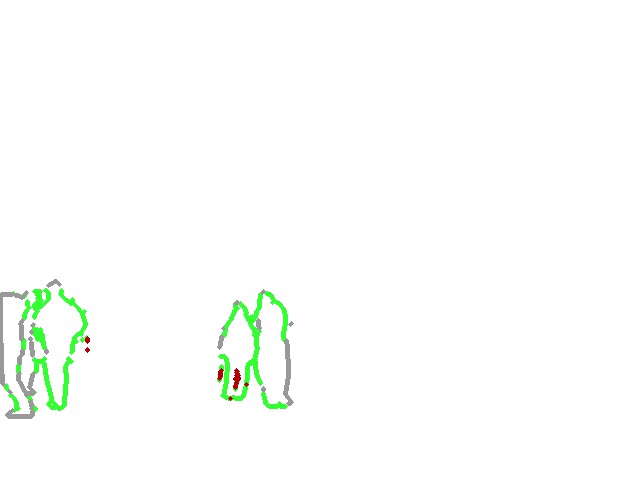} & \includegraphics[width=0.14\textwidth]{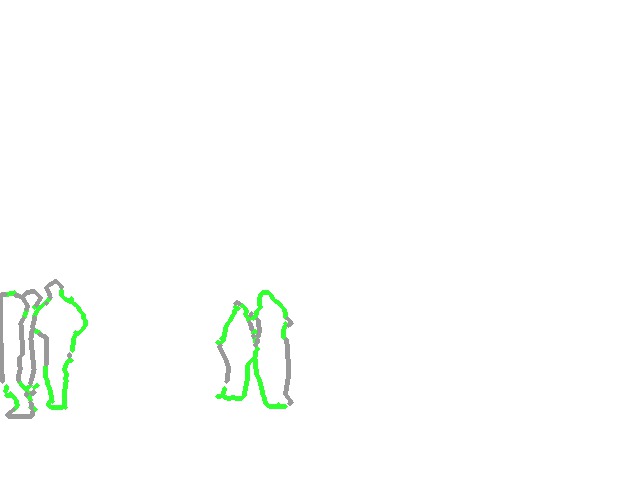}\tabularnewline
\hline 
\includegraphics[width=0.14\textwidth]{figures/sup_material/coco_results/COCO_val2014_000000175831_im} & \includegraphics[width=0.14\textwidth]{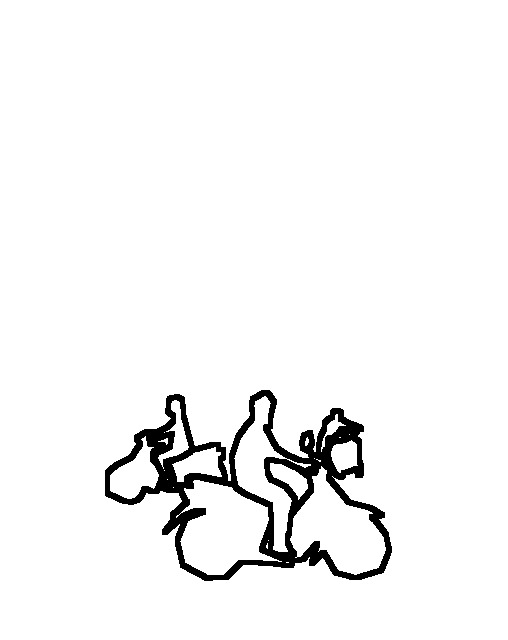} & \includegraphics[width=0.14\textwidth]{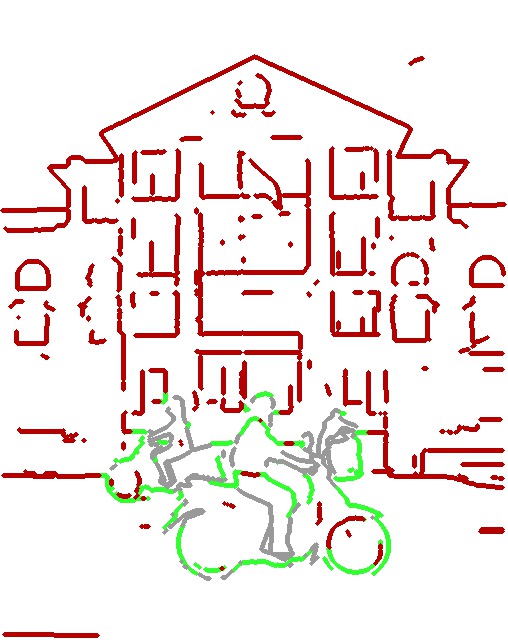} & \includegraphics[width=0.14\textwidth]{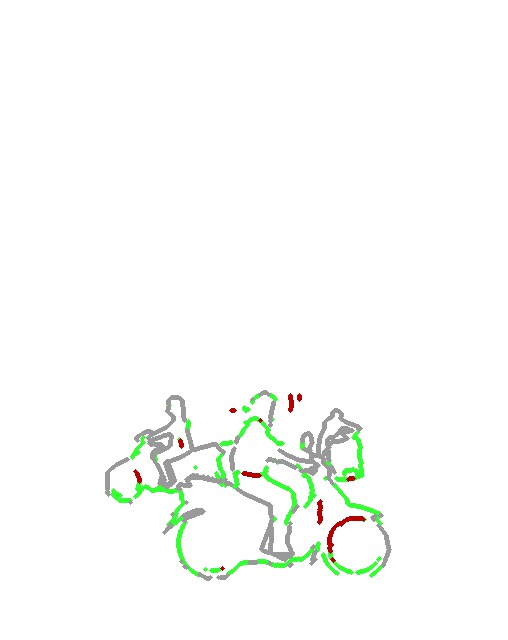} & \includegraphics[width=0.14\textwidth]{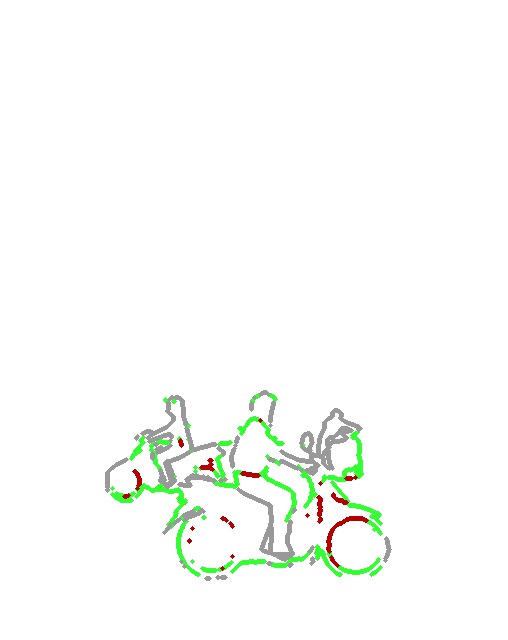} & \includegraphics[width=0.14\textwidth]{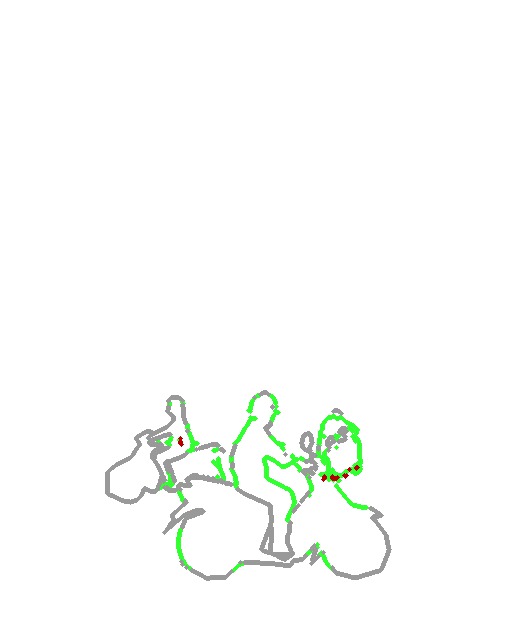} & \includegraphics[width=0.14\textwidth]{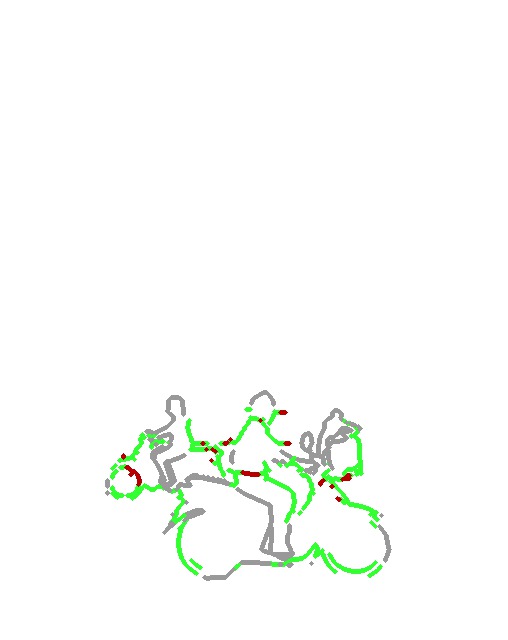}\tabularnewline
\hline 
\includegraphics[width=0.14\textwidth]{figures/sup_material/coco_results/COCO_val2014_000000188945_im} & \includegraphics[width=0.14\textwidth]{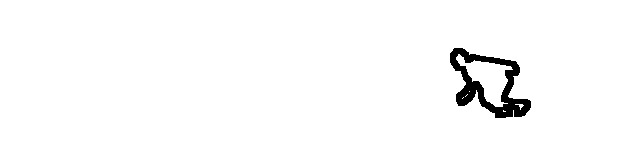} & \includegraphics[width=0.14\textwidth]{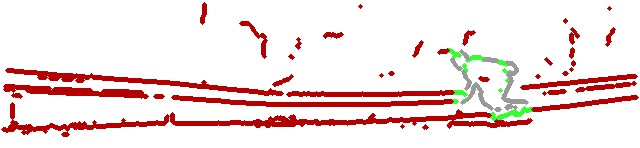} & \includegraphics[width=0.14\textwidth]{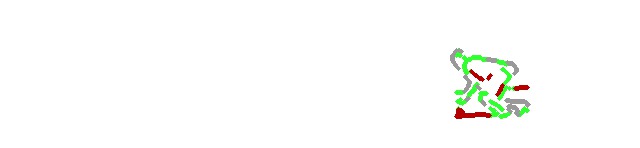} & \includegraphics[width=0.14\textwidth]{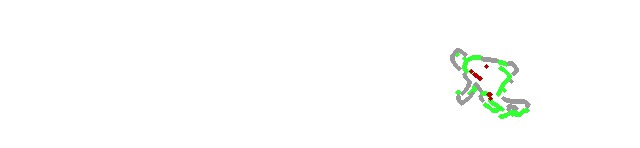} & \includegraphics[width=0.14\textwidth]{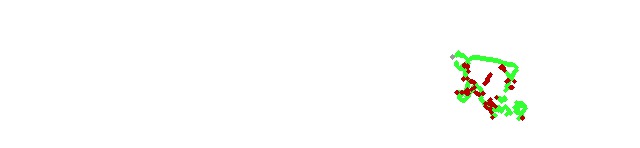} & \includegraphics[width=0.14\textwidth]{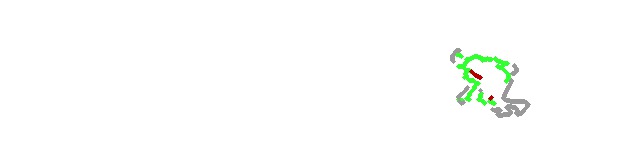}\tabularnewline
\hline 
\includegraphics[width=0.14\textwidth]{figures/sup_material/coco_results/COCO_val2014_000000224567_im} & \includegraphics[width=0.14\textwidth]{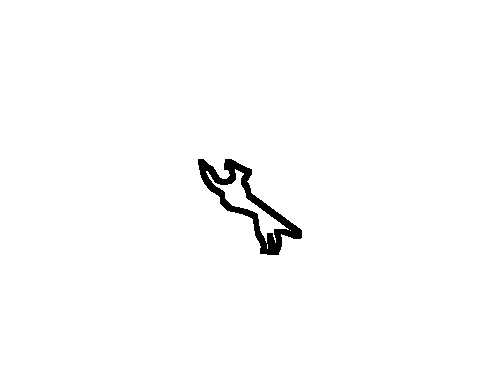} & \includegraphics[width=0.14\textwidth]{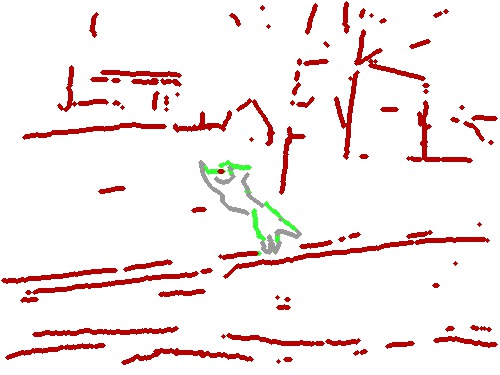} & \includegraphics[width=0.14\textwidth]{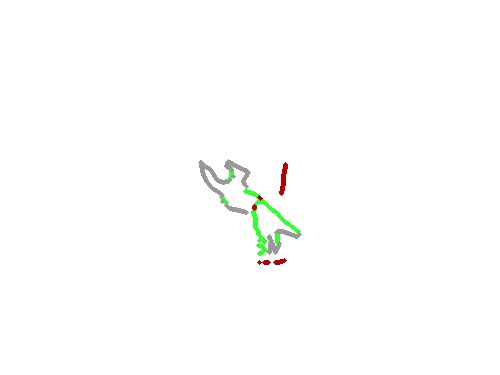} & \includegraphics[width=0.14\textwidth]{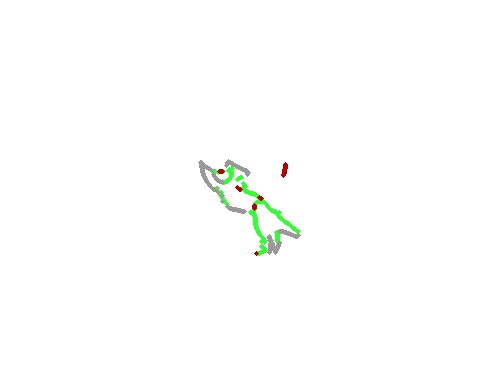} & \includegraphics[width=0.14\textwidth]{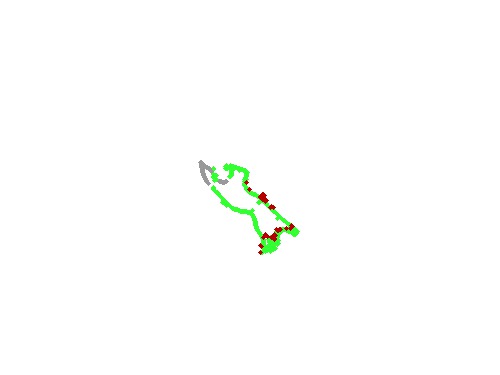} & \includegraphics[width=0.14\textwidth]{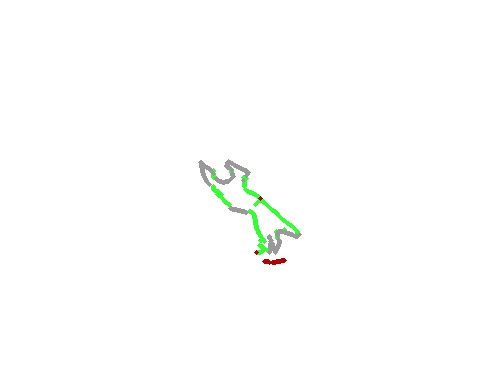}\tabularnewline
\hline 
\includegraphics[width=0.14\textwidth]{figures/sup_material/coco_results/COCO_val2014_000000416837_im} & \includegraphics[width=0.14\textwidth]{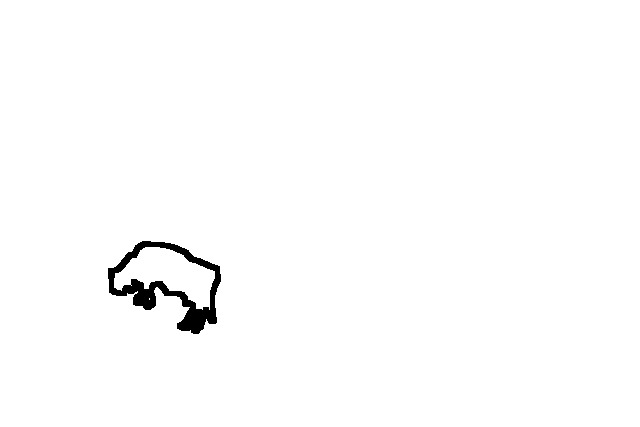} & \includegraphics[width=0.14\textwidth]{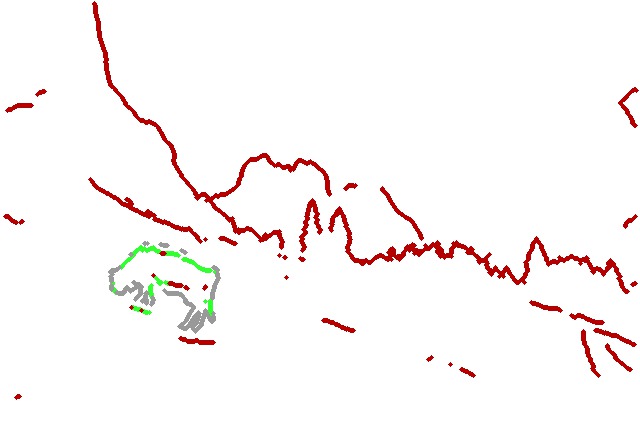} & \includegraphics[width=0.14\textwidth]{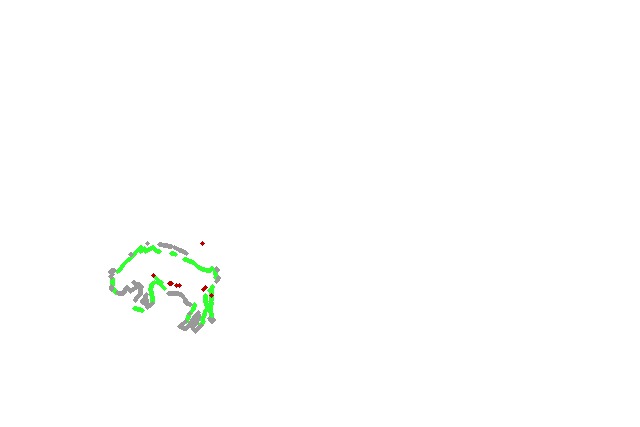} & \includegraphics[width=0.14\textwidth]{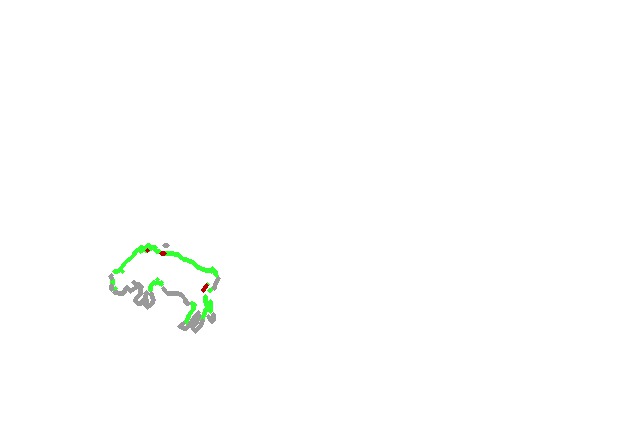} & \includegraphics[width=0.14\textwidth]{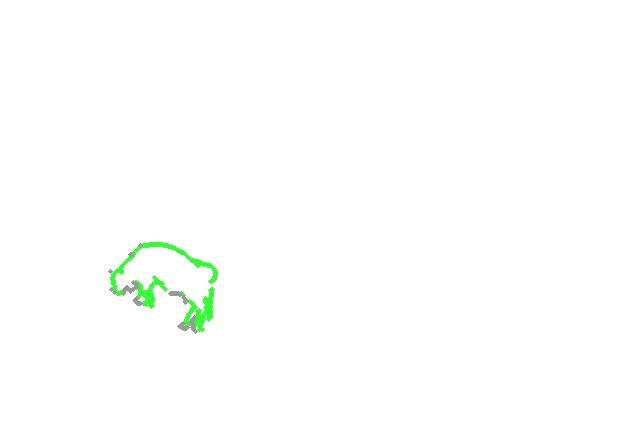} & \includegraphics[width=0.14\textwidth]{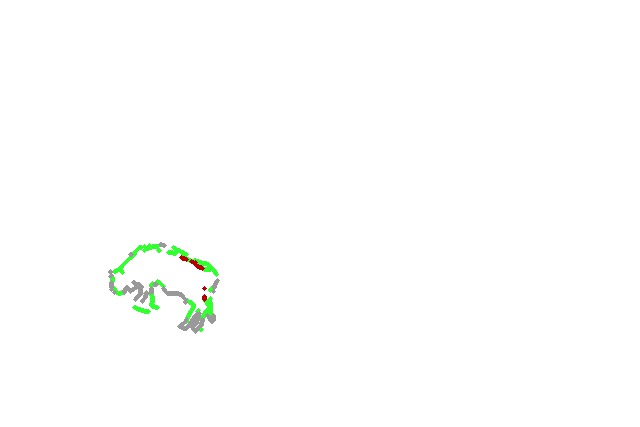}\tabularnewline
\hline 
\multicolumn{1}{c}{\selectlanguage{english}%
{\small{}\rule{0pt}{2.6ex}}\foreignlanguage{british}{{\small{}Image}}\selectlanguage{british}%
} & \multicolumn{1}{c}{{\small{}Ground truth}} & \multicolumn{1}{c}{{\small{}SE(BSDS)}} & \multicolumn{1}{c}{{\small{}$\mbox{Det.}\negmedspace+\negmedspace\mbox{SE}\left(\mbox{COCO}\right)$}} & \multicolumn{1}{c}{{\small{}$\mbox{Det.}\negmedspace+\negmedspace\mbox{SE}\left(\mathbf{weak}\right)$}} & \multicolumn{1}{c}{{\small{}$\mbox{Det.}\negthickspace+\negthickspace\mbox{HED}\left(\textrm{COCO}\right)$}} & \multicolumn{1}{c}{{\small{}$\mbox{Det.}\negthickspace+\negthickspace\mbox{HED}\left(\mathbf{weak}\right)$}}\tabularnewline
\end{tabular}

\protect\caption{\label{fig:Qualitative-results-coco}Qualitative results on COCO.
$\left(\cdot\right)$ denotes the data used for training. Red/green
indicate false/true positive pixels, grey is missing recall. All methods
are shown at $50\%$ recall. {\small{}$\mbox{Det.}\negmedspace+\negmedspace\mbox{SE}\left(\mathbf{weak}\right)$}
denotes the model {\small{}$\mbox{Det.}\negmedspace+\negmedspace\mbox{SE}\left(\mbox{SeSe}_{+}\cap\mbox{BBs}\right)$}
and {\small{}$\mbox{Det.}\negthickspace+\negthickspace\mbox{HED}\left(\mathbf{weak}\right)$}
denotes {\small{}$\mbox{Det.}\negthickspace+\negthickspace\mbox{HED}\left(\textrm{cons.\mbox{ S\&G}\ensuremath{\cap}BBs}\right)$}.
Object-specific boundaries differ from generic boundaries (such as
the ones detected by SE(BSDS)). By using an object detector we can
suppress non-object boundaries and focus boundary detection on the
classes of interest. The proposed weakly supervised techniques allow
to achieve high quality boundary estimates that are similar to the
ones obtained by fully supervised methods. }
}\endgroup
\arrayrulecolor{black}
\end{figure*}

Figure \ref{fig:Qualitative-results-coco} shows examples of boundary
detection on COCO. This figure complements Table 3 from the main paper.
Our proposed weak-supervision techniques achieve competitive performance
with fully supervised results for object-specific boundaries.

\section{\label{sec:Detailed-results-SBD}Detailed results for SBD}

\paragraph{Per-class curves}

\begin{figure*}
\vspace{-0.5em}
\includegraphics[bb=0bp 90bp 612bp 700bp,clip,width=0.245\textwidth,height=0.185\textheight]{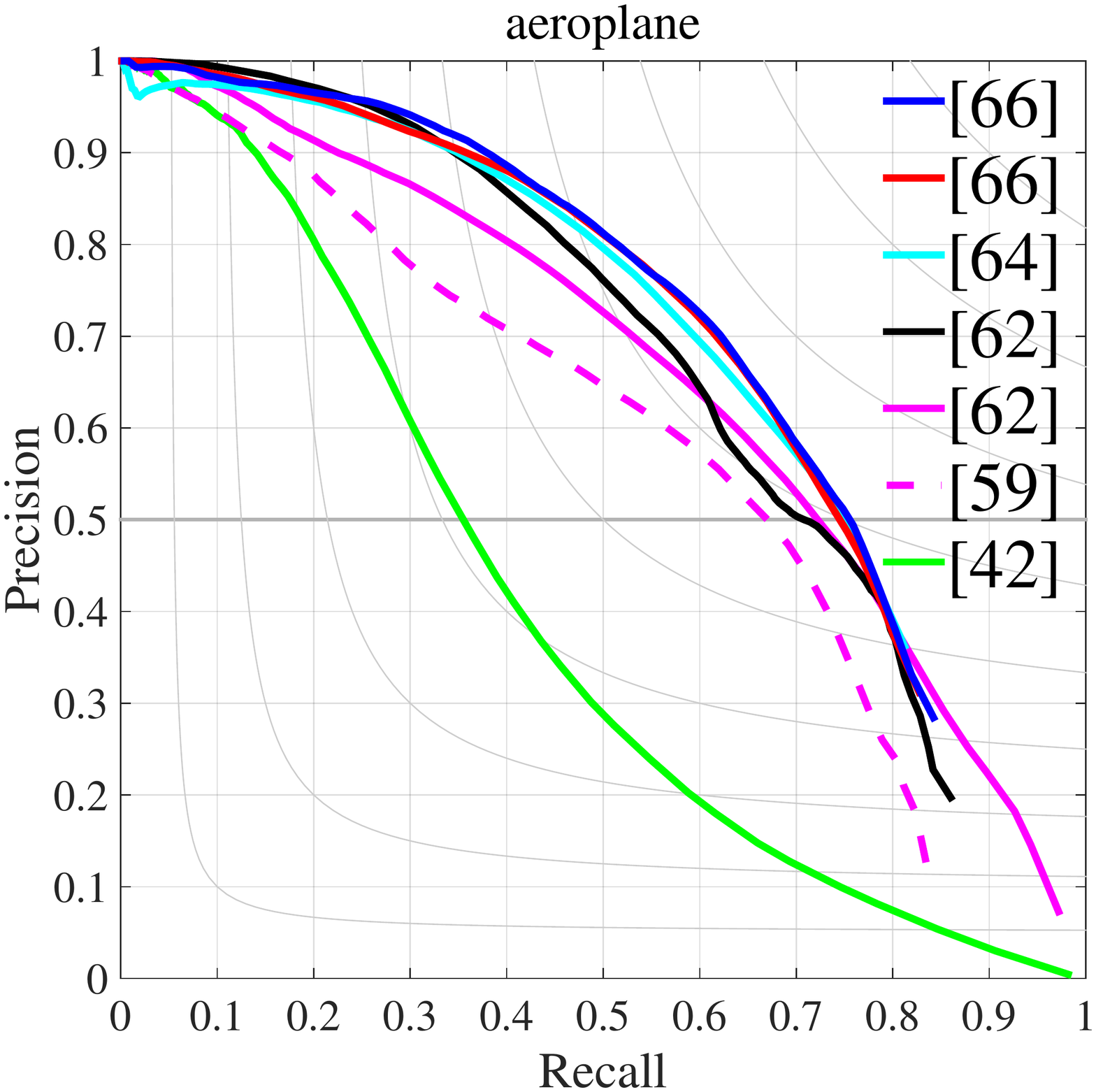}\includegraphics[bb=0bp 90bp 612bp 700bp,clip,width=0.245\textwidth,height=0.185\textheight]{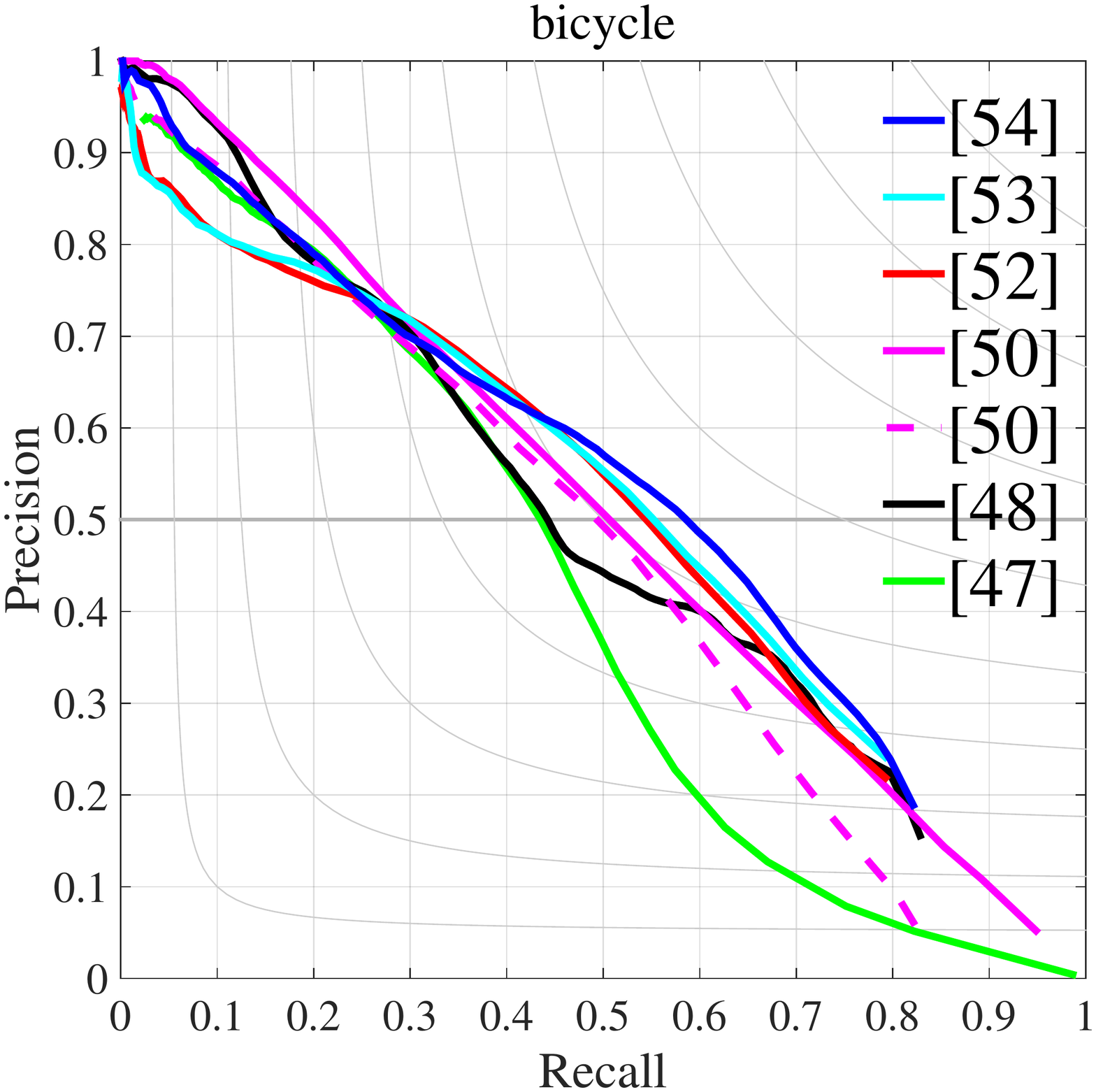}\includegraphics[bb=0bp 90bp 612bp 700bp,clip,width=0.245\textwidth,height=0.185\textheight]{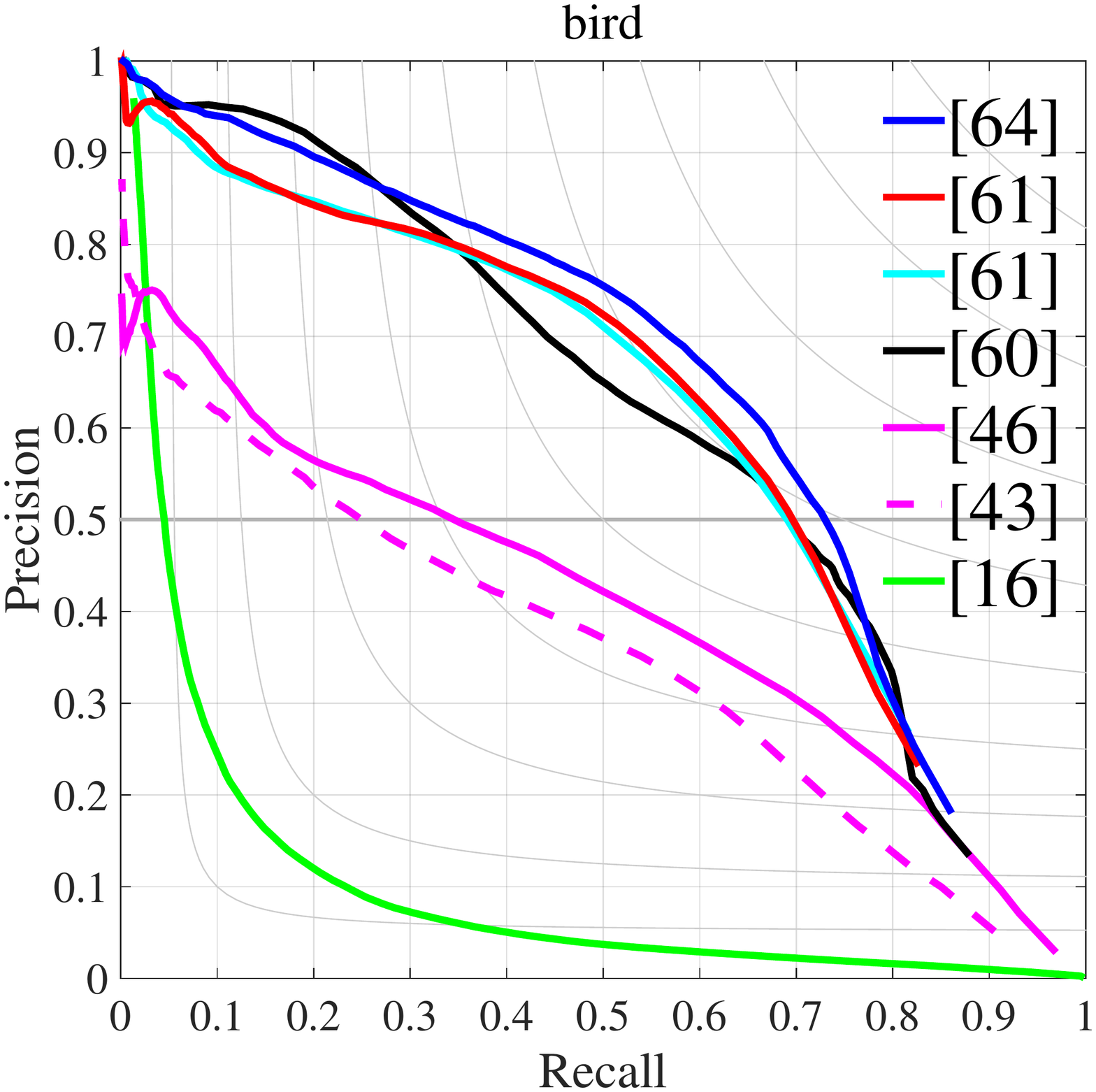}\includegraphics[bb=0bp 90bp 612bp 700bp,clip,width=0.245\textwidth,height=0.185\textheight]{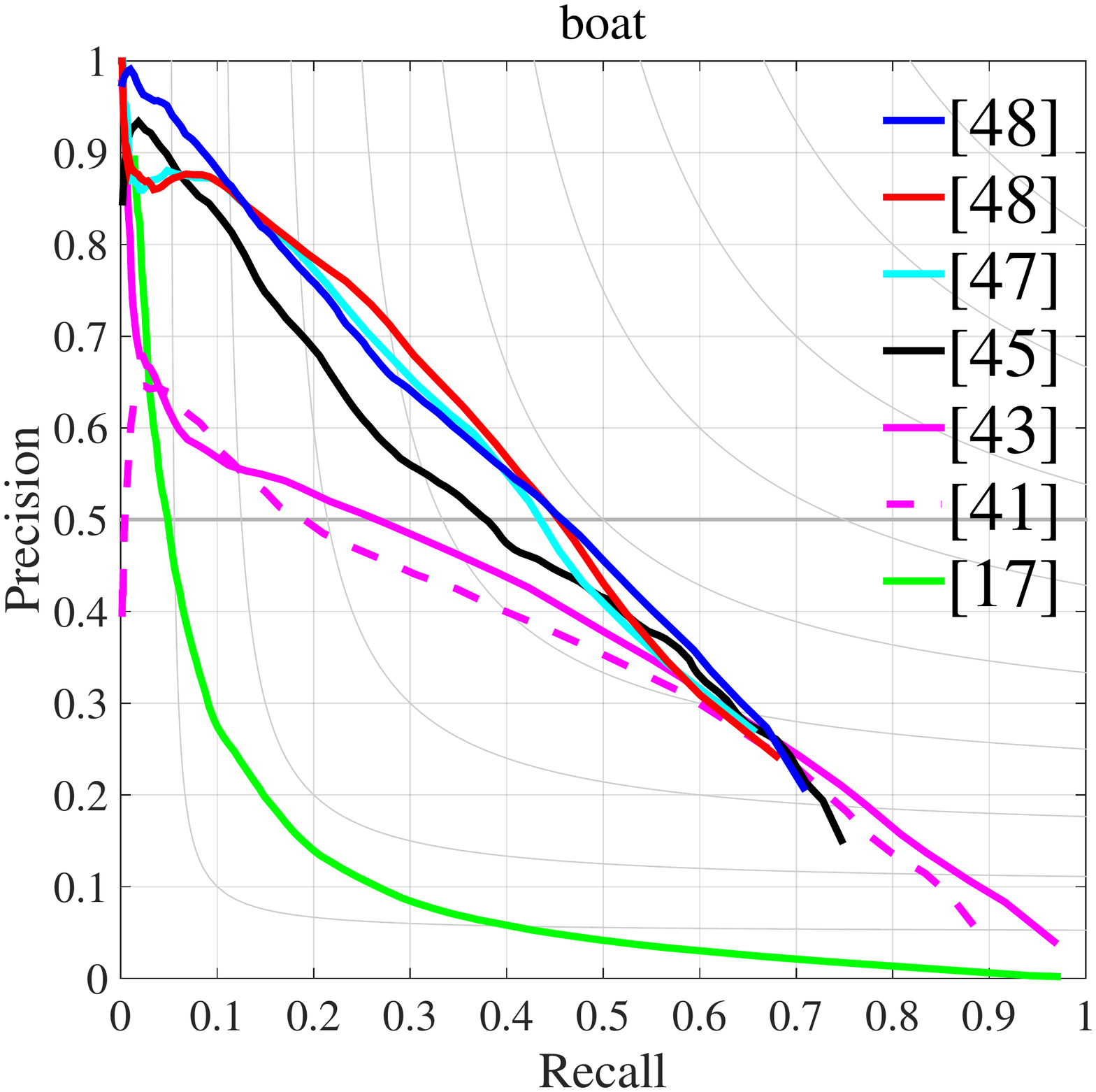}

\vfill{}

\includegraphics[bb=0bp 90bp 612bp 700bp,clip,width=0.245\textwidth,height=0.185\textheight]{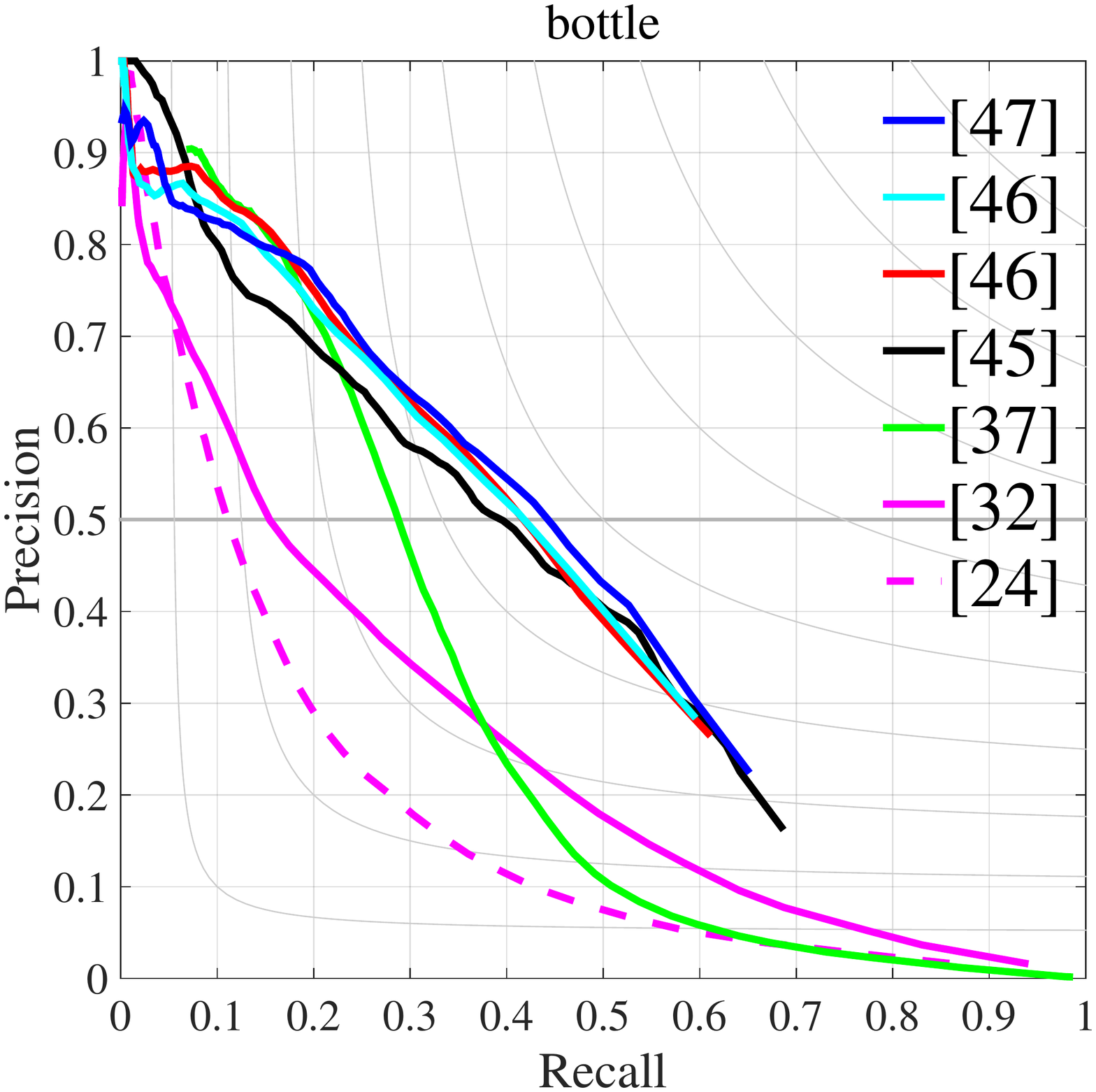}\includegraphics[bb=0bp 90bp 612bp 700bp,clip,width=0.245\textwidth,height=0.185\textheight]{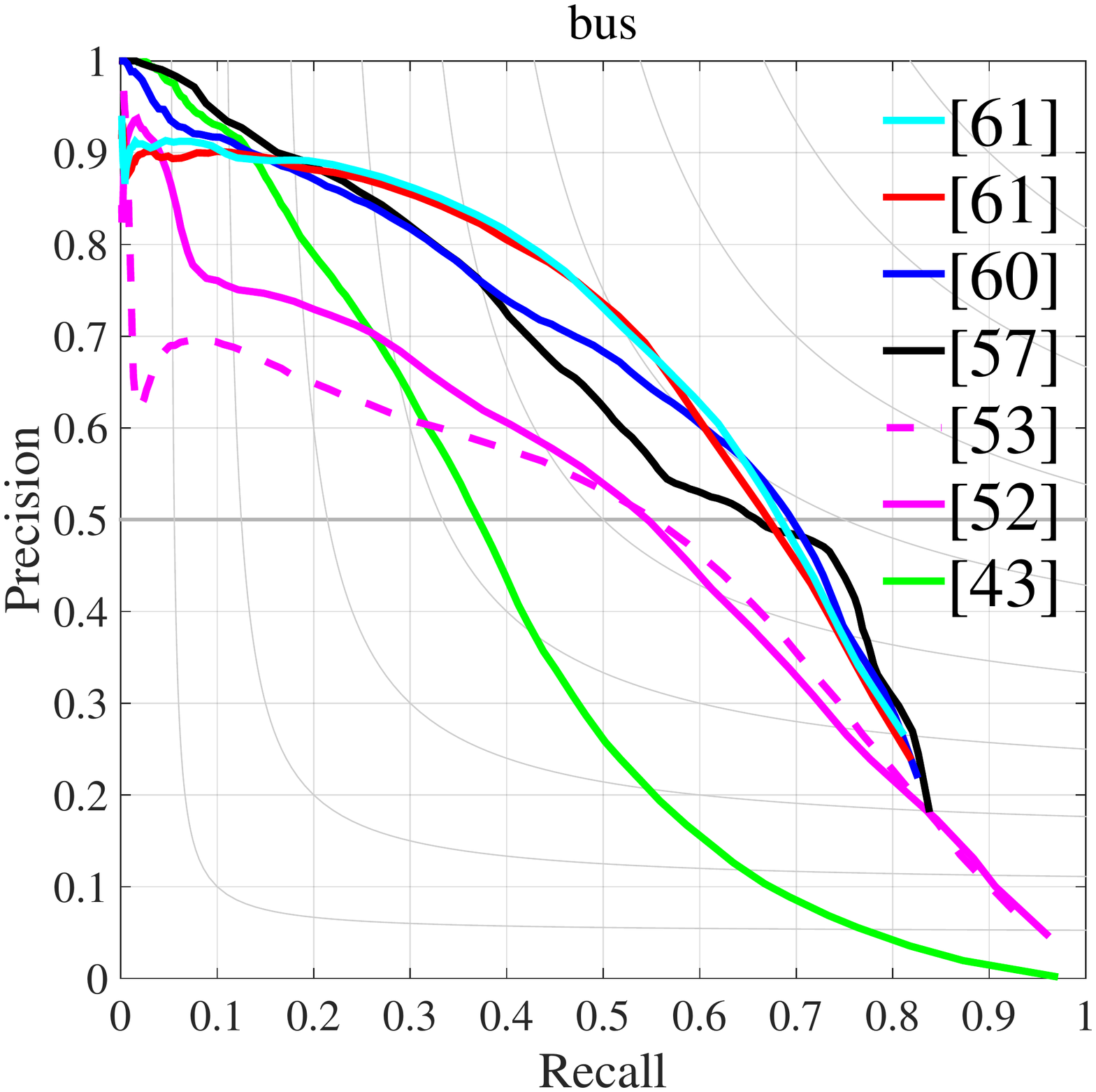}\includegraphics[bb=0bp 90bp 612bp 700bp,clip,width=0.245\textwidth,height=0.185\textheight]{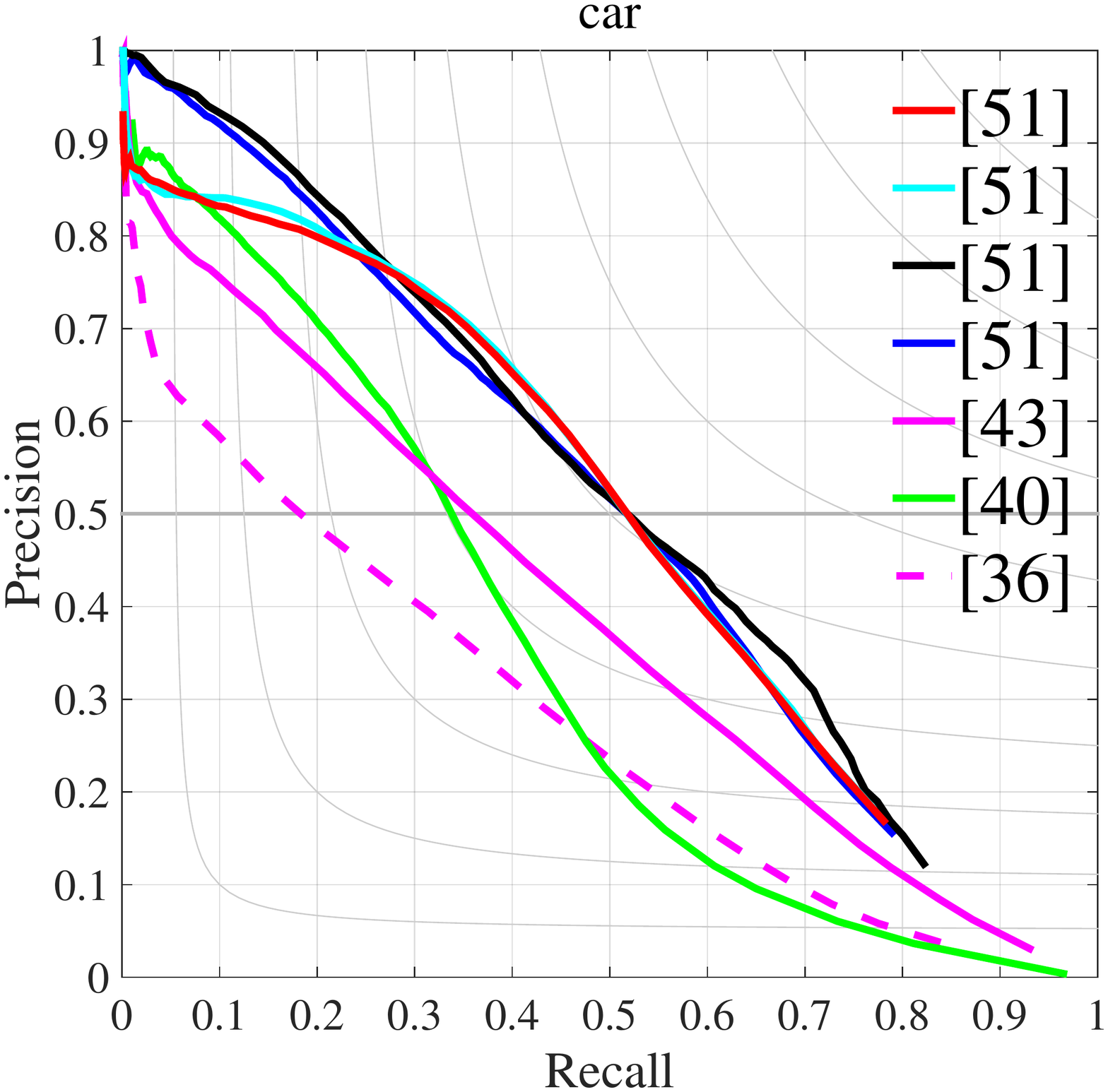}\includegraphics[bb=0bp 90bp 612bp 700bp,clip,width=0.245\textwidth,height=0.185\textheight]{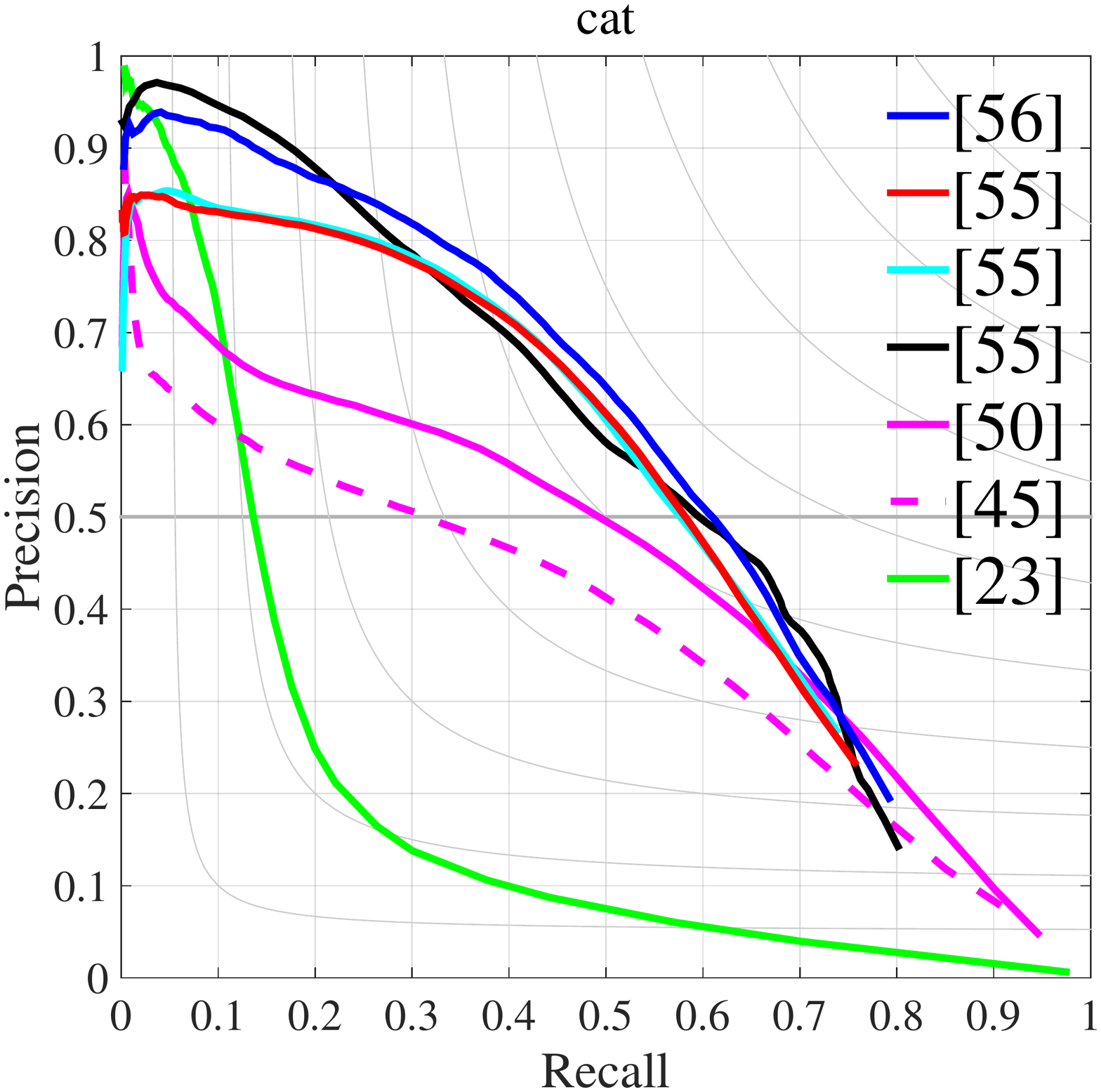}

\vfill{}

\includegraphics[bb=0bp 90bp 612bp 700bp,clip,width=0.245\textwidth,height=0.185\textheight]{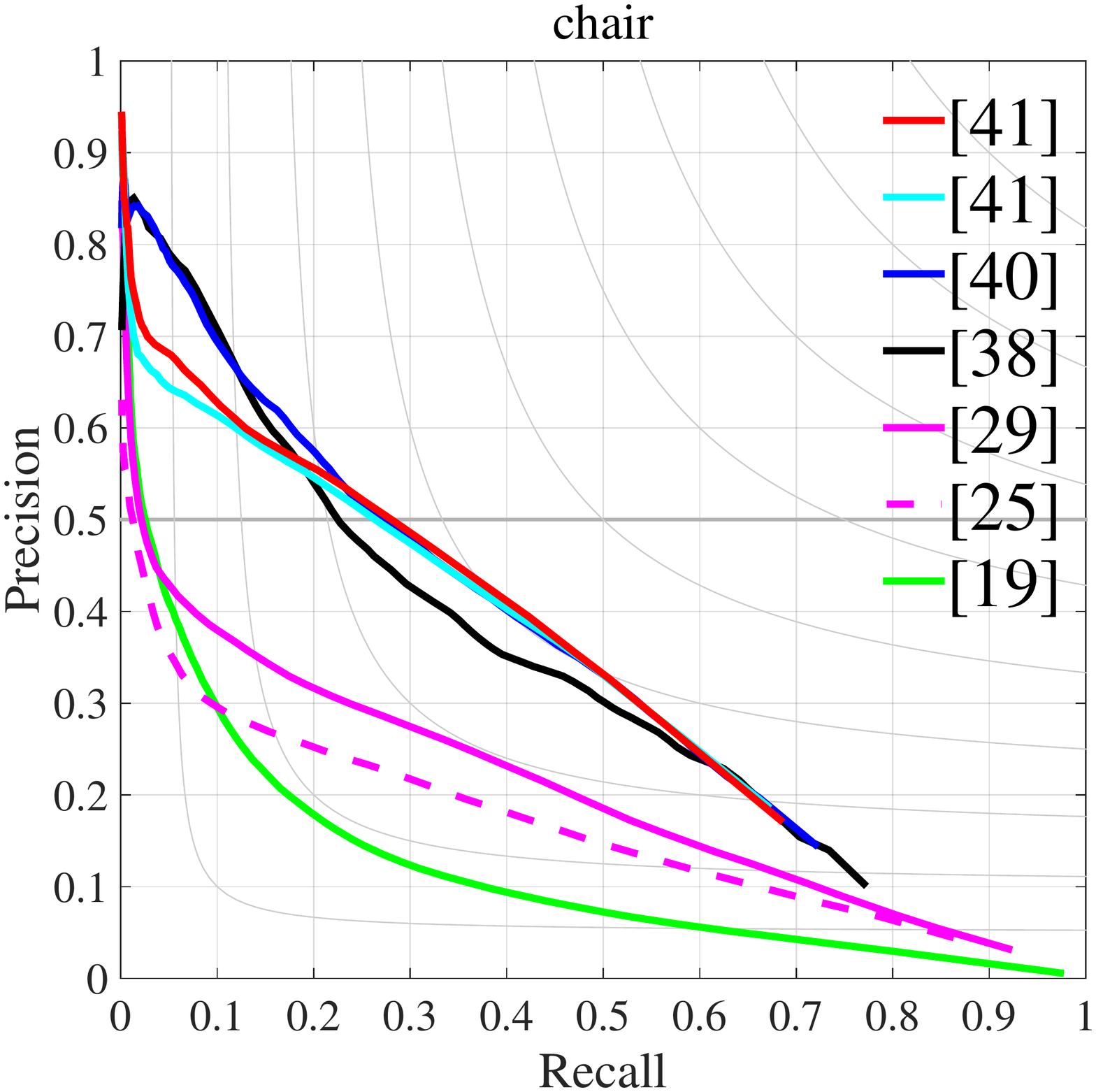}\includegraphics[bb=0bp 90bp 612bp 700bp,clip,width=0.245\textwidth,height=0.185\textheight]{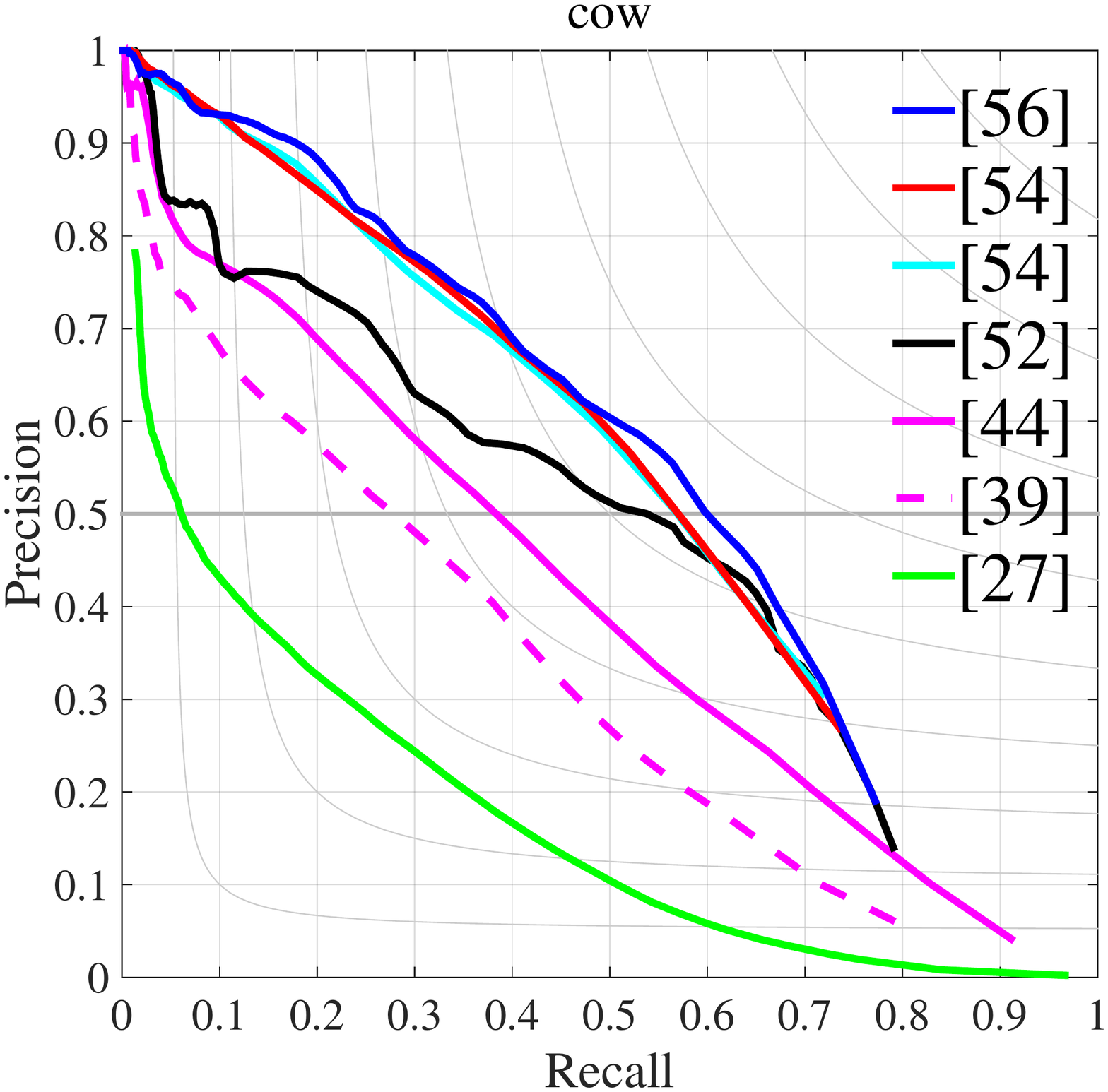}\includegraphics[bb=0bp 90bp 612bp 700bp,clip,width=0.245\textwidth,height=0.185\textheight]{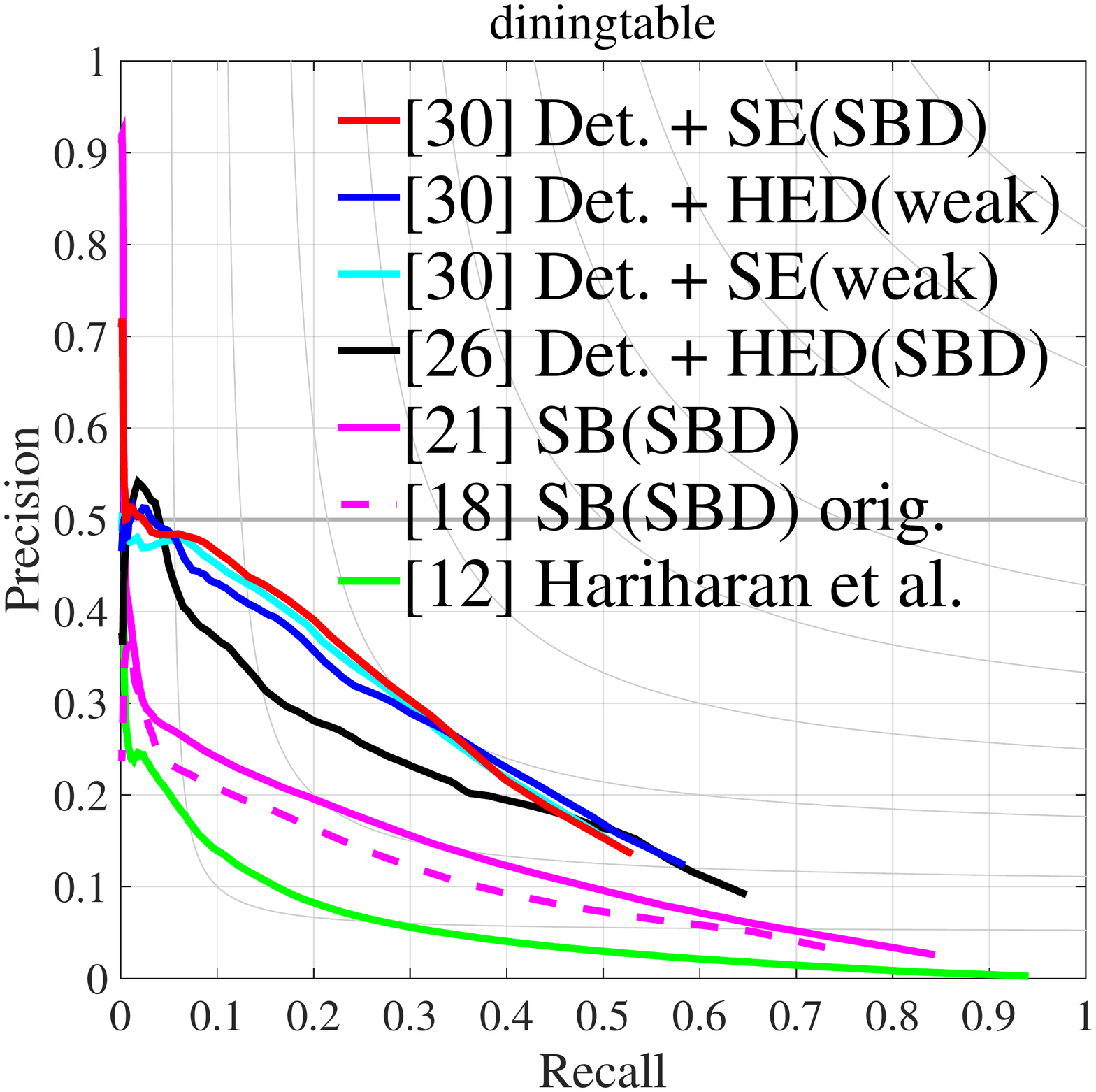}\includegraphics[bb=0bp 90bp 612bp 700bp,clip,width=0.245\textwidth,height=0.185\textheight]{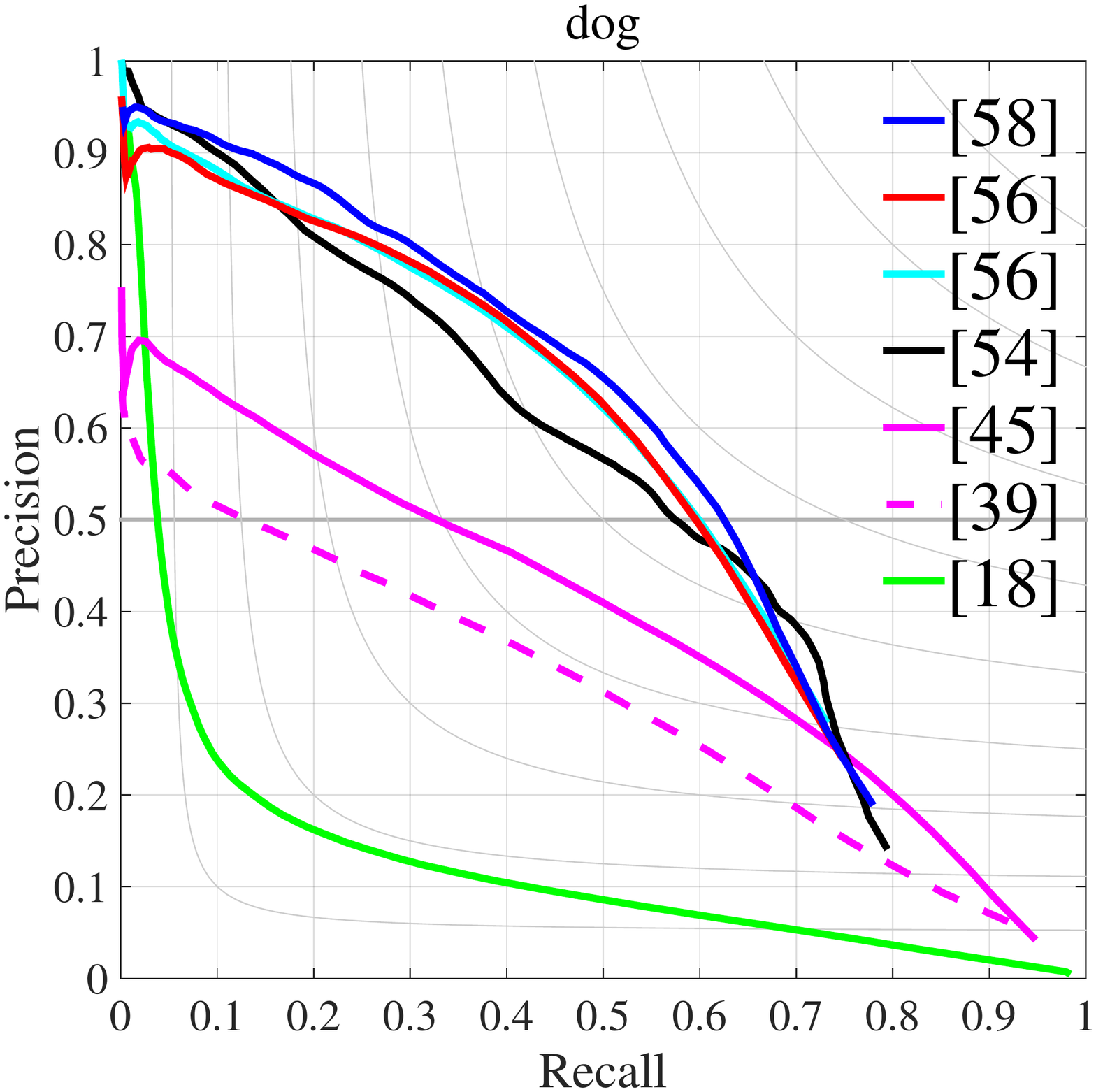}

\vfill{}

\includegraphics[bb=0bp 90bp 612bp 700bp,clip,width=0.245\textwidth,height=0.185\textheight]{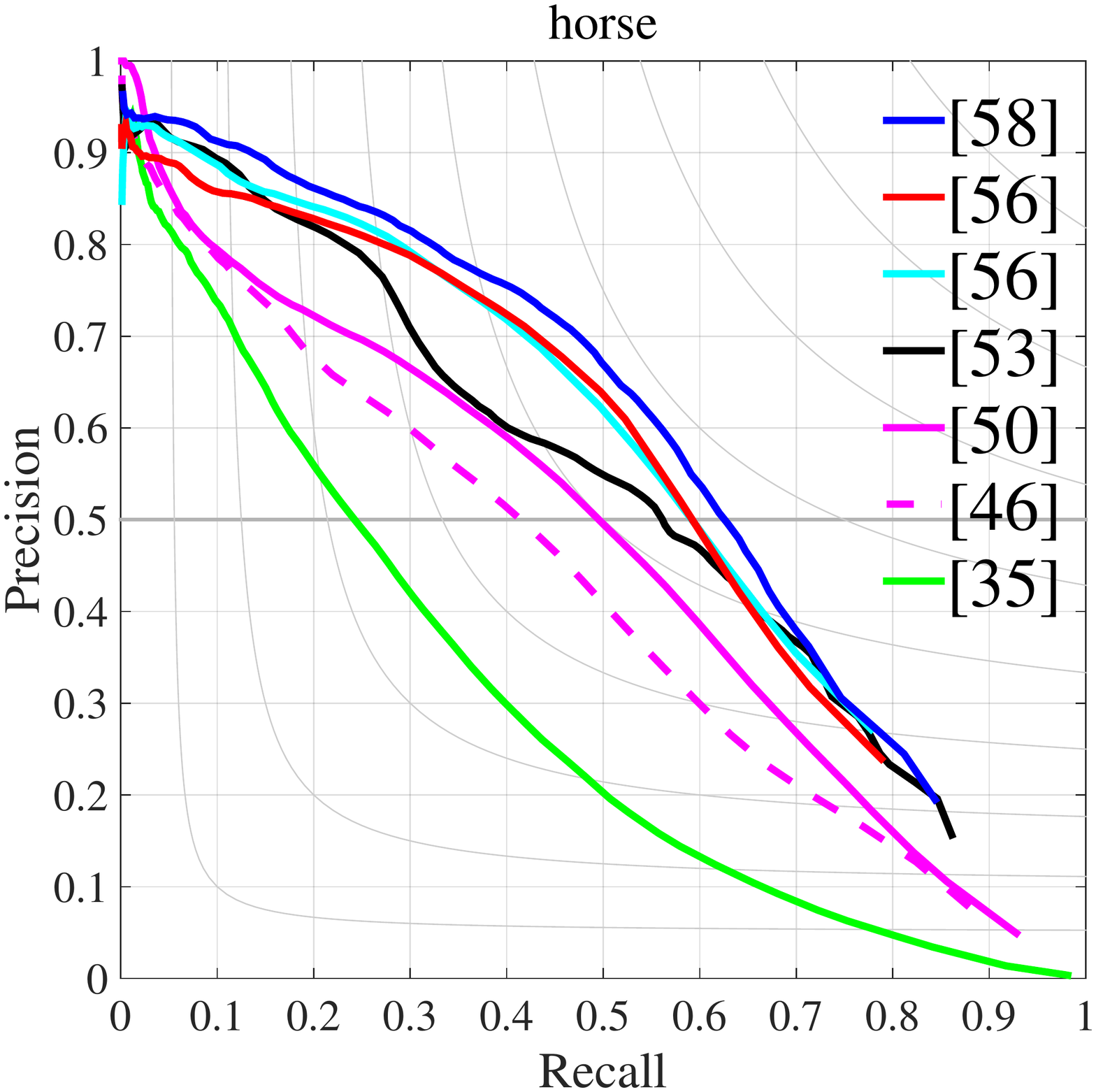}\includegraphics[bb=0bp 90bp 612bp 700bp,clip,width=0.245\textwidth,height=0.185\textheight]{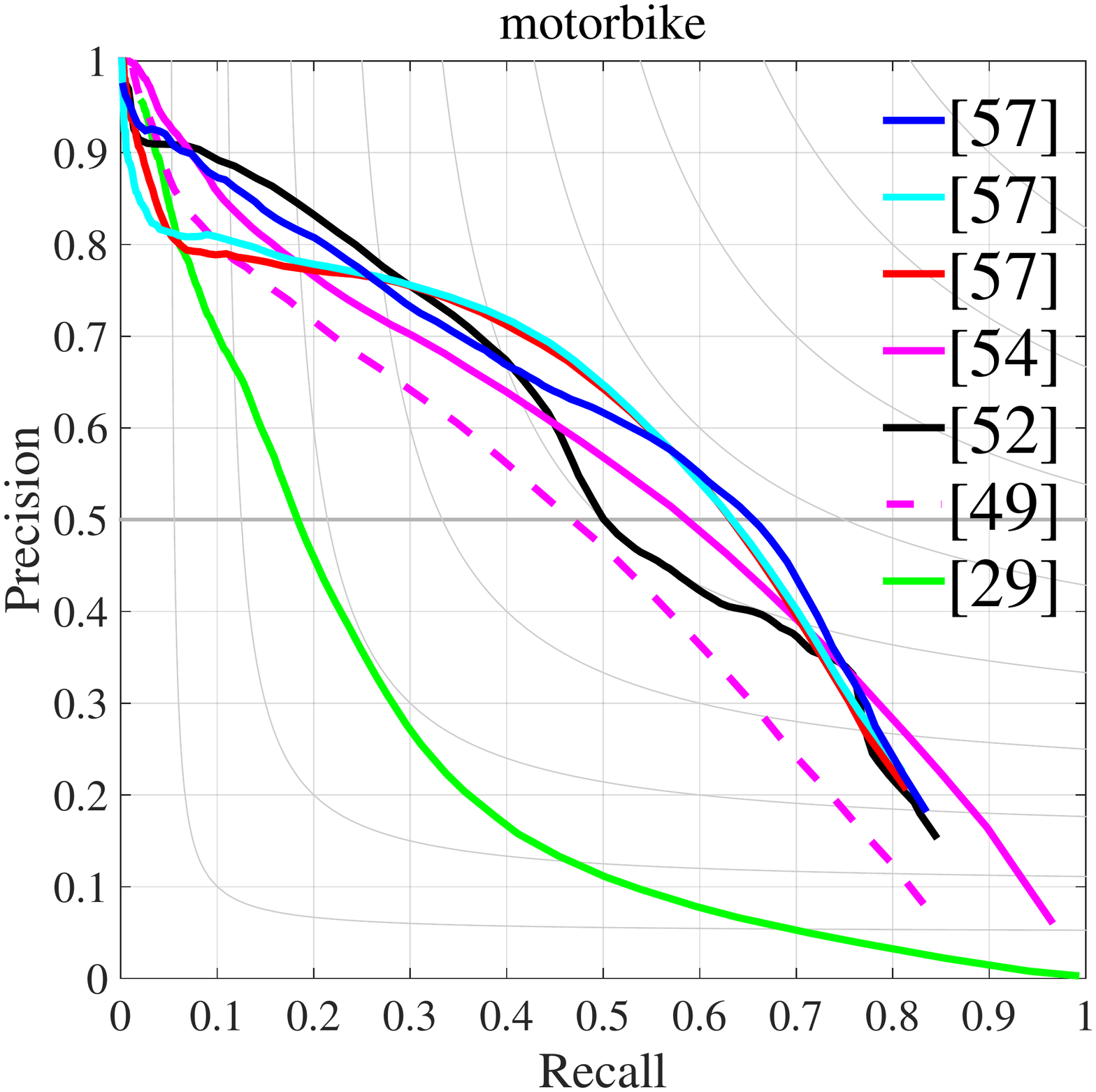}\includegraphics[bb=0bp 90bp 612bp 700bp,clip,width=0.245\textwidth,height=0.185\textheight]{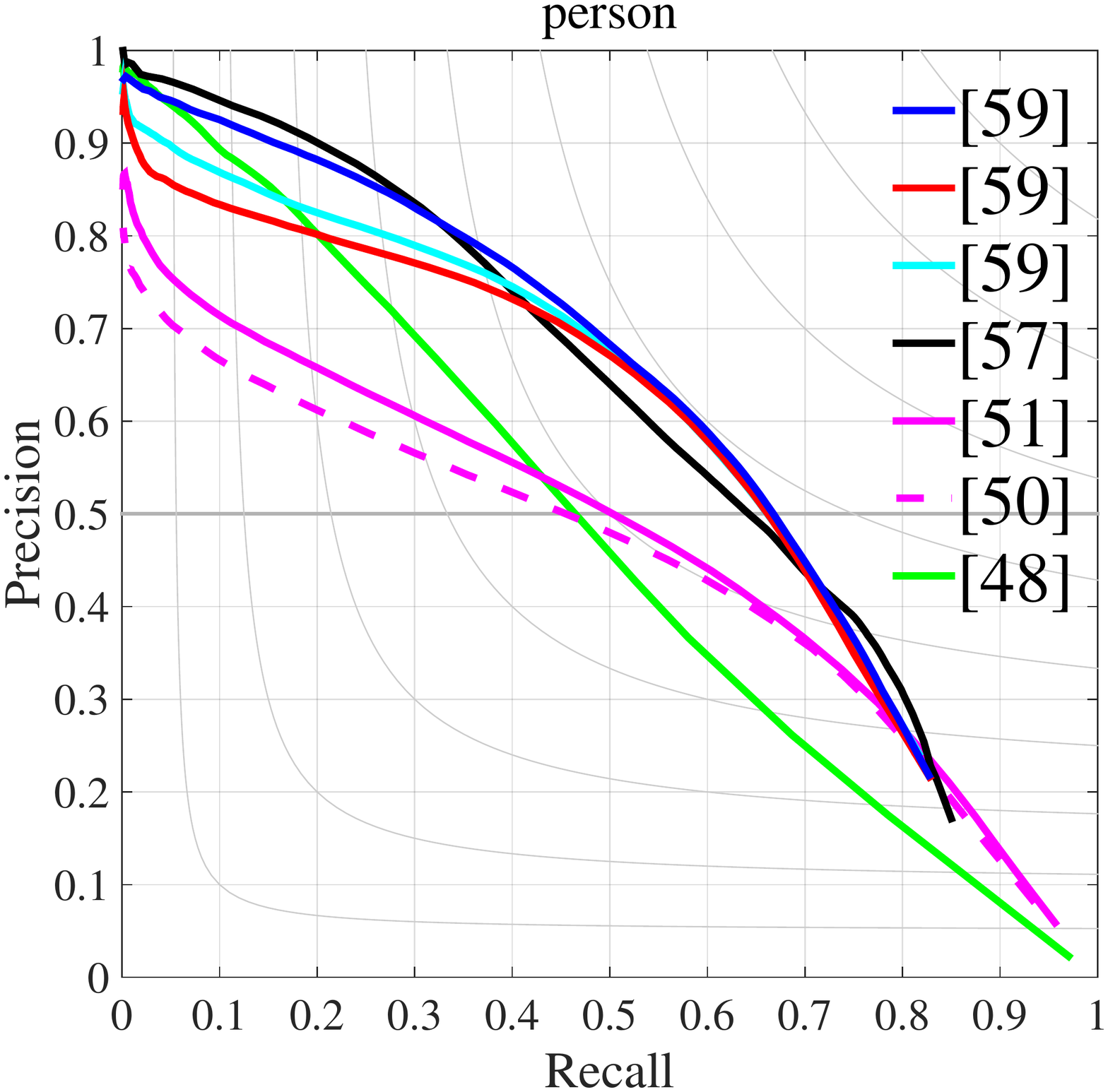}\includegraphics[bb=0bp 90bp 612bp 700bp,clip,width=0.245\textwidth,height=0.185\textheight]{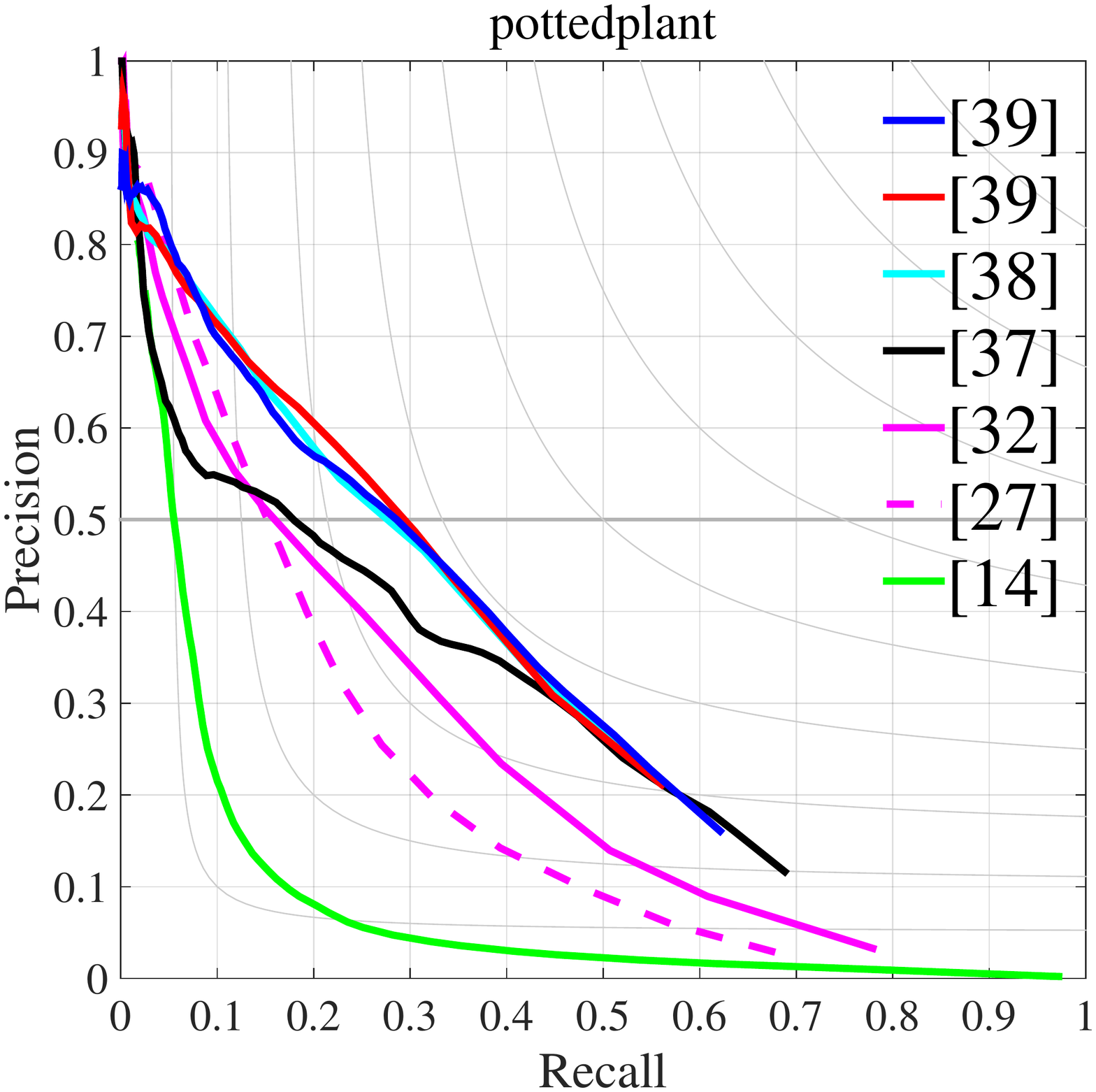}

\vfill{}

\includegraphics[bb=0bp 90bp 612bp 700bp,clip,width=0.245\textwidth,height=0.185\textheight]{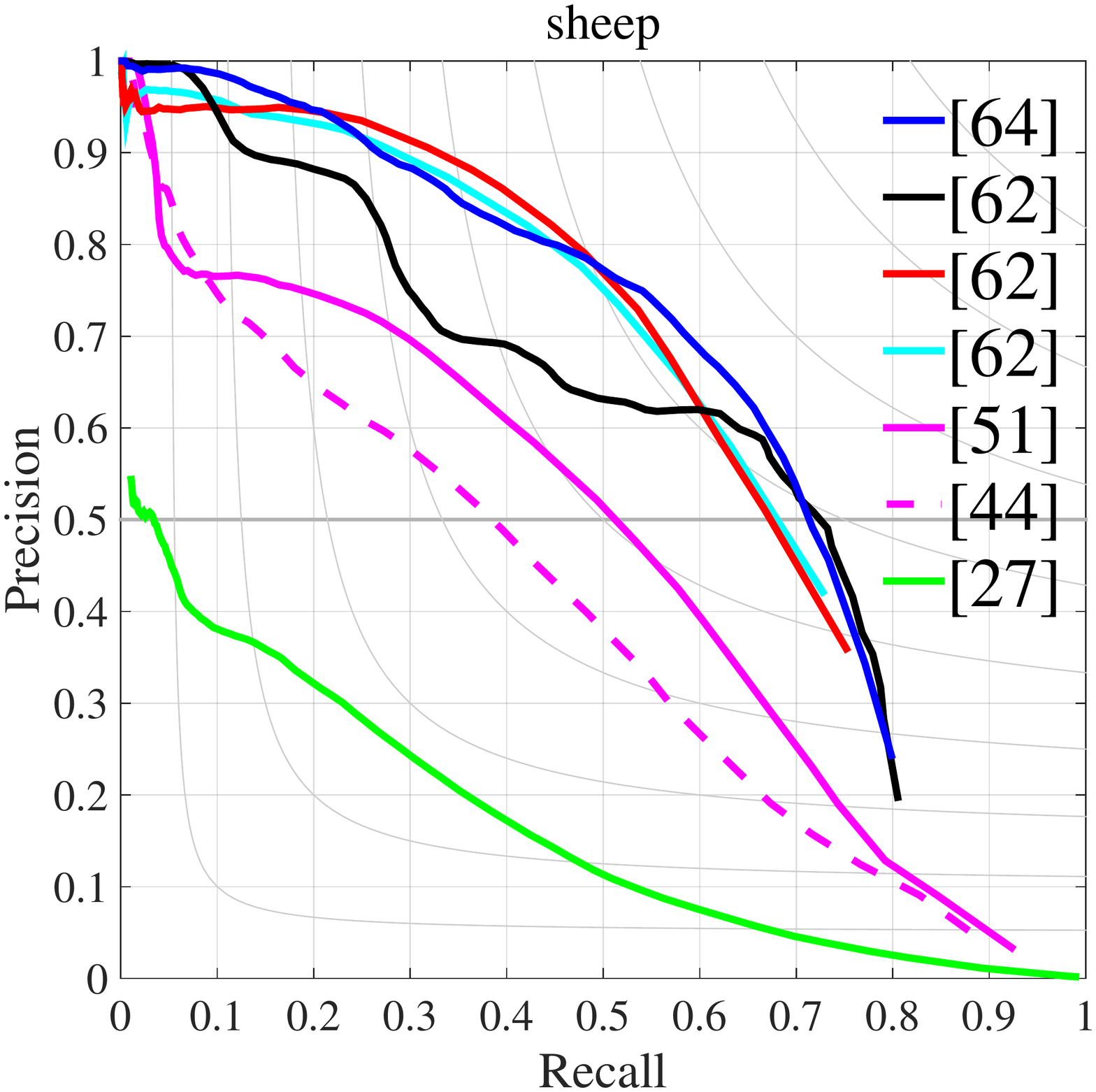}\includegraphics[bb=0bp 90bp 612bp 700bp,clip,width=0.245\textwidth,height=0.185\textheight]{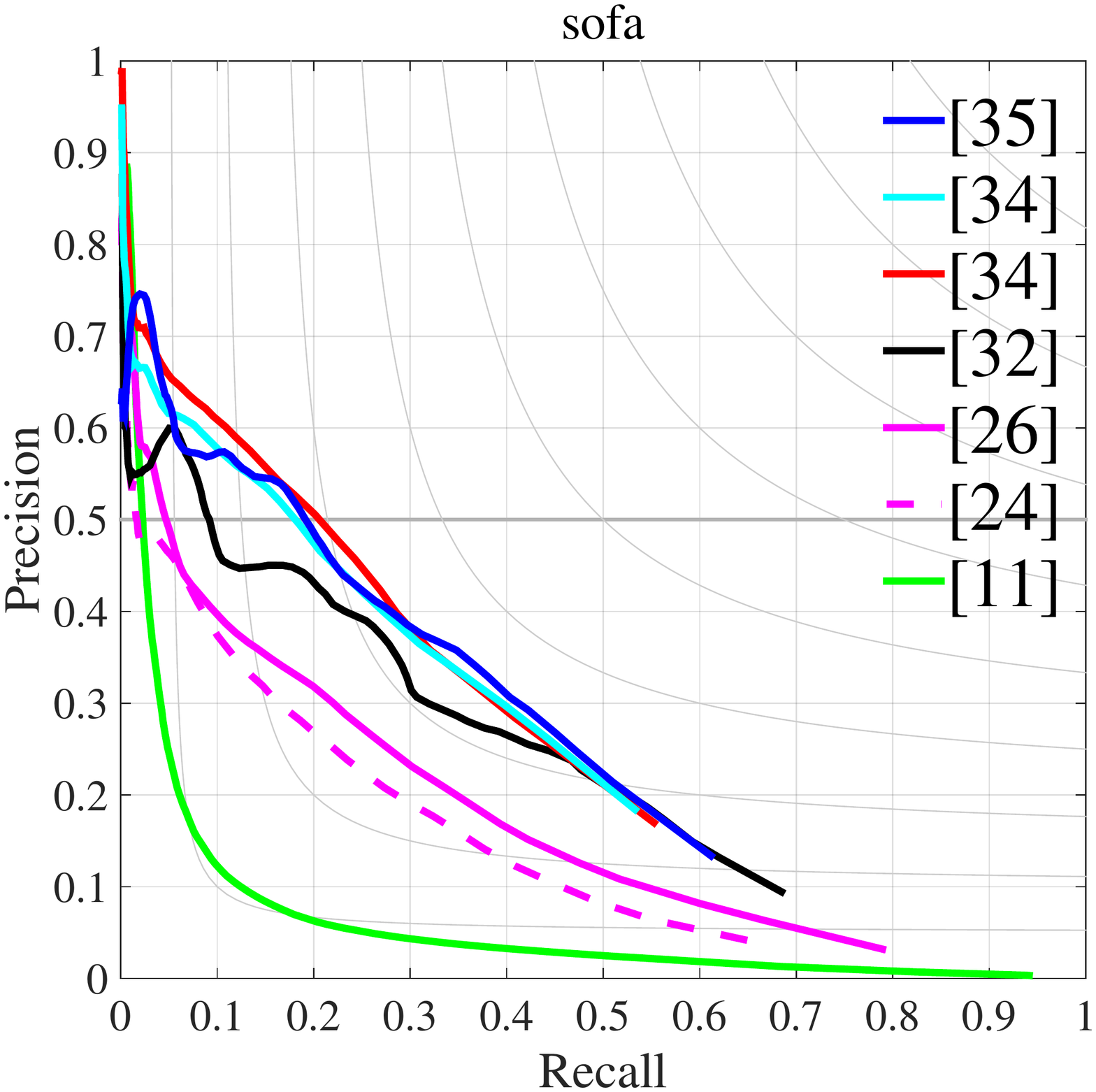}\includegraphics[bb=0bp 90bp 612bp 700bp,clip,width=0.245\textwidth,height=0.185\textheight]{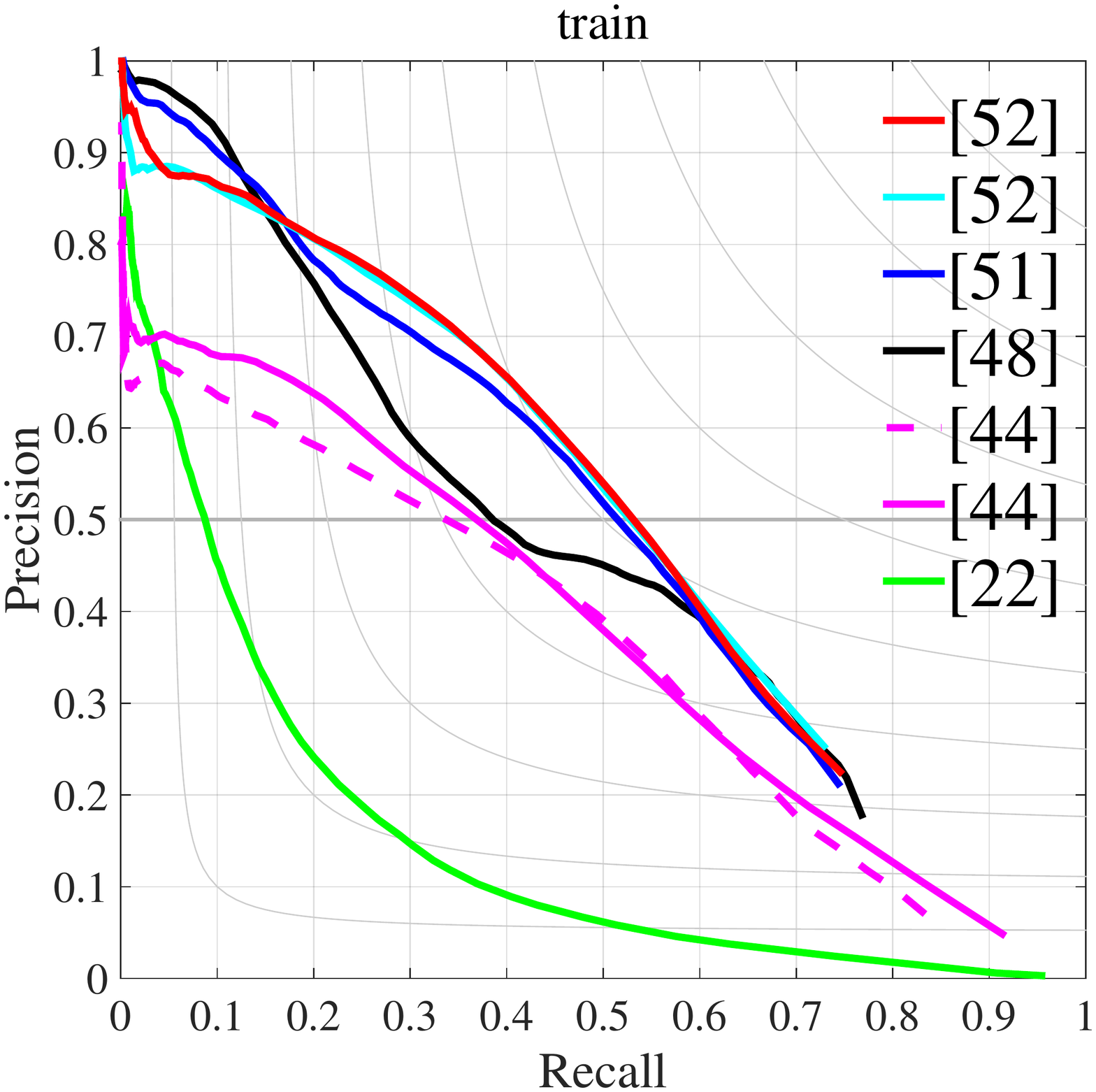}\includegraphics[bb=0bp 90bp 612bp 700bp,clip,width=0.245\textwidth,height=0.185\textheight]{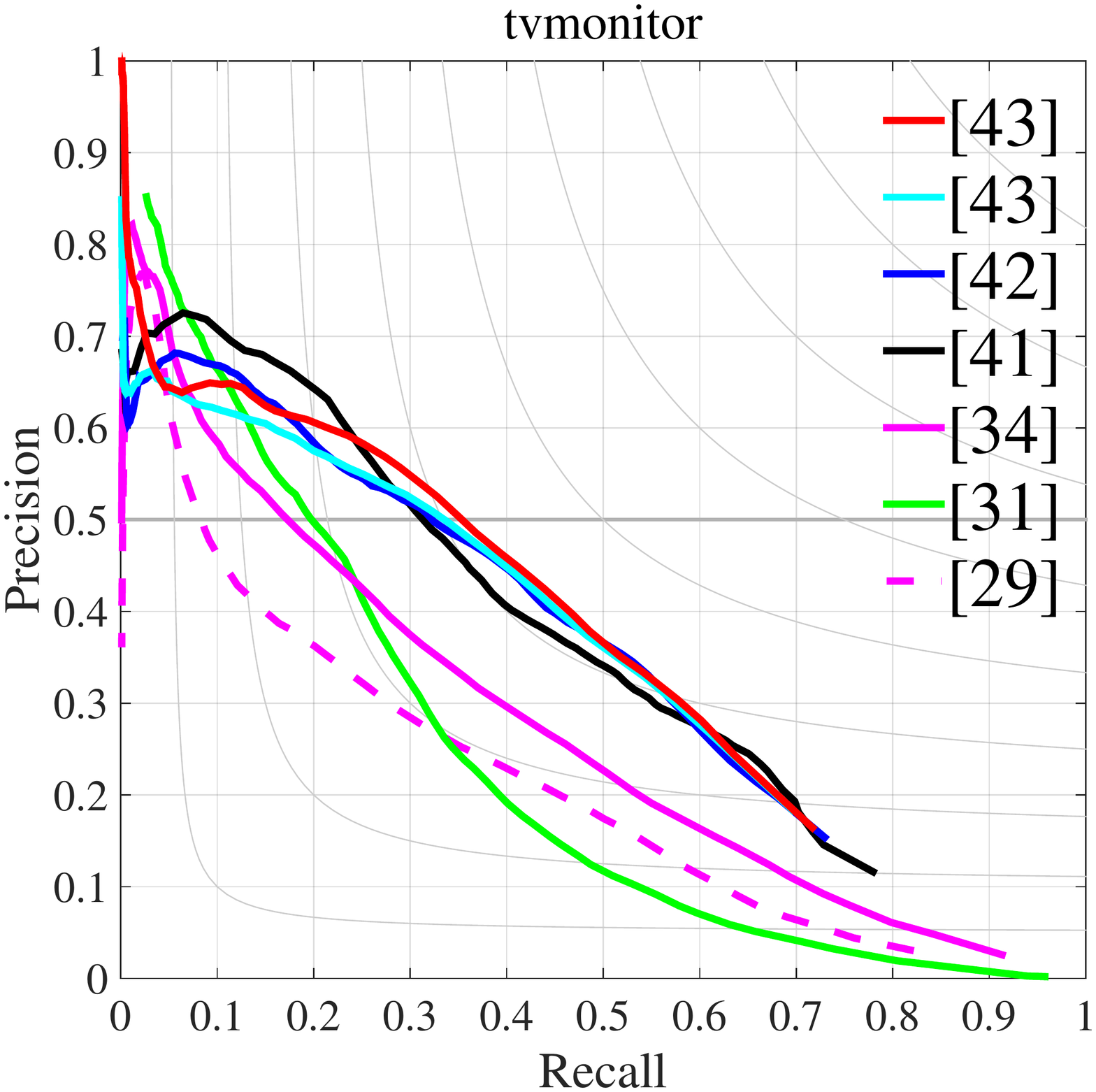}\protect\caption{\label{fig:SBD-boundary-PR}SBD boundary PR curves per class.$\left(\cdot\right)$
denotes the data used for training. Legend indicates AP numbers. {\small{}$\mbox{Det.}\negmedspace+\negmedspace\mbox{SE}\left(\mathbf{weak}\right)$}
denotes the model {\small{}$\mbox{Det.}\negmedspace+\negmedspace\mbox{SE}\left(\mbox{MCG}_{+}\cap\mbox{BBs}\right)$}
and {\small{}$\mbox{Det.}\negmedspace+\negmedspace\mbox{HED}\left(\mathbf{weak}\right)$}
denotes {\small{}$\mbox{Det.}\negthickspace+\negthickspace\mbox{HED}\left(\textrm{cons.\mbox{ S\&G}\ensuremath{\cap}BBs}\right)$}.
For all classes our weakly supervised results are on par to the fully
supervised ones.}
\end{figure*}

Figure \ref{fig:SBD-boundary-PR} shows the per class performance
of the proposed weakly supervised boundary variants trained with SE
and HED on the SBD dataset \cite{Hariharan2011Iccv} (this figure
is a breakdown of Figure 9 in the main paper). In contrast to VOC
and COCO we move from object boundaries to class specific object boundaries.
We are interested in external boundaries of all annotated objects
of the specific semantic class and all internal boundaries are ignored
during evaluation following the benchmark \cite{Hariharan2011Iccv}.
As Figure \ref{fig:SBD-boundary-PR} shows our weakly supervised approach
considerably outperforms \cite{Uijlings2015Cvpr,Hariharan2011Iccv}
on all 20 classes.\\
Compared to VOC, SE and HED results are most similar between each
other because the evaluation protocol focuses on the external object
boundaries (ignoring internal object boundaries), where both methods
equally well. Compared to VOC, we also notice that $\mbox{Det.}\negthickspace+\negthickspace\mbox{HED}\left(\textrm{cons.\mbox{ S\&G}\ensuremath{\cap}BBs}\right)$
performs better than $\mbox{Det.}\negthickspace+\negthickspace\mbox{HED}\left(\mbox{SBD}\right)$,
we attribute this to the ``consensus'' aspect of our generated annotations
(see BSDS results, Section 4 of main paper).

\section{\label{sec:SBD-boundaries-examples}SBD boundary detection examples}

Figure \ref{fig:Qualitative-results-sbd} shows examples of boundary
detection on the SBD dataset. As the quantitative results indicate,
qualitatively our weakly supervised results are on par to the fully
supervised ones.

\newpage{}

\newpage{}

\begin{figure*}
\vspace{-0.5em}
\begingroup{\arrayrulecolor{lightgray}
\setlength{\tabcolsep}{0pt} 
\renewcommand{\arraystretch}{0}

\begin{tabular}[b]{c|c|c||c|c||c|c}
\multicolumn{7}{c}{aeroplane\foreignlanguage{english}{\rule[-1.0ex]{0pt}{0pt}}}\tabularnewline
\hline 
\multicolumn{1}{|c|}{\includegraphics[width=0.14\textwidth]{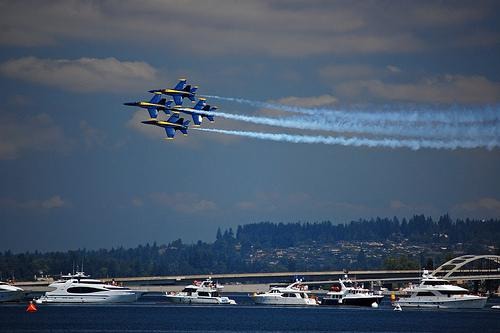}} & \includegraphics[width=0.14\textwidth]{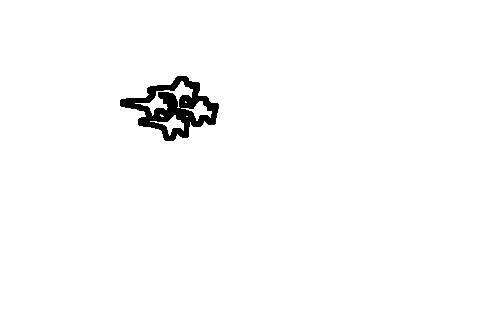} & \includegraphics[width=0.14\textwidth]{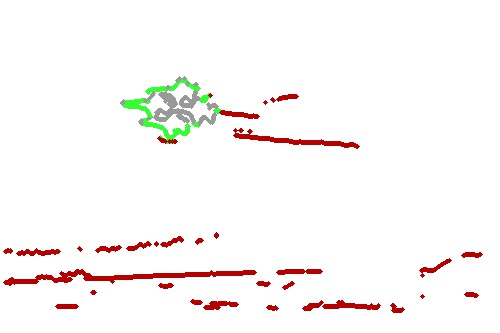} & \includegraphics[width=0.14\textwidth]{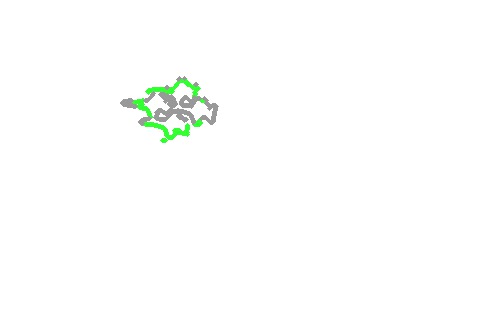} & \includegraphics[width=0.14\textwidth]{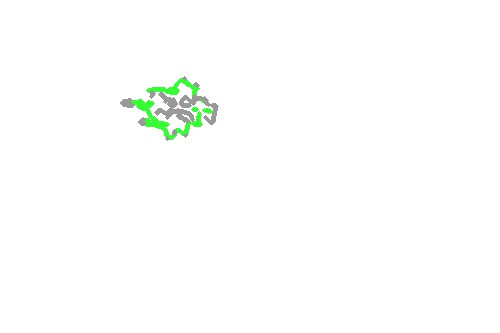} & \includegraphics[width=0.14\textwidth]{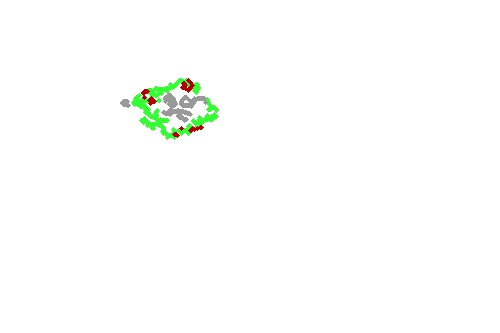} & \multicolumn{1}{c|}{\includegraphics[width=0.14\textwidth]{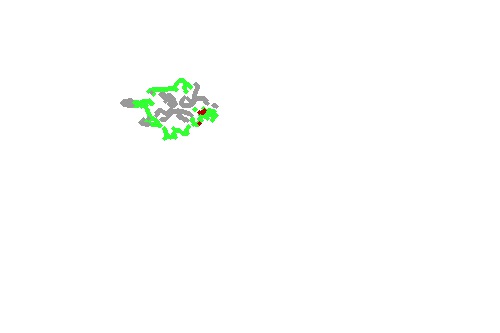}}\tabularnewline
\hline 
\multicolumn{7}{c}{bicycle\foreignlanguage{english}{\rule{0pt}{2.2ex}\rule[-1.0ex]{0pt}{0pt}}}\tabularnewline
\hline 
\multicolumn{1}{|c|}{\includegraphics[width=0.14\textwidth]{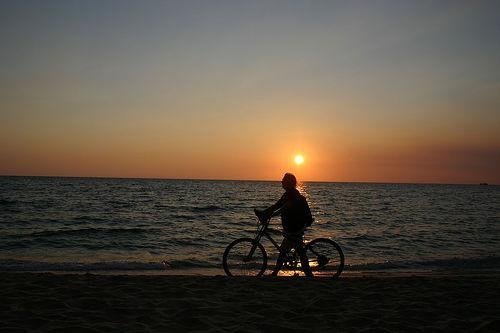}} & \includegraphics[width=0.14\textwidth]{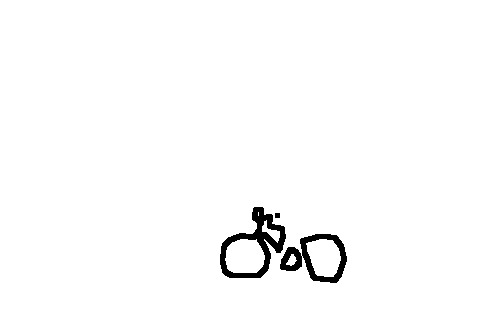} & \includegraphics[width=0.14\textwidth]{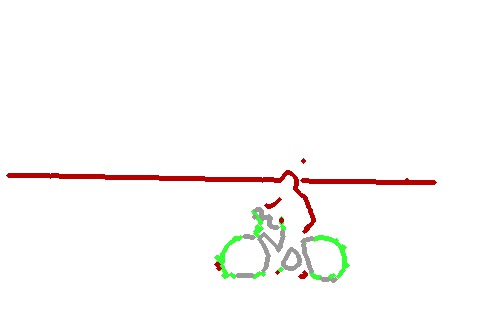} & \includegraphics[width=0.14\textwidth]{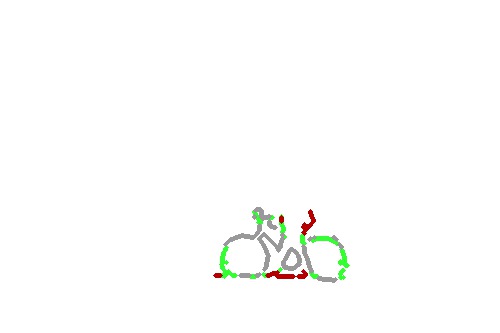} & \includegraphics[width=0.14\textwidth]{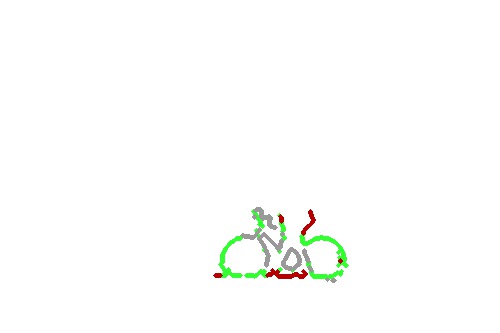} & \includegraphics[width=0.14\textwidth]{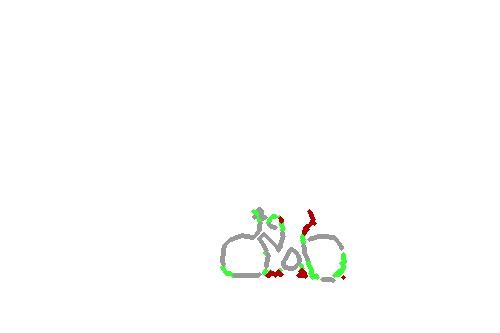} & \multicolumn{1}{c|}{\includegraphics[width=0.14\textwidth]{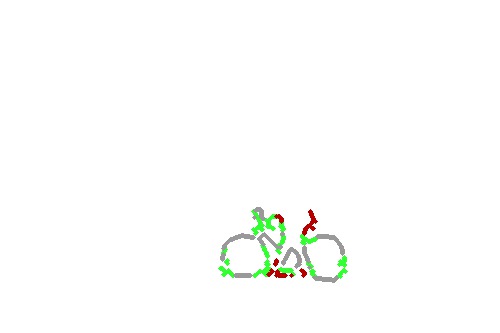}}\tabularnewline
\hline 
\multicolumn{7}{c}{bird\foreignlanguage{english}{\rule{0pt}{2.2ex}\rule[-1.0ex]{0pt}{0pt}}}\tabularnewline
\hline 
\multicolumn{1}{|c|}{\includegraphics[width=0.14\textwidth]{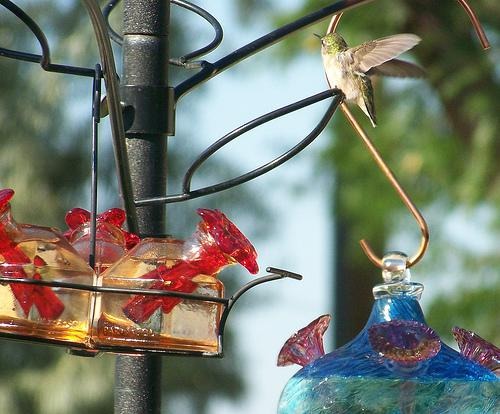}} & \includegraphics[width=0.14\textwidth]{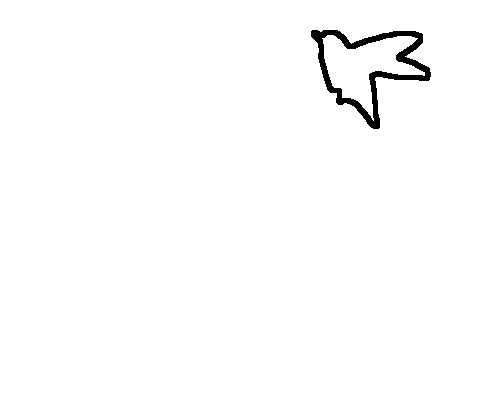} & \includegraphics[width=0.14\textwidth]{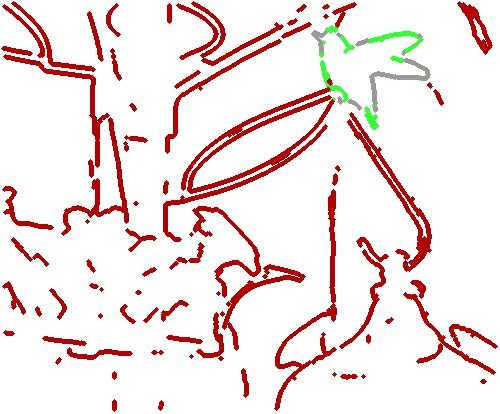} & \includegraphics[width=0.14\textwidth]{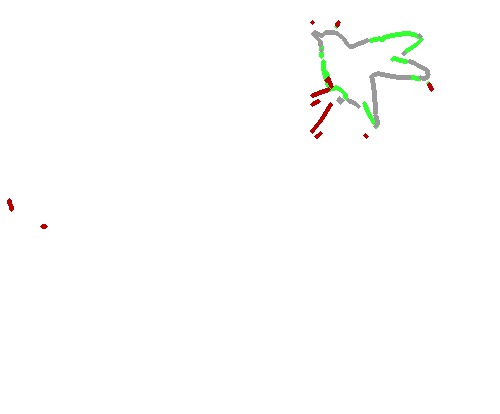} & \includegraphics[width=0.14\textwidth]{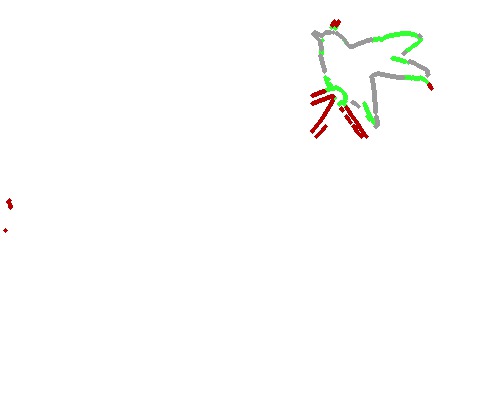} & \includegraphics[width=0.14\textwidth]{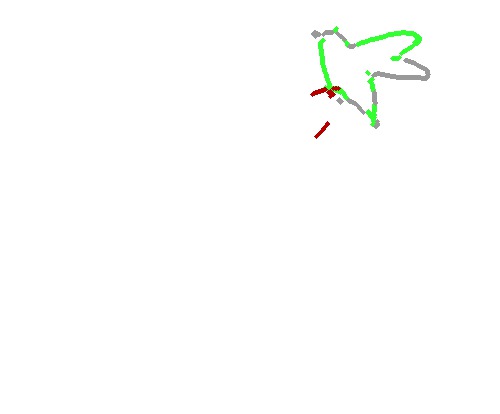} & \multicolumn{1}{c|}{\includegraphics[width=0.14\textwidth]{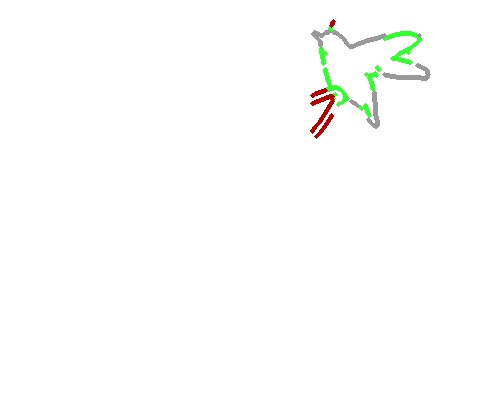}}\tabularnewline
\hline 
\multicolumn{7}{c}{boat\foreignlanguage{english}{\rule{0pt}{2.2ex}\rule[-1.0ex]{0pt}{0pt}}}\tabularnewline
\hline 
\multicolumn{1}{|c|}{\includegraphics[width=0.14\textwidth]{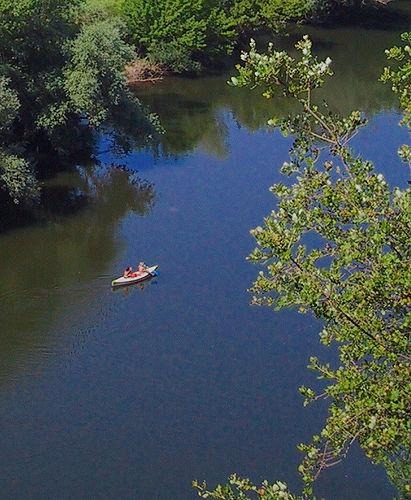}} & \includegraphics[width=0.14\textwidth]{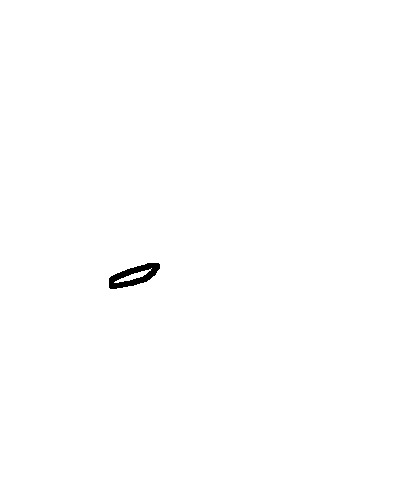} & \includegraphics[width=0.14\textwidth]{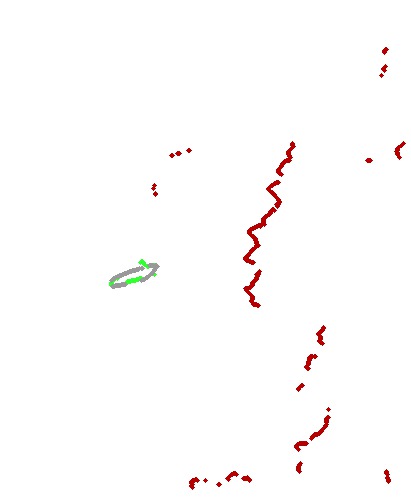} & \includegraphics[width=0.14\textwidth]{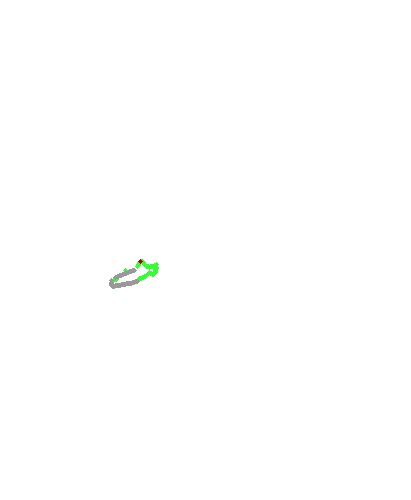} & \includegraphics[width=0.14\textwidth]{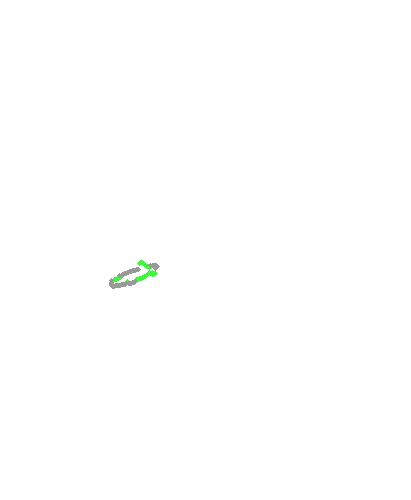} & \includegraphics[width=0.14\textwidth]{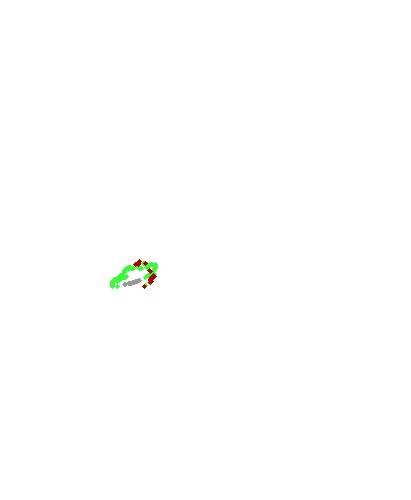} & \multicolumn{1}{c|}{\includegraphics[width=0.14\textwidth]{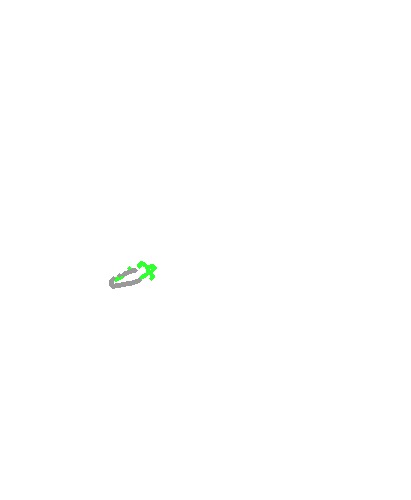}}\tabularnewline
\hline 
\multicolumn{7}{c}{bottle\foreignlanguage{english}{\rule{0pt}{2.2ex}\rule[-1.0ex]{0pt}{0pt}}}\tabularnewline
\hline 
\multicolumn{1}{|c|}{\includegraphics[width=0.14\textwidth]{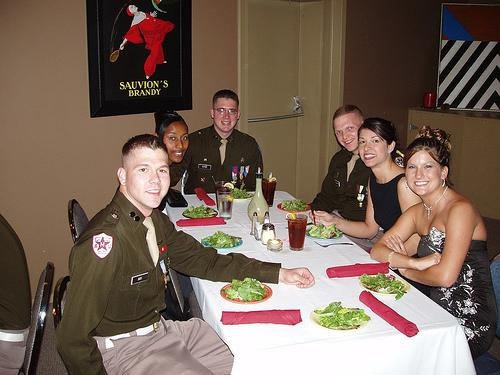}} & \includegraphics[width=0.14\textwidth]{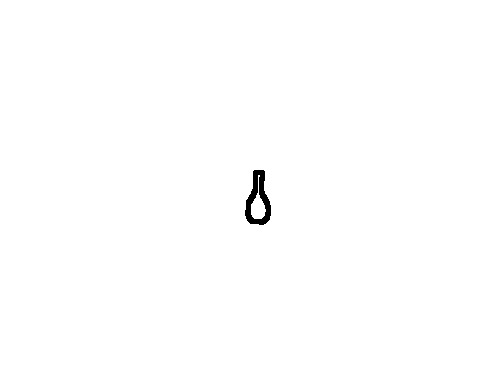} & \includegraphics[width=0.14\textwidth]{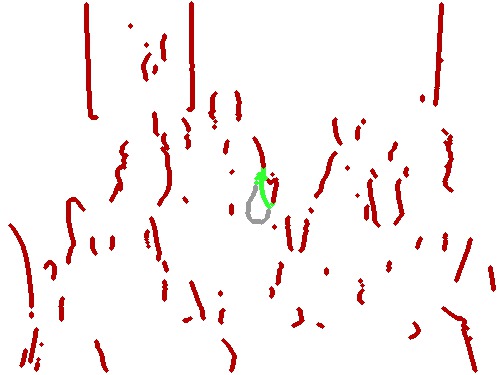} & \includegraphics[width=0.14\textwidth]{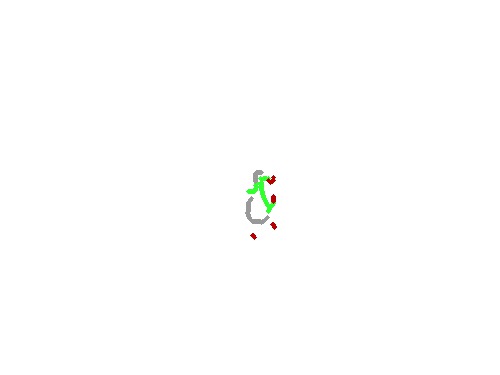} & \includegraphics[width=0.14\textwidth]{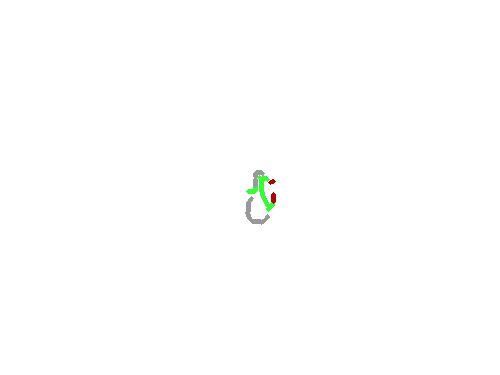} & \includegraphics[width=0.14\textwidth]{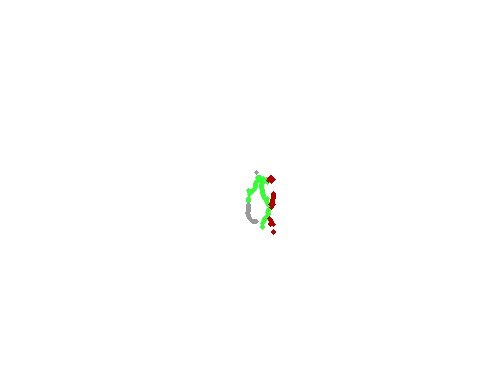} & \multicolumn{1}{c|}{\includegraphics[width=0.14\textwidth]{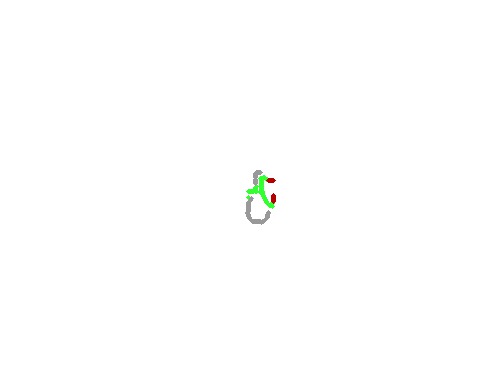}}\tabularnewline
\hline 
\multicolumn{7}{c}{bus\foreignlanguage{english}{\rule{0pt}{2.2ex}\rule[-1.0ex]{0pt}{0pt}}}\tabularnewline
\hline 
\multicolumn{1}{|c|}{\includegraphics[width=0.14\textwidth]{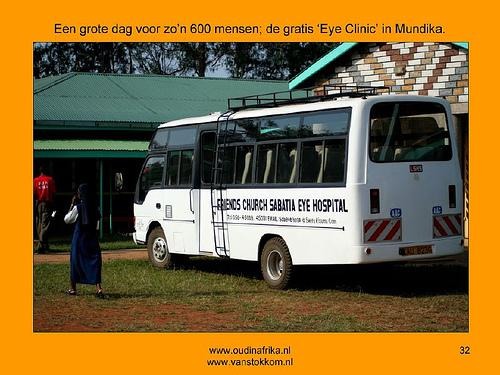}} & \includegraphics[width=0.14\textwidth]{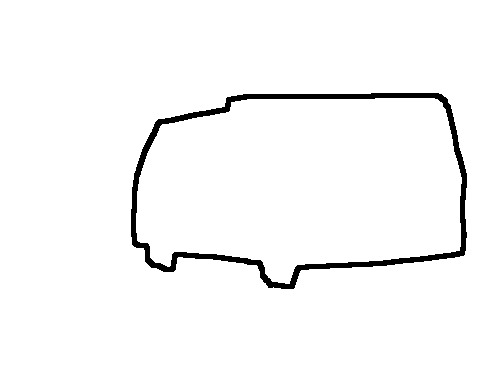} & \includegraphics[width=0.14\textwidth]{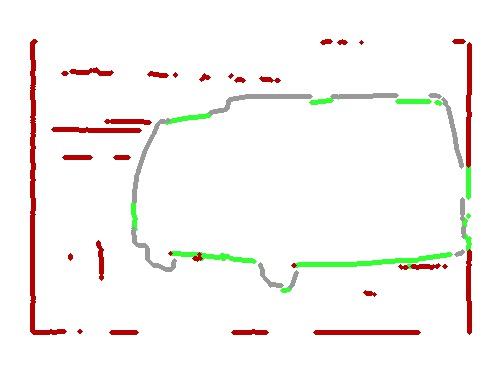} & \includegraphics[width=0.14\textwidth]{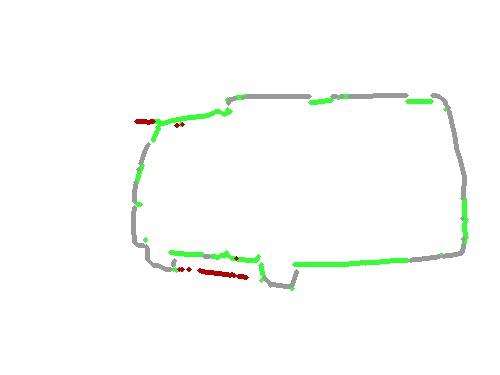} & \includegraphics[width=0.14\textwidth]{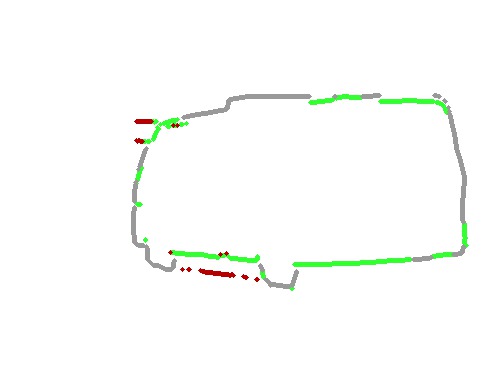} & \includegraphics[width=0.14\textwidth]{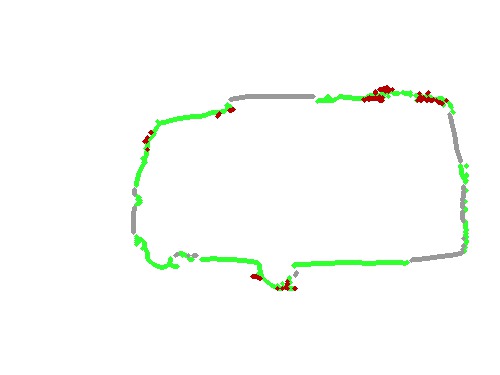} & \multicolumn{1}{c|}{\includegraphics[width=0.14\textwidth]{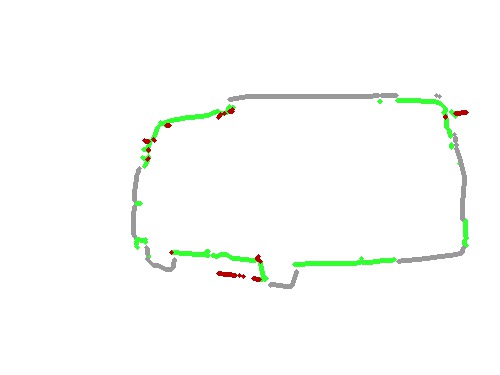}}\tabularnewline
\hline 
\multicolumn{7}{c}{car\foreignlanguage{english}{\rule{0pt}{2.2ex}\rule[-1.0ex]{0pt}{0pt}}}\tabularnewline
\hline 
\multicolumn{1}{|c|}{\includegraphics[width=0.14\textwidth]{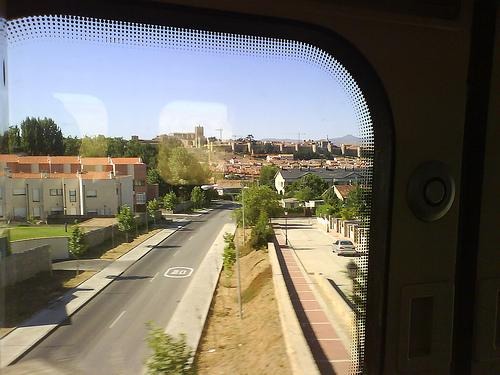}} & \includegraphics[width=0.14\textwidth]{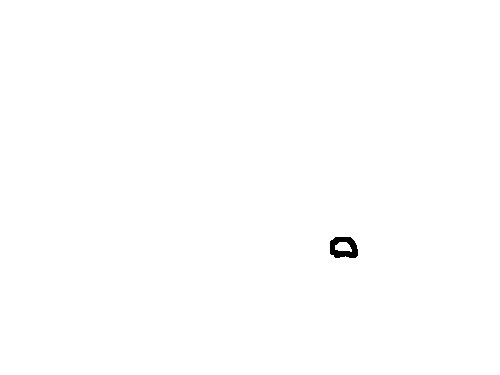} & \includegraphics[width=0.14\textwidth]{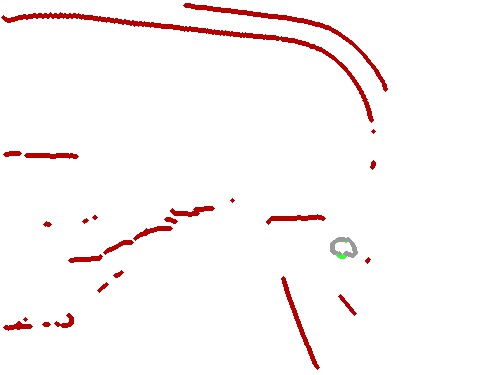} & \includegraphics[width=0.14\textwidth]{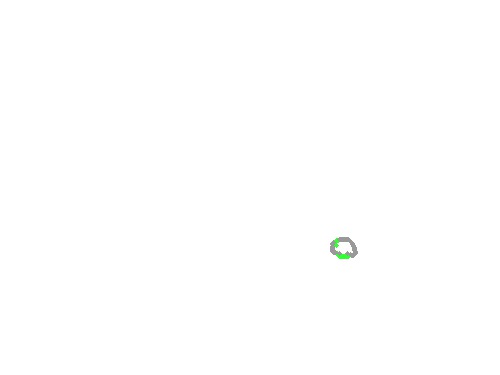} & \includegraphics[width=0.14\textwidth]{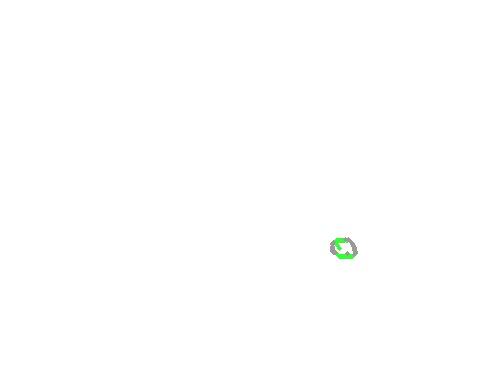} & \includegraphics[width=0.14\textwidth]{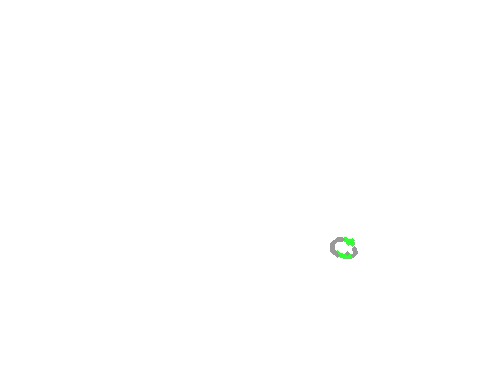} & \multicolumn{1}{c|}{\includegraphics[width=0.14\textwidth]{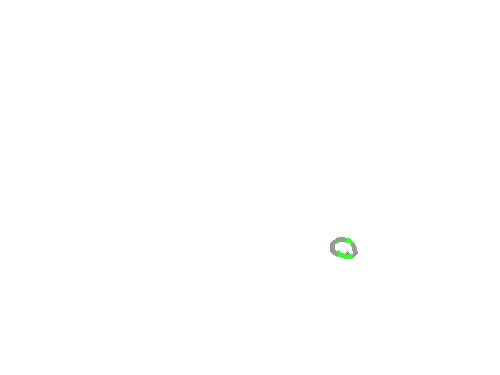}}\tabularnewline
\hline 
\multicolumn{1}{c}{\selectlanguage{english}%
{\small{}\rule{0pt}{2.6ex}}\foreignlanguage{british}{{\small{}Image}}\selectlanguage{british}%
} & \multicolumn{1}{c}{{\small{}Ground truth}} & \multicolumn{1}{c}{{\small{}SB(SBD)}} & \multicolumn{1}{c}{{\small{}$\mbox{Det.}\negmedspace+\negmedspace\mbox{SE}\left(\mbox{SBD}\right)$}} & \multicolumn{1}{c}{{\small{}$\mbox{Det.}\negmedspace+\negmedspace\mbox{SE}\left(\mathbf{\mathbf{weak}}\right)$}} & \multicolumn{1}{c}{{\small{}$\mbox{Det.}\negmedspace+\negmedspace\mbox{HED}\left(\mbox{SBD}\right)$}} & {\small{}$\mbox{Det.}\negthickspace+\negthickspace\mbox{HED}\left(\mathbf{weak}\right)$}\tabularnewline
\end{tabular}

}\endgroup
\arrayrulecolor{black}\protect\caption{Qualitative results on SBD. Red/green indicates false/true positives
pixels, grey is missing recall. All methods shown at $50\%$ recall.
{\small{}$\mbox{Det.}\negmedspace+\negmedspace\mbox{SE}\left(\mathbf{weak}\right)$}
denotes the model {\small{}$\mbox{Det.}\negmedspace+\negmedspace\mbox{SE}\left(\mbox{MCG}_{+}\cap\mbox{BBs}\right)$}
and {\small{}$\mbox{Det.}\negmedspace+\negmedspace\mbox{HED}\left(\mathbf{weak}\right)$}
denotes {\small{}$\mbox{Det.}\negthickspace+\negthickspace\mbox{HED}\left(\textrm{cons.\mbox{ S\&G}\ensuremath{\cap}BBs}\right)$}.
As the quantitative results indicate, qualitatively our weakly supervised
results are on par to the fully supervised ones.}
\end{figure*}

\setcounter{figure}{17}

\begin{figure*}
\begingroup{\arrayrulecolor{lightgray}
\setlength{\tabcolsep}{0pt} 
\renewcommand{\arraystretch}{0}

\begin{tabular}[b]{c|c|c||c|c||c|c}
\multicolumn{7}{c}{cat\foreignlanguage{english}{\rule[-1.0ex]{0pt}{0pt}}}\tabularnewline
\hline 
\multicolumn{1}{|c|}{\includegraphics[width=0.14\textwidth]{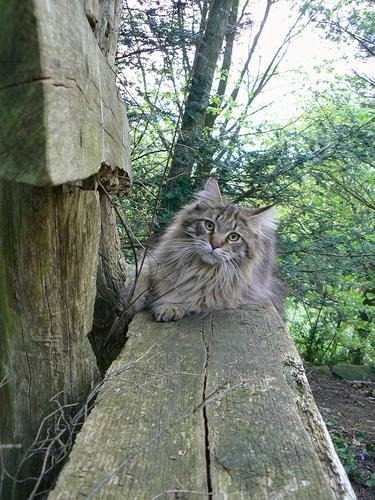}} & \includegraphics[width=0.14\textwidth]{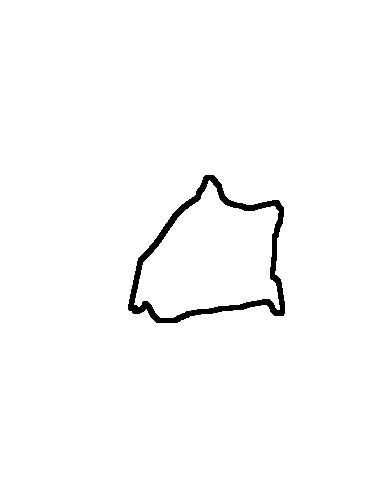} & \includegraphics[width=0.14\textwidth]{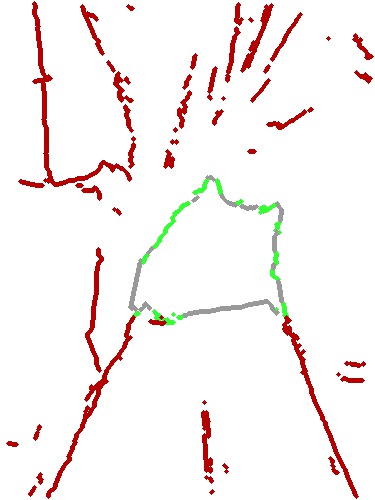} & \includegraphics[width=0.14\textwidth]{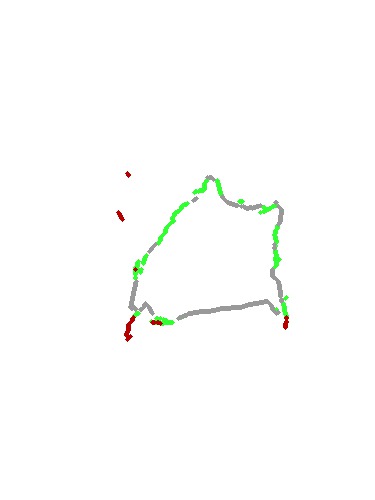} & \includegraphics[width=0.14\textwidth]{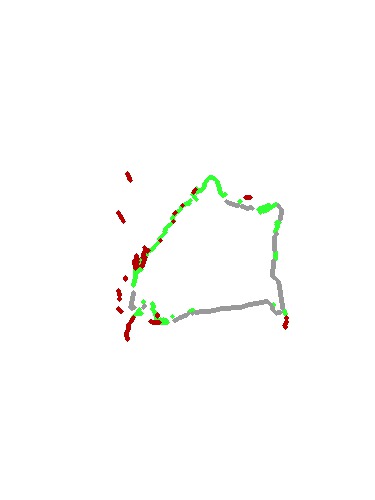} & \includegraphics[width=0.14\textwidth]{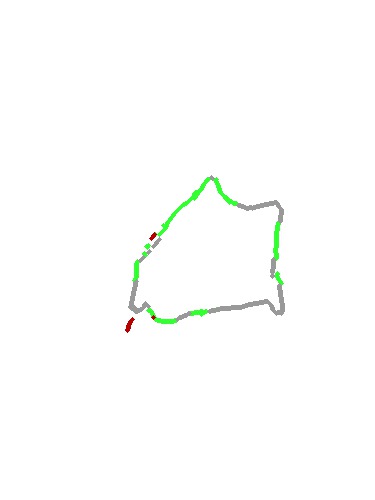} & \multicolumn{1}{c|}{\includegraphics[width=0.14\textwidth]{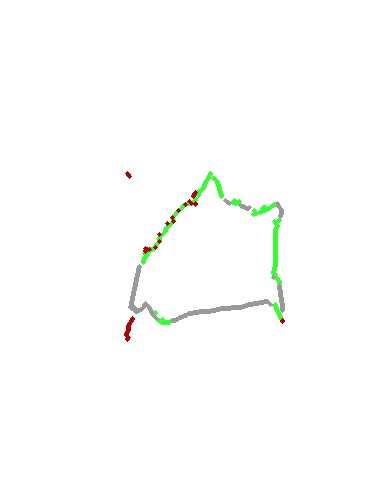}}\tabularnewline
\hline 
\multicolumn{7}{c}{chair\foreignlanguage{english}{\rule{0pt}{2.2ex}\rule[-1.0ex]{0pt}{0pt}}}\tabularnewline
\hline 
\multicolumn{1}{|c|}{\includegraphics[width=0.14\textwidth]{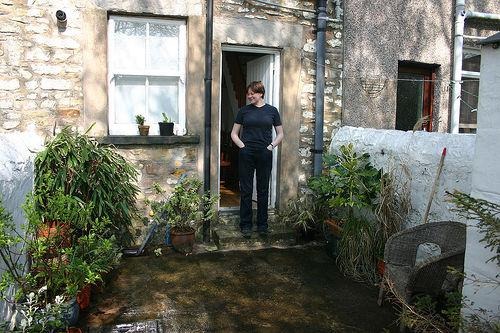}} & \includegraphics[width=0.14\textwidth]{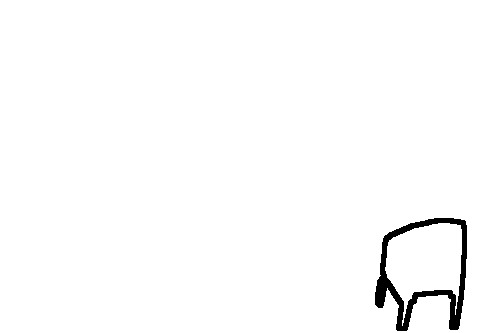} & \includegraphics[width=0.14\textwidth]{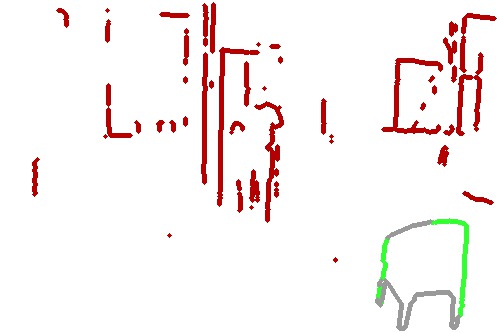} & \includegraphics[width=0.14\textwidth]{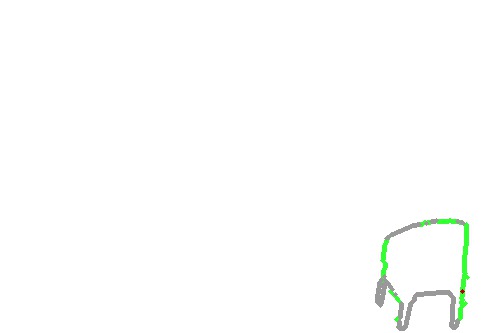} & \includegraphics[width=0.14\textwidth]{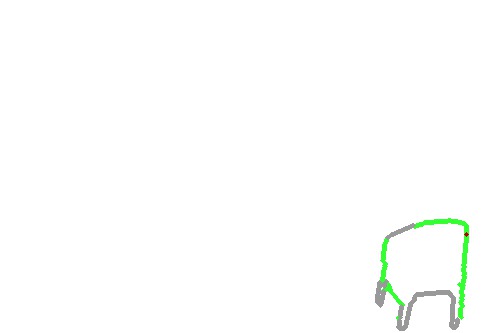} & \includegraphics[width=0.14\textwidth]{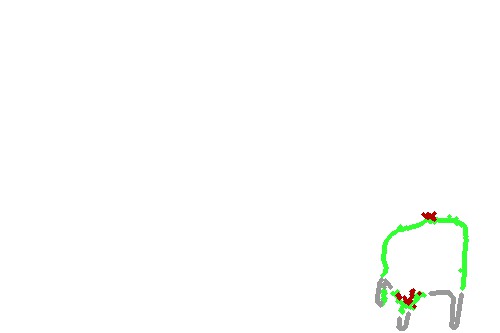} & \multicolumn{1}{c|}{\includegraphics[width=0.14\textwidth]{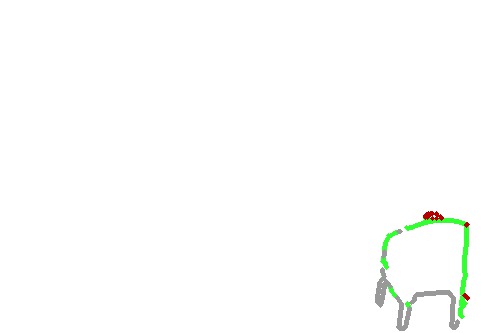}}\tabularnewline
\hline 
\multicolumn{7}{c}{cow\foreignlanguage{english}{\rule{0pt}{2.2ex}\rule[-1.0ex]{0pt}{0pt}}}\tabularnewline
\hline 
\multicolumn{1}{|c|}{\includegraphics[width=0.14\textwidth]{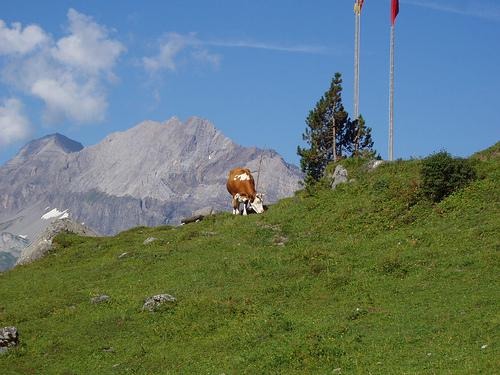}} & \includegraphics[width=0.14\textwidth]{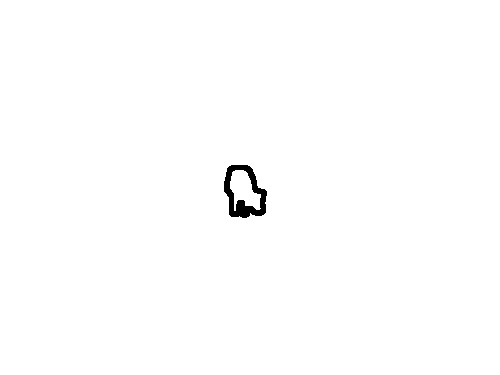} & \includegraphics[width=0.14\textwidth]{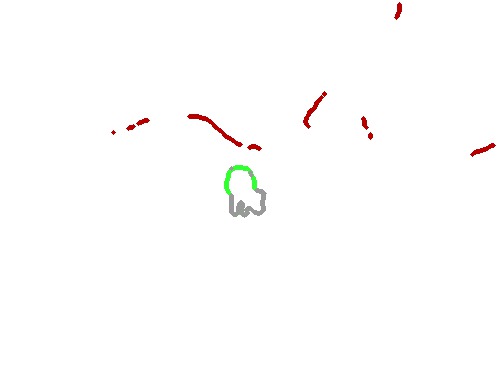} & \includegraphics[width=0.14\textwidth]{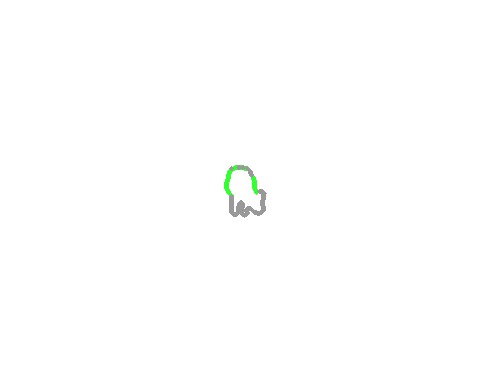} & \includegraphics[width=0.14\textwidth]{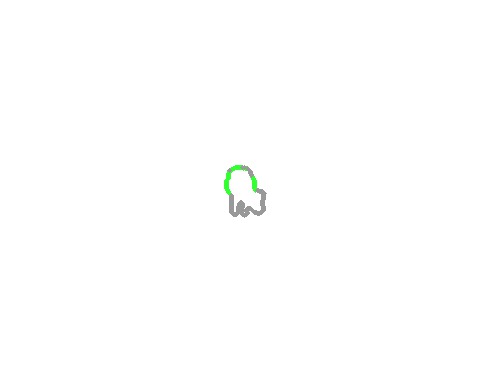} & \includegraphics[width=0.14\textwidth]{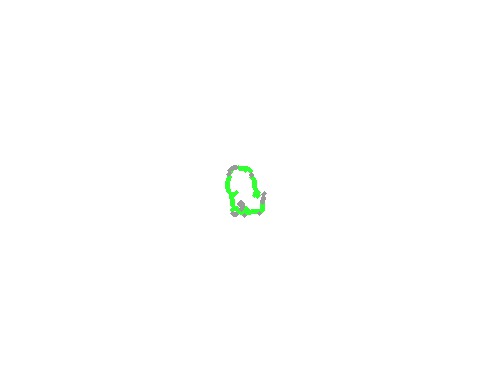} & \multicolumn{1}{c|}{\includegraphics[width=0.14\textwidth]{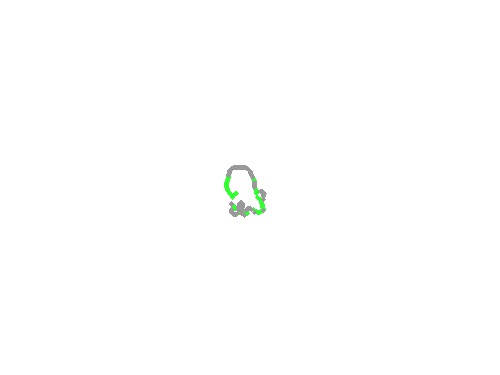}}\tabularnewline
\hline 
\multicolumn{7}{c}{dining table\foreignlanguage{english}{\rule{0pt}{2.2ex}\rule[-1.0ex]{0pt}{0pt}}}\tabularnewline
\hline 
\multicolumn{1}{|c|}{\includegraphics[width=0.14\textwidth]{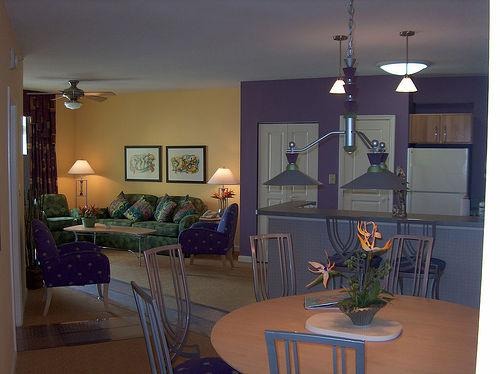}} & \includegraphics[width=0.14\textwidth]{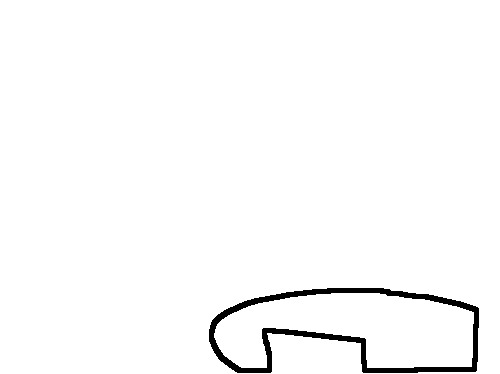} & \includegraphics[width=0.14\textwidth]{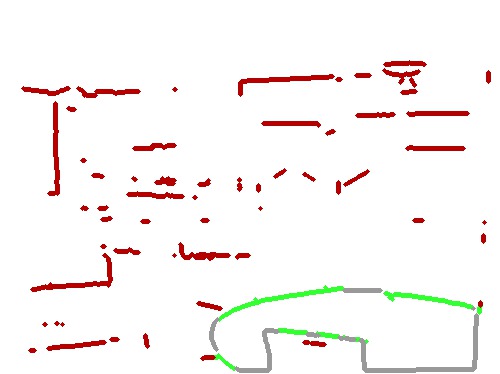} & \includegraphics[width=0.14\textwidth]{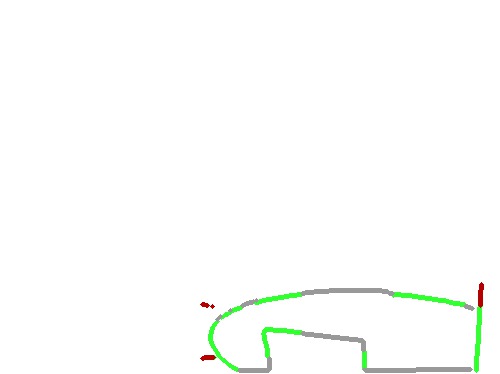} & \includegraphics[width=0.14\textwidth]{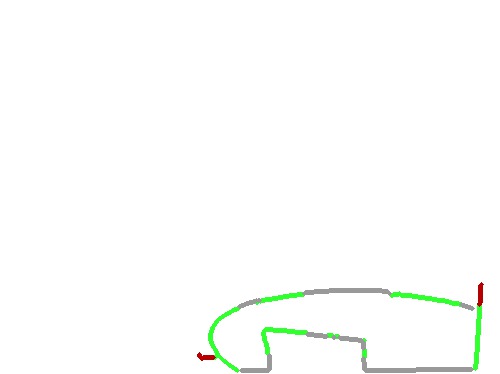} & \includegraphics[width=0.14\textwidth]{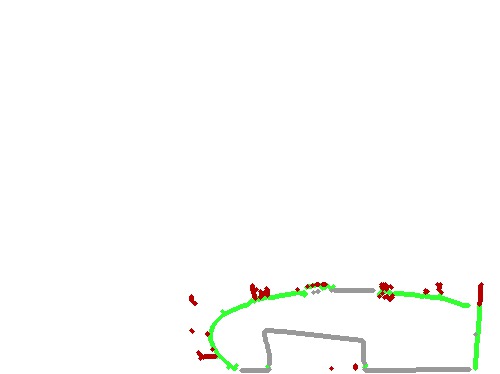} & \multicolumn{1}{c|}{\includegraphics[width=0.14\textwidth]{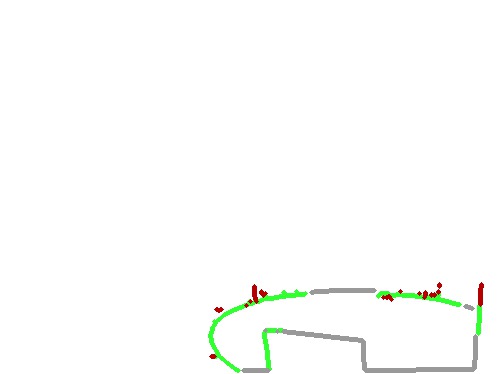}}\tabularnewline
\hline 
\multicolumn{7}{c}{dog\foreignlanguage{english}{\rule{0pt}{2.2ex}\rule[-1.0ex]{0pt}{0pt}}}\tabularnewline
\hline 
\multicolumn{1}{|c|}{\includegraphics[width=0.14\textwidth]{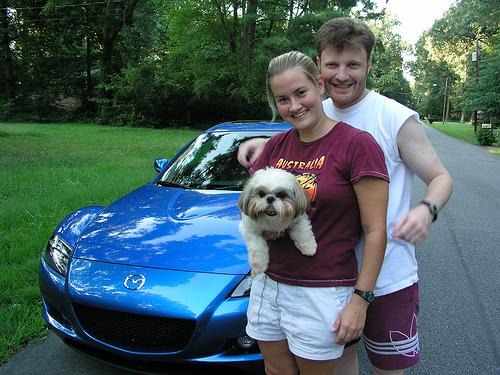}} & \includegraphics[width=0.14\textwidth]{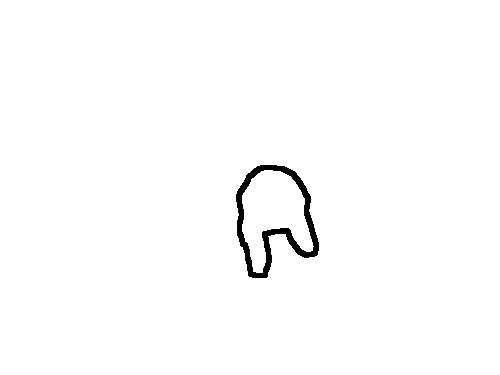} & \includegraphics[width=0.14\textwidth]{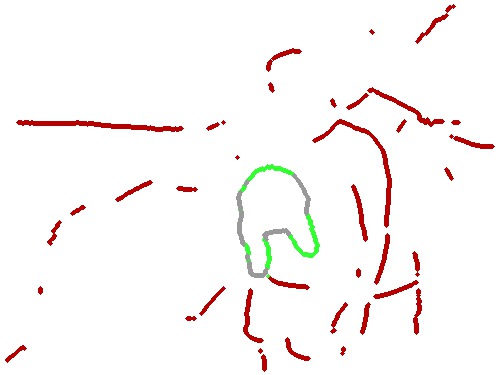} & \includegraphics[width=0.14\textwidth]{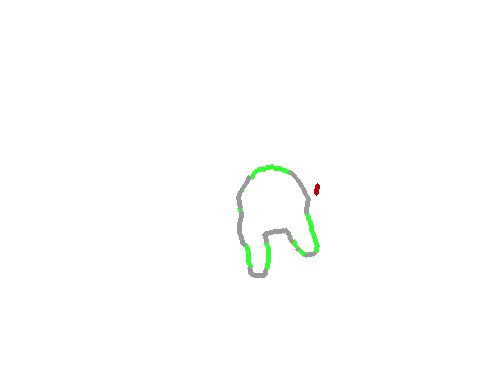} & \includegraphics[width=0.14\textwidth]{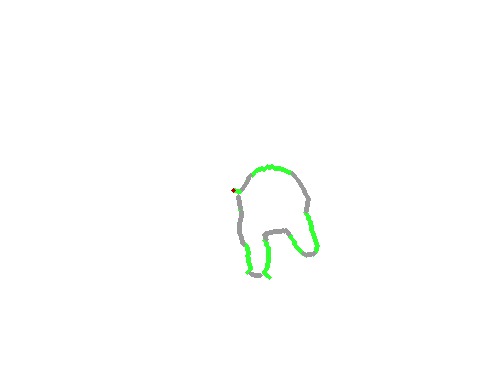} & \includegraphics[width=0.14\textwidth]{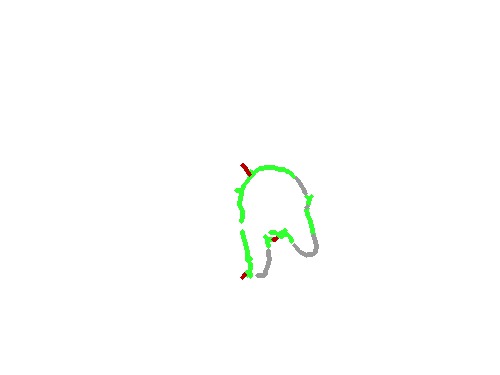} & \multicolumn{1}{c|}{\includegraphics[width=0.14\textwidth]{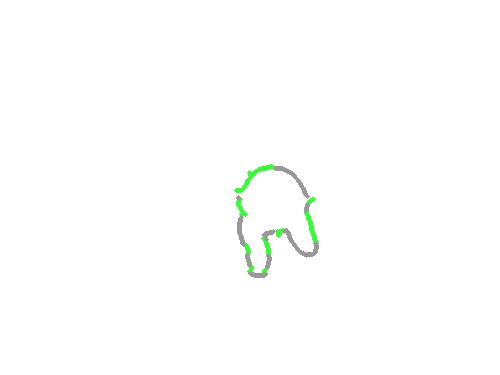}}\tabularnewline
\hline 
\multicolumn{7}{c}{horse\foreignlanguage{english}{\rule{0pt}{2.2ex}\rule[-1.0ex]{0pt}{0pt}}}\tabularnewline
\hline 
\multicolumn{1}{|c|}{\includegraphics[width=0.14\textwidth]{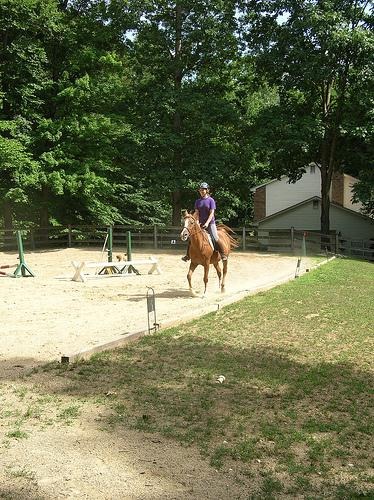}} & \includegraphics[width=0.14\textwidth]{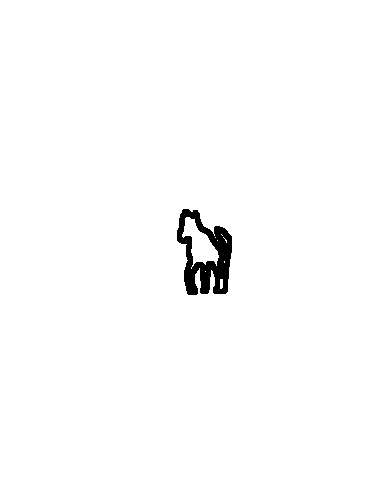} & \includegraphics[width=0.14\textwidth]{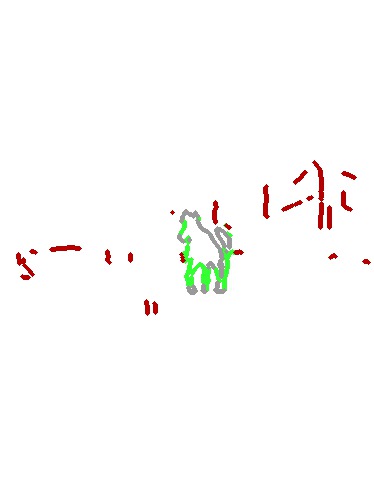} & \includegraphics[width=0.14\textwidth]{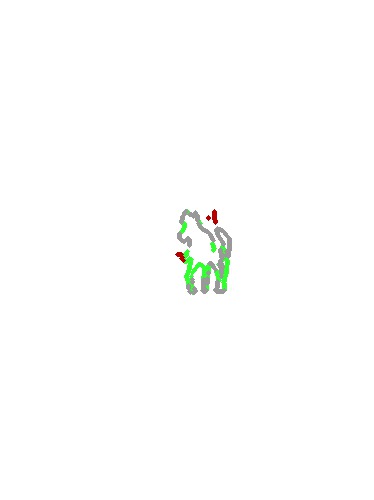} & \includegraphics[width=0.14\textwidth]{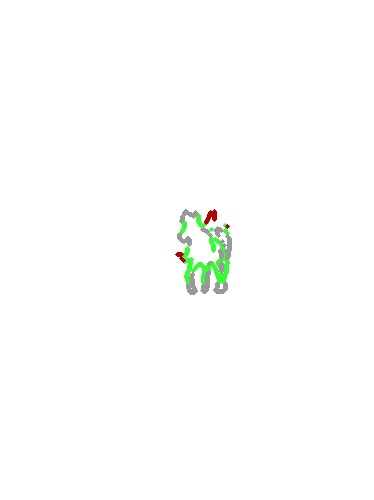} & \includegraphics[width=0.14\textwidth]{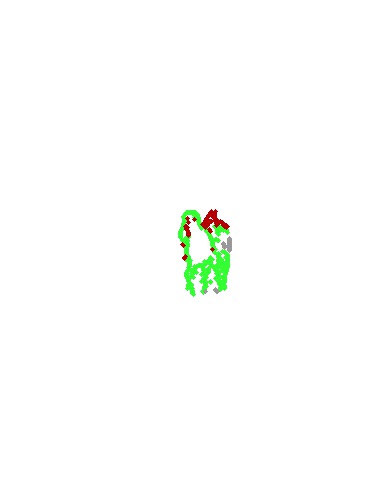} & \multicolumn{1}{c|}{\includegraphics[width=0.14\textwidth]{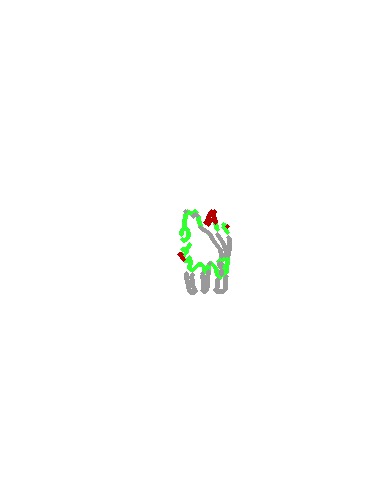}}\tabularnewline
\hline 
\multicolumn{7}{c}{motorbike\foreignlanguage{english}{\rule{0pt}{2.2ex}\rule[-1.0ex]{0pt}{0pt}}}\tabularnewline
\hline 
\multicolumn{1}{|c|}{\includegraphics[width=0.14\textwidth]{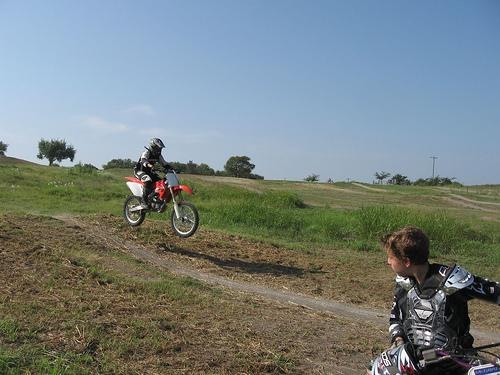}} & \includegraphics[width=0.14\textwidth]{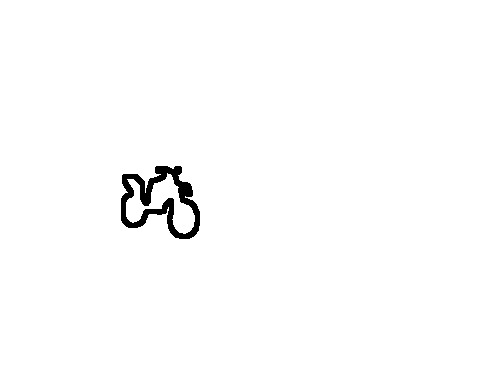} & \includegraphics[width=0.14\textwidth]{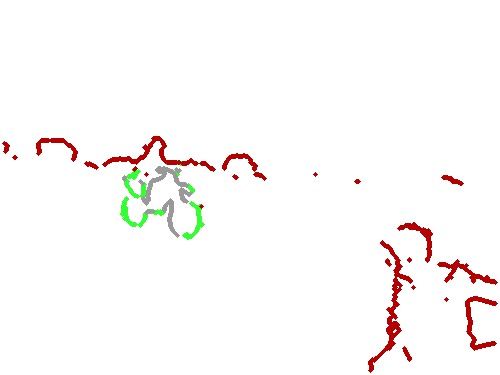} & \includegraphics[width=0.14\textwidth]{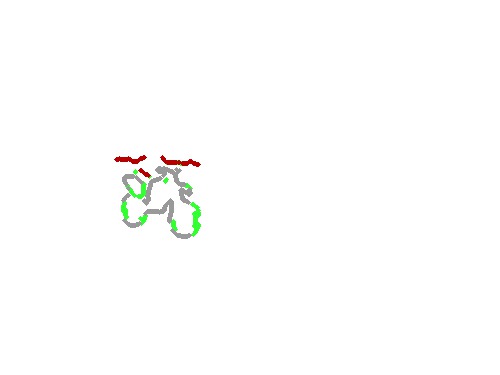} & \includegraphics[width=0.14\textwidth]{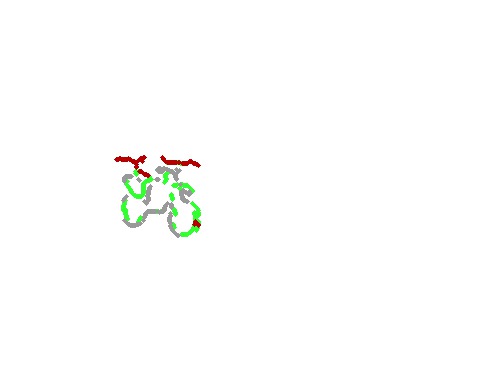} & \includegraphics[width=0.14\textwidth]{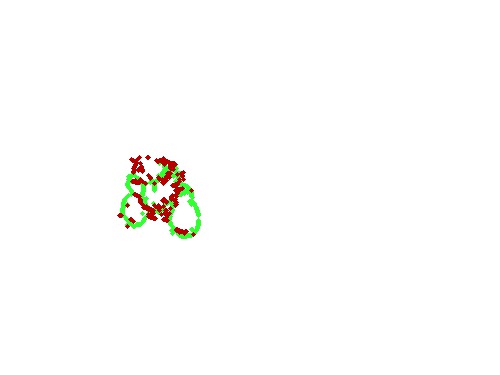} & \multicolumn{1}{c|}{\includegraphics[width=0.14\textwidth]{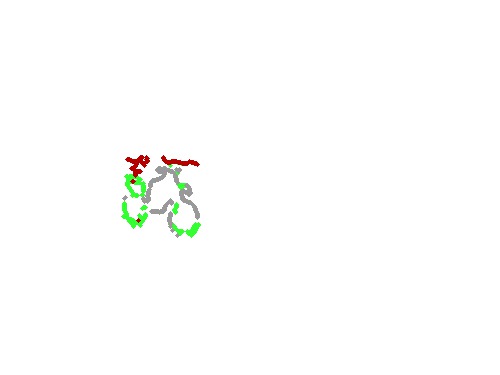}}\tabularnewline
\hline 
\multicolumn{7}{c}{}\tabularnewline
\hline 
\multicolumn{1}{|c|}{} &  &  &  &  &  & \multicolumn{1}{c|}{}\tabularnewline
\hline 
\multicolumn{1}{c}{\selectlanguage{english}%
{\small{}\rule{0pt}{2.6ex}}\foreignlanguage{british}{{\small{}Image}}\selectlanguage{british}%
} & \multicolumn{1}{c}{{\small{}Ground truth}} & \multicolumn{1}{c}{{\small{}SB(SBD)}} & \multicolumn{1}{c}{{\small{}$\mbox{Det.}\negmedspace+\negmedspace\mbox{SE}\left(\mbox{SBD}\right)$}} & \multicolumn{1}{c}{{\small{}$\mbox{Det.}\negmedspace+\negmedspace\mbox{SE}\left(\mathbf{\mathbf{weak}}\right)$}} & \multicolumn{1}{c}{{\small{}$\mbox{Det.}\negmedspace+\negmedspace\mbox{HED}\left(\mbox{SBD}\right)$}} & {\small{}$\mbox{Det.}\negthickspace+\negthickspace\mbox{HED}\left(\mathbf{weak}\right)$}\tabularnewline
\end{tabular}

}\endgroup
\arrayrulecolor{black}\protect\caption{Qualitative results on SBD. Red/green indicates false/true positives
pixels, grey is missing recall. All methods shown at $50\%$ recall.
{\small{}$\mbox{Det.}\negmedspace+\negmedspace\mbox{SE}\left(\mathbf{weak}\right)$}
denotes the model {\small{}$\mbox{Det.}\negmedspace+\negmedspace\mbox{SE}\left(\mbox{MCG}_{+}\cap\mbox{BBs}\right)$}
and {\small{}$\mbox{Det.}\negmedspace+\negmedspace\mbox{HED}\left(\mathbf{weak}\right)$}
denotes {\small{}$\mbox{Det.}\negthickspace+\negthickspace\mbox{HED}\left(\textrm{cons.\mbox{ S\&G}\ensuremath{\cap}BBs}\right)$}.
As the quantitative results indicate, qualitatively our weakly supervised
results are on par to the fully supervised ones.}
\end{figure*}

\setcounter{figure}{17}

\begin{figure*}
\begingroup{\arrayrulecolor{lightgray}
\setlength{\tabcolsep}{0pt} 
\renewcommand{\arraystretch}{0}

\begin{tabular}[b]{c|c|c||c|c||c|c}
\multicolumn{7}{c}{person\foreignlanguage{english}{\rule[-1.0ex]{0pt}{0pt}}}\tabularnewline
\hline 
\multicolumn{1}{|c|}{\includegraphics[width=0.14\textwidth]{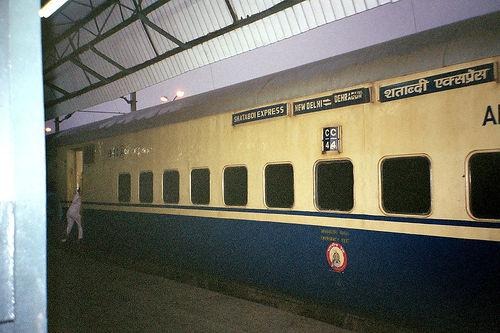}} & \includegraphics[width=0.14\textwidth]{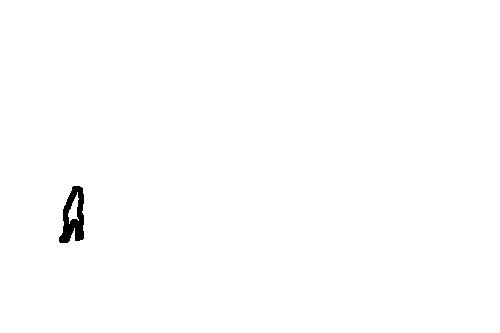} & \includegraphics[width=0.14\textwidth]{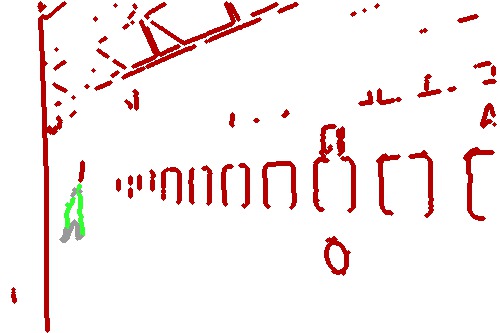} & \includegraphics[width=0.14\textwidth]{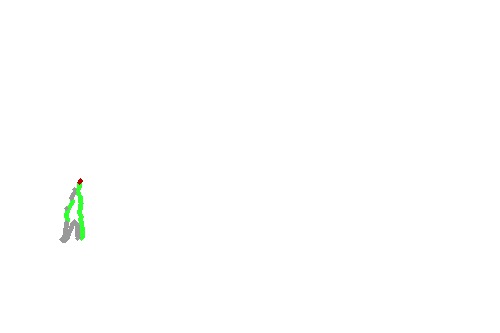} & \includegraphics[width=0.14\textwidth]{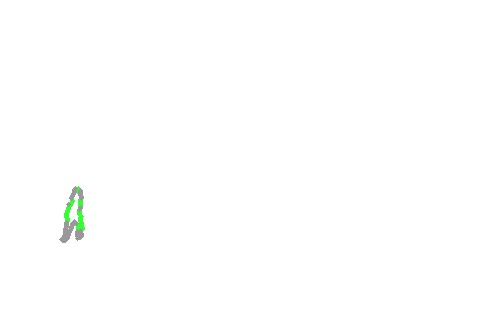} & \includegraphics[width=0.14\textwidth]{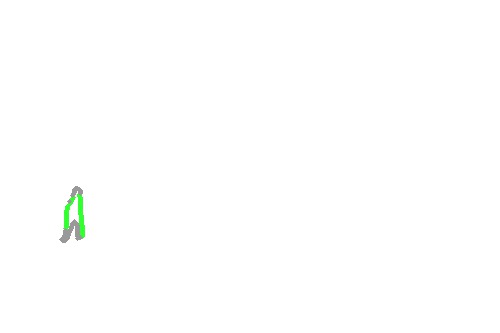} & \multicolumn{1}{c|}{\includegraphics[width=0.14\textwidth]{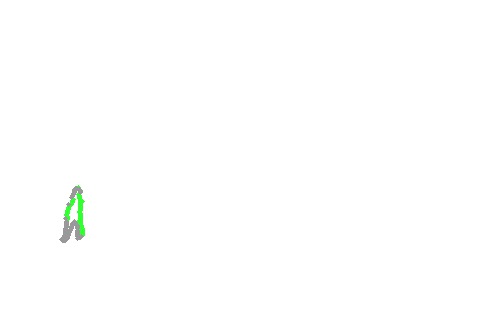}}\tabularnewline
\hline 
\multicolumn{7}{c}{potted plant\foreignlanguage{english}{\rule{0pt}{2.2ex}\rule[-1.0ex]{0pt}{0pt}}}\tabularnewline
\hline 
\multicolumn{1}{|c|}{\includegraphics[width=0.14\textwidth]{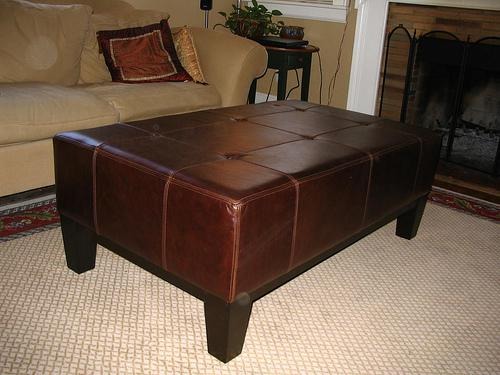}} & \includegraphics[width=0.14\textwidth]{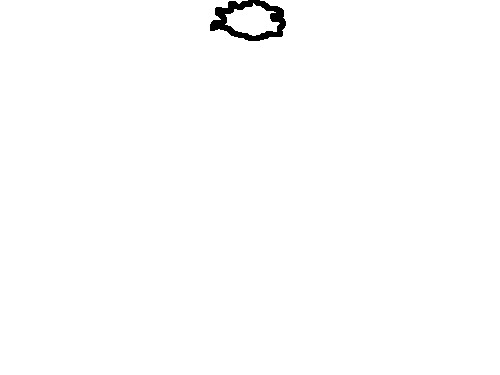} & \includegraphics[width=0.14\textwidth]{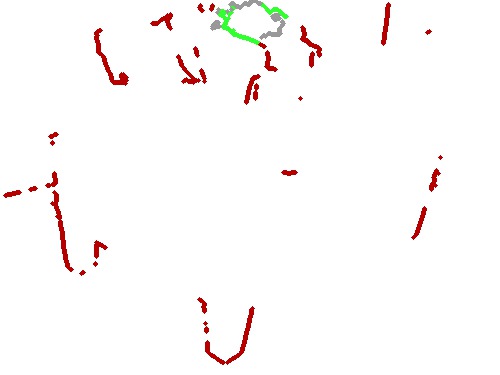} & \includegraphics[width=0.14\textwidth]{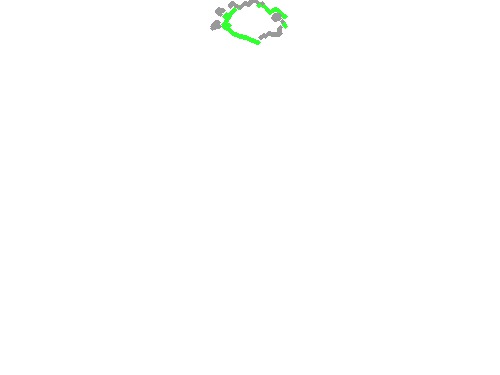} & \includegraphics[width=0.14\textwidth]{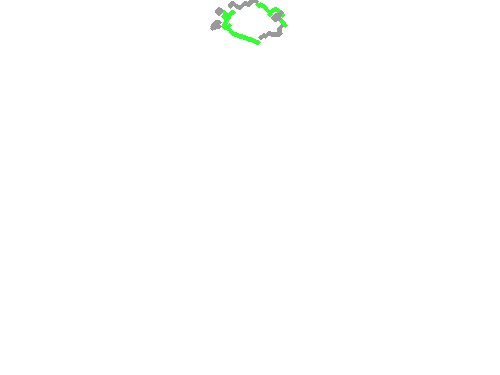} & \includegraphics[width=0.14\textwidth]{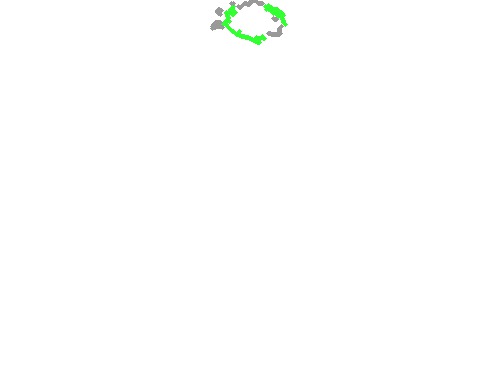} & \multicolumn{1}{c|}{\includegraphics[width=0.14\textwidth]{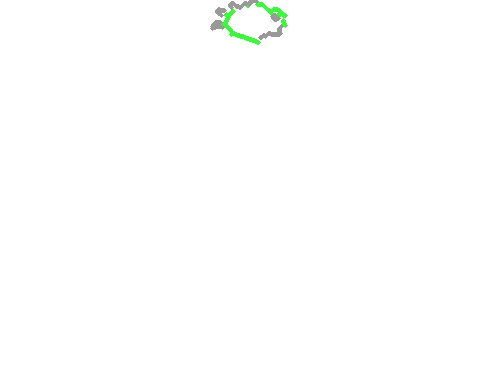}}\tabularnewline
\hline 
\multicolumn{7}{c}{sheep\foreignlanguage{english}{\rule{0pt}{2.2ex}\rule[-1.0ex]{0pt}{0pt}}}\tabularnewline
\hline 
\multicolumn{1}{|c|}{\includegraphics[width=0.14\textwidth]{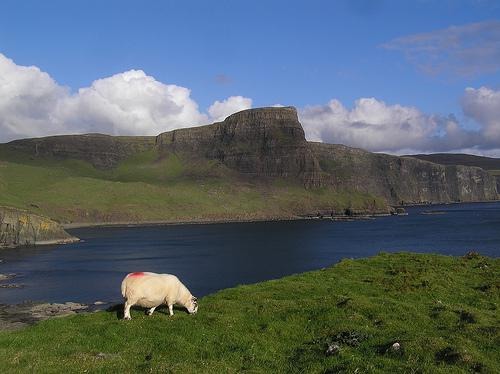}} & \includegraphics[width=0.14\textwidth]{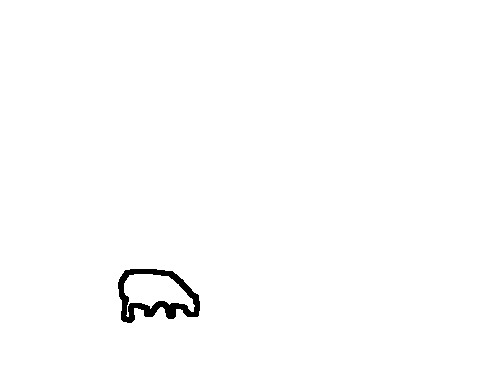} & \includegraphics[width=0.14\textwidth]{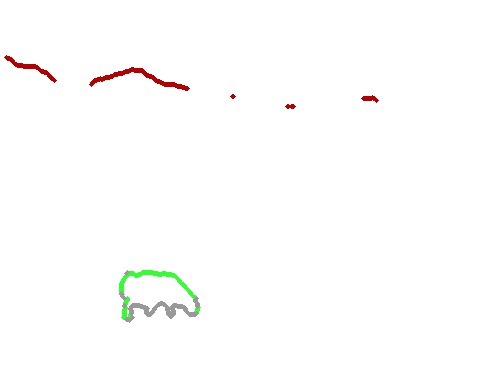} & \includegraphics[width=0.14\textwidth]{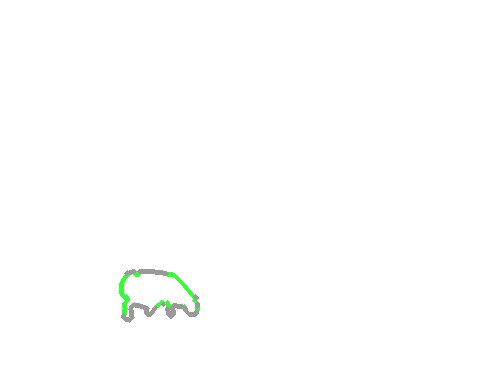} & \includegraphics[width=0.14\textwidth]{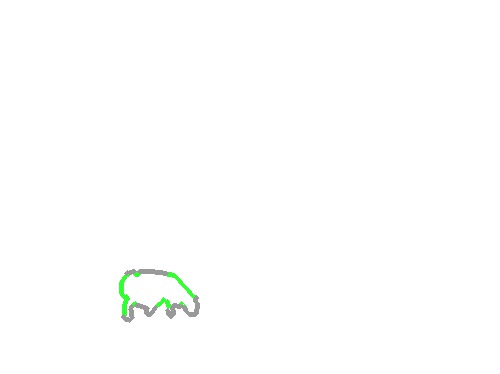} & \includegraphics[width=0.14\textwidth]{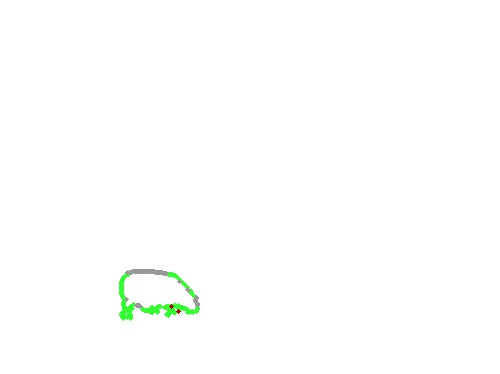} & \multicolumn{1}{c|}{\includegraphics[width=0.14\textwidth]{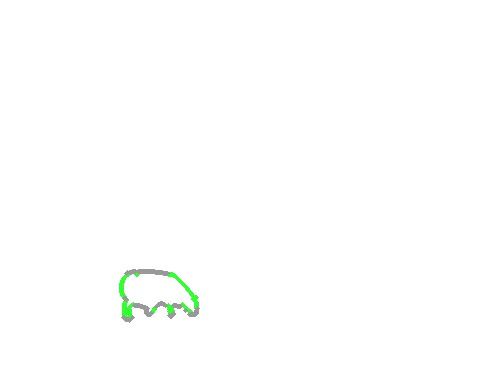}}\tabularnewline
\hline 
\multicolumn{7}{c}{sofa\foreignlanguage{english}{\rule{0pt}{2.2ex}\rule[-1.0ex]{0pt}{0pt}}}\tabularnewline
\hline 
\multicolumn{1}{|c|}{\includegraphics[width=0.14\textwidth]{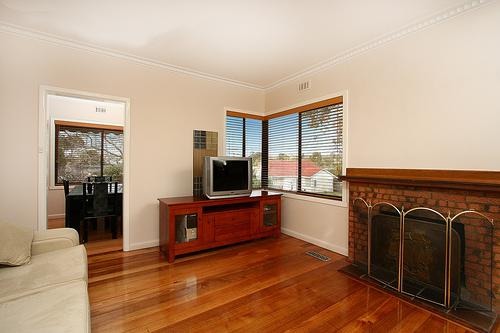}} & \includegraphics[width=0.14\textwidth]{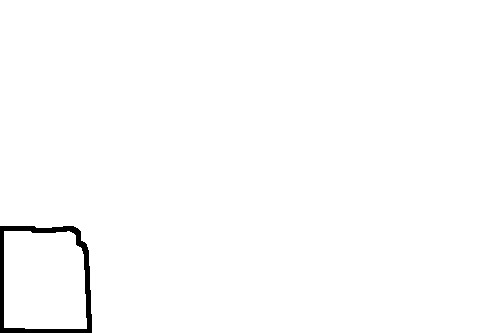} & \includegraphics[width=0.14\textwidth]{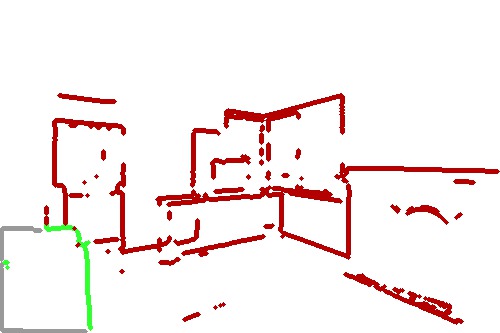} & \includegraphics[width=0.14\textwidth]{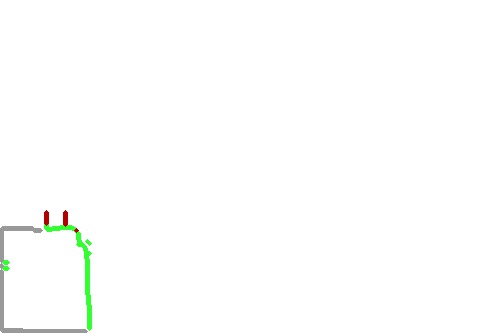} & \includegraphics[width=0.14\textwidth]{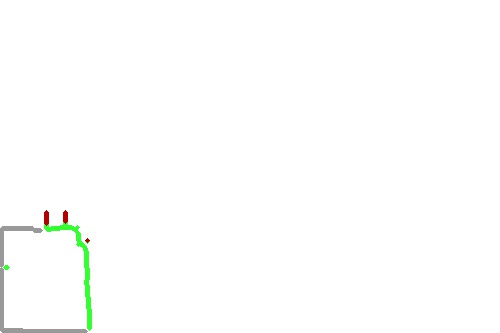} & \includegraphics[width=0.14\textwidth]{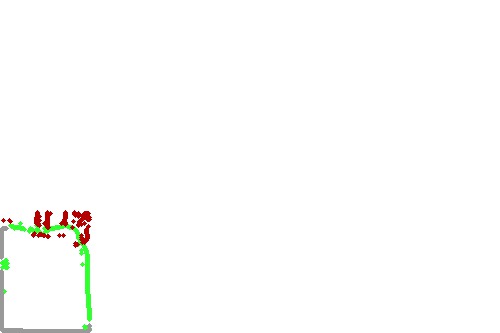} & \multicolumn{1}{c|}{\includegraphics[width=0.14\textwidth]{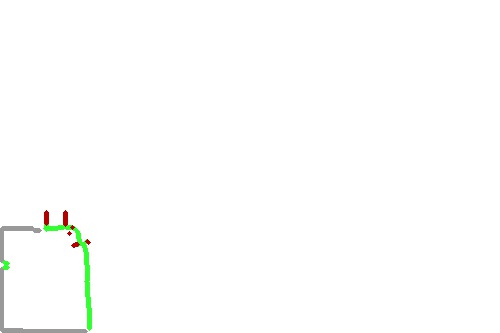}}\tabularnewline
\hline 
\multicolumn{7}{c}{train\foreignlanguage{english}{\rule{0pt}{2.2ex}\rule[-1.0ex]{0pt}{0pt}}}\tabularnewline
\hline 
\multicolumn{1}{|c|}{\includegraphics[width=0.14\textwidth]{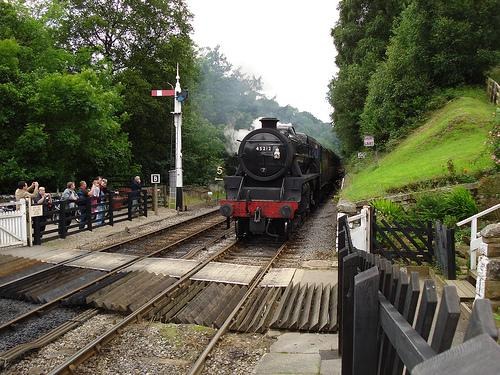}} & \includegraphics[width=0.14\textwidth]{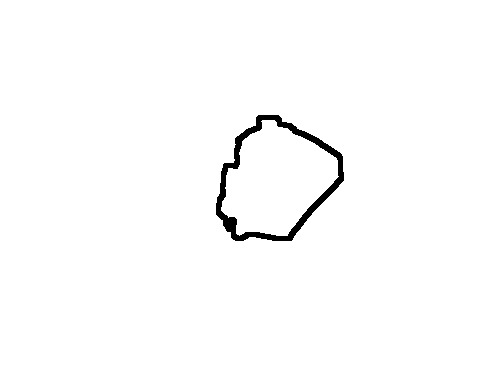} & \includegraphics[width=0.14\textwidth]{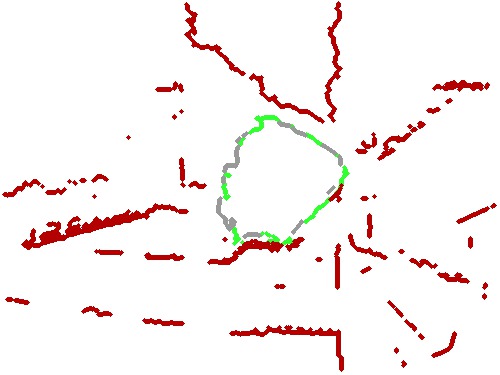} & \includegraphics[width=0.14\textwidth]{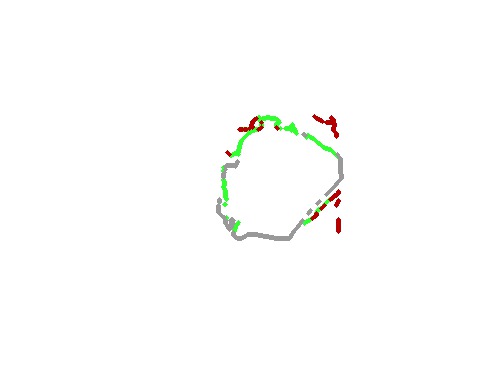} & \includegraphics[width=0.14\textwidth]{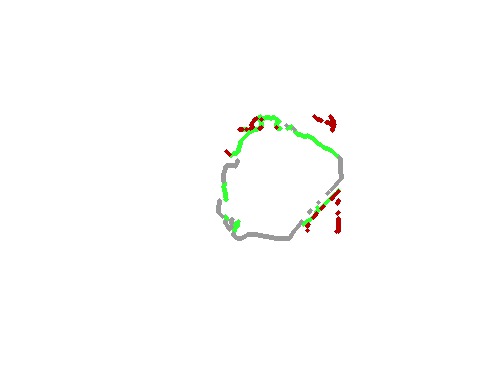} & \includegraphics[width=0.14\textwidth]{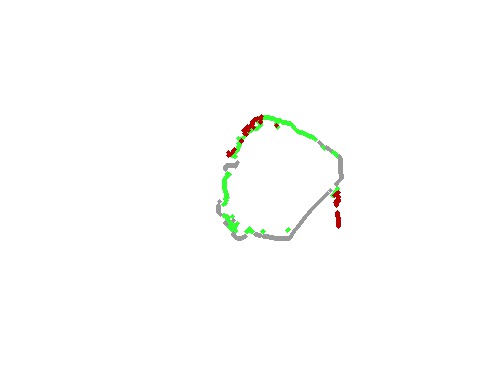} & \multicolumn{1}{c|}{\includegraphics[width=0.14\textwidth]{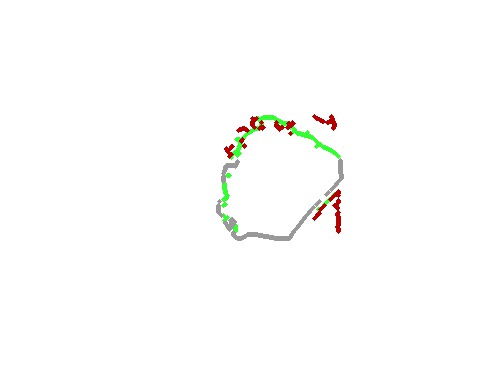}}\tabularnewline
\hline 
\multicolumn{7}{c}{tv monitor\foreignlanguage{english}{\rule{0pt}{2.2ex}\rule[-1.0ex]{0pt}{0pt}}}\tabularnewline
\hline 
\multicolumn{1}{|c|}{\includegraphics[width=0.14\textwidth]{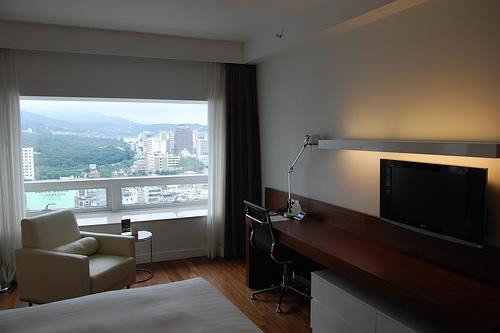}} & \includegraphics[width=0.14\textwidth]{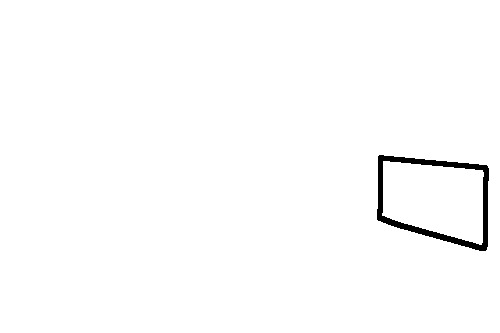} & \includegraphics[width=0.14\textwidth]{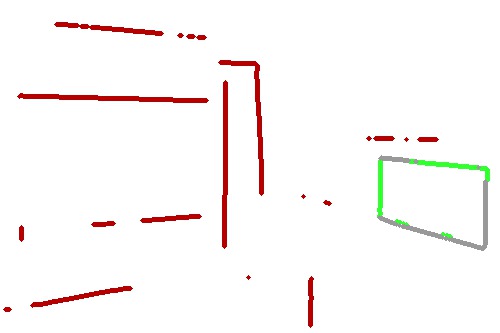} & \includegraphics[width=0.14\textwidth]{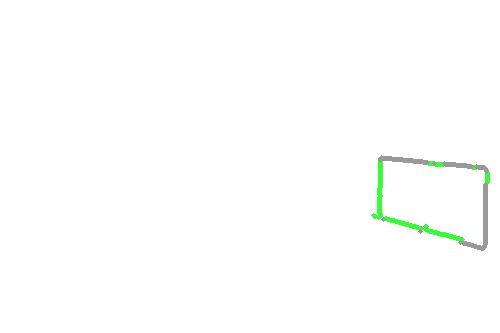} & \includegraphics[width=0.14\textwidth]{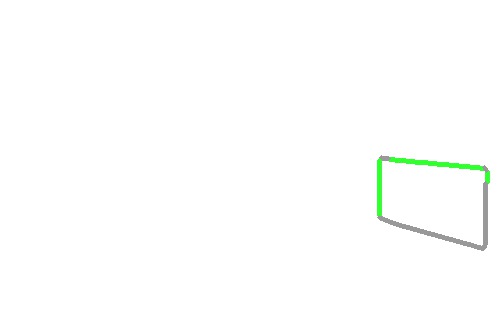} & \includegraphics[width=0.14\textwidth]{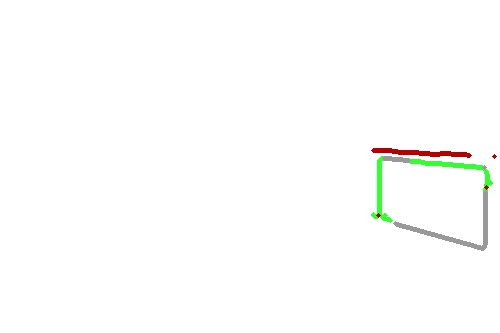} & \multicolumn{1}{c|}{\includegraphics[width=0.14\textwidth]{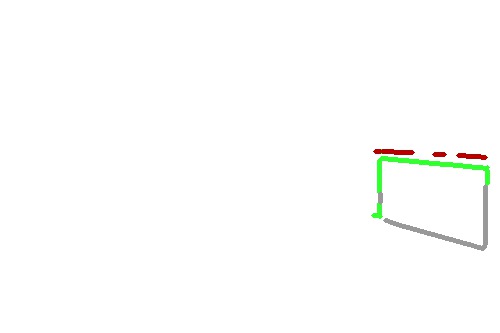}}\tabularnewline
\hline 
\multicolumn{1}{c}{\selectlanguage{english}%
{\small{}\rule{0pt}{2.6ex}}\foreignlanguage{british}{{\small{}Image}}\selectlanguage{british}%
} & \multicolumn{1}{c}{{\small{}Ground truth}} & \multicolumn{1}{c}{{\small{}SB(SBD)}} & \multicolumn{1}{c}{{\small{}$\mbox{Det.}\negmedspace+\negmedspace\mbox{SE}\left(\mbox{SBD}\right)$}} & \multicolumn{1}{c}{{\small{}$\mbox{Det.}\negmedspace+\negmedspace\mbox{SE}\left(\mathbf{\mathbf{weak}}\right)$}} & \multicolumn{1}{c}{{\small{}$\mbox{Det.}\negmedspace+\negmedspace\mbox{HED}\left(\mbox{SBD}\right)$}} & {\small{}$\mbox{Det.}\negthickspace+\negthickspace\mbox{HED}\left(\mathbf{weak}\right)$}\tabularnewline
\end{tabular}

}\endgroup
\arrayrulecolor{black}\protect\caption{\label{fig:Qualitative-results-sbd}Qualitative results on SBD. Red/green
indicates false/true positives pixels, grey is missing recall. All
methods shown at $50\%$ recall. {\small{}$\mbox{Det.}\negmedspace+\negmedspace\mbox{SE}\left(\mathbf{weak}\right)$}
denotes the model {\small{}$\mbox{Det.}\negmedspace+\negmedspace\mbox{SE}\left(\mbox{MCG}_{+}\cap\mbox{BBs}\right)$}
and {\small{}$\mbox{Det.}\negmedspace+\negmedspace\mbox{HED}\left(\mathbf{weak}\right)$}
denotes {\small{}$\mbox{Det.}\negthickspace+\negthickspace\mbox{HED}\left(\textrm{cons.\mbox{ S\&G}\ensuremath{\cap}BBs}\right)$}.
As the quantitative results indicate, qualitatively our weakly supervised
results are on par to the fully supervised ones.}
\end{figure*}

\end{document}